\documentclass[10pt]{article}
\usepackage{arxiv}

\usepackage{microtype}
\usepackage{graphicx}
\usepackage{subfigure}
\usepackage{wrapfig}
\usepackage{booktabs} 
\usepackage{natbib}
\usepackage{hyperref}

\usepackage{amsmath}
\usepackage{amssymb}
\usepackage{mathtools}
\usepackage{amsthm}

\usepackage{hyperref}
\usepackage{url}

\usepackage{epsfig} 
\usepackage{times} 
\usepackage{booktabs}
\usepackage{wrapfig}
\usepackage{xcolor}
\usepackage{subfigure}

\usepackage{algorithm}
\usepackage{algorithmicx}
\usepackage{enumitem}
\setlist{nosep}

\usepackage{pifont}

\usepackage{float}
\usepackage{xspace}

\usepackage[capitalize,noabbrev]{cleveref}

\theoremstyle{plain}
\newtheorem{theorem}{Theorem}[section]
\newtheorem*{theorem*}{Theorem}
\newtheorem{proposition}[theorem]{Proposition}
\newtheorem{example}[theorem]{Example}
\newtheorem{lemma}[theorem]{Lemma}

\theoremstyle{definition}

\newtheorem{assumption}[theorem]{Assumption}
\theoremstyle{remark}

\usepackage[textsize=tiny]{todonotes}

\ifodd 1

\newcommand{\ty}[1]{\textbf{\color{blue}(Tongxin: #1)}}
\newcommand{\com}[1]{\textbf{\color{red}(comment: #1)}}
\newcommand{\res}[1]{\textbf{\color{magenta}(RESPONSE: #1)}} 
\newcommand{\zc}[1]{\textbf{\color{red}(Zhongzhu: #1)}} 
\newcommand{\kj}[1]{\textbf{\color{orange}(Kun: #1)}}

\newcommand{\expg}[0]{\texttt{ExpGaussian}}

\else

\newcommand{\ty}[1]{}
\newcommand{\zc}[1]{}
\newcommand{\kj}[1]{}

\newcommand{\com}[1]{}
\newcommand{\res}[1]{}

\newcommand{\expg}[0]{\texttt{ExpGaussian}}
\fi

\title{Performative Federated Learning: \\A Solution to Model-Dependent and Heterogeneous Distribution Shifts}

\author{%
  Kun Jin \footnotemark[1]\\
  University of Michigan \\
  \texttt{kunj@umich.edu}
  \And
  Tongxin Yin \footnotemark[1] \\
  University of Michigan \\
  \texttt{tyin@umich.edu}
  \And
  Zhongzhu Chen \footnotemark[1] \\
  University of Michigan\\
  \texttt{zhongzhc@umich.edu} \\
  \And
  Zeyu Sun\\
  University of Michigan\\
  \texttt{zeyusun@umich.edu}\\
  \AND
  Xueru Zhang \\
  The Ohio State University \\
  \texttt{zhang.12807@osu.edu}
  \And
  Yang Liu \\
  University of California, Santa Cruz\\
  ByteDance AI Lab \\
  \texttt{yangliu@ucsc.edu}
  \And
  Mingyan Liu \\
  University of Michigan \\
  \texttt{mingyan@umich.edu}
}

\begin{document}
\maketitle
\renewcommand{\thefootnote}{\fnsymbol{footnote}}
\footnotetext[1]{These authors contributed equally to this work.}

\begin{abstract}
We consider a federated learning (FL) system consisting of multiple clients and a server, where the clients aim to collaboratively learn a common decision model from their distributed data. Unlike the conventional FL framework that assumes the client's data is static, we consider scenarios where the clients' data distributions may be reshaped by the deployed decision model. In this work, we leverage the idea of distribution shift mappings in \textit{performative} \textit{prediction} to formalize this model-dependent data distribution shift and propose a performative federated learning framework. 
We first introduce necessary and sufficient conditions for the existence of a unique performative stable solution  and characterize its distance to the performative optimal solution. Then we propose the performative \texttt{FedAvg} algorithm and show that it converges to the performative stable solution at a rate of  $\mathcal{O}(1/T)$ under both full and partial participation schemes. In particular, we use novel proof techniques and show how the clients' heterogeneity influences the convergence. Numerical results validate our analysis and provide valuable insights into real-world applications.
\end{abstract}

\section{Introduction} \label{sec:intro}

Traditional learning problems typically assume data distributions to be static. For applications such as face recognition, this is largely true and designing algorithms under such an assumption in general does not impact learning efficacy.  This,  however, is not true in many other domains.  In some cases, there may be a natural evolution and shift in the distribution, e.g., in weather and climate data, in which case new data need to be acquired periodically and the algorithm re-trained to remain up to date.  In other cases, the distribution shift is the result of the very learning outcome, when individuals respond to the algorithmic decisions they are subjected to.  For instance, when users with certain accents perceive larger-than-acceptable errors from a speech recognition software and therefore stop using it, this can directly impact the type of speech samples collected by the software used for training the next generation of the product.  Another example is ``gaming the algorithm'', where users through honest or dishonest means attempt to improve critical features so as to obtain a favorable decision by the algorithm (e.g., in loan approvals or job applications).  This again can directly lead to the distributional change in features and label that the algorithm relies on for decision making. 

This latter type of distribution shifts, one of particular interest, where
the deployed model itself can trigger changes in the data distribution and influence the objective, said to be \emph{performative}. Performing prediction in the presence of such distribution shift is called \emph{performative prediction} \cite{ICML_2020_Perdomo_PP}. Typical scenarios of performative prediction include \emph{strategic learning} \cite{Hardt2016, dong2018strategic, Milli2019, Hu2019,  braverman2020role, chen2020strategic, pmlr-v119-miller20b, shavit2020causal,Haghtalab2020, Kleinberg2020,NEURIPS_2021_Zrnic}.

Performative prediction has been primarily studied in a centralized setting, with fruitful literature including the convergence analysis \cite{NIPS_2020_Mendler, ArXiv_2020_Drusvyatskiy, ArXiv_2020_Brown, AISTATS_2022_Li,IEEE_CSL_2021_Wood} and algorithm development \cite{ICML_2021_Zachary,AISTATS_2022_Zachary,ICML_2021_Miller,AAAI_2022_Ray}.

In modern large-scale machine learning, distributed learning offers greater privacy protection and better avoids the computational resource bottlenecks compared to centralized learning, and federated learning (FL) is one of the most popular examples.
Here the issue of distribution shift is further compounded due to data heterogeneity in a distributed setting. Specifically, the distributed data sources can be heterogeneous in nature, and their respective distribution shifts can also be different.  
Prior works in FL systems that address data distribution shifts, such as \cite{guo2021towards,casado2022concept,DBLP:journals/corr/abs-2002-08782,hosseinalipour2022parallel,zhu2021diurnal,eichner2019semi,ding2020distributed}, typically do not consider shifts in local distributions 
at the client end induced by the model. In this work, we propose the \emph{performative federated learning} framework to study and handle such data shifts in FL.

Extending the current results in performative prediction to the decentralized FL has a number of challenges. To highlight a few: 1) \emph{Data heterogeneity:} As already one of the major difficulties in FL, tackling data heterogeneity faces additional challenges when taking the disparity of client distribution shift into consideration. 2) \emph{Central $\rightleftarrows$ Local:} During training, clients receive the aggregated model at certain steps and train from it. While fitting better as an entity, such aggregation may fail to fit well on each client, which may lead to more severe shifting issues. 3) \emph{heterogeneity in shift:} some clients may be more sensitive to the deployed decisions and have more drastic data shifts than other clients, e.g., due to different manipulation costs in strategic learning.

Toward this end, we formally introduce the {\em performative FedAvg} algorithm, or \texttt{P-FedAvg}, and establish its convergence.  Our main findings are as follows. 
\begin{itemize}
    \setlength\itemsep{0.1em}

    \item We prove the uniqueness of the performative stable (PS) solution reached by the algorithm, and show that it is a provable approximation to the performative optimal (PO) solution under mild conditions. Both solutions will be formally defined in Section \ref{subsec:system_settings_objectives}. More interestingly, we show that the stable solution has its own game-theoretic interpretation as the fixed point of the best response dynamics. 
    \item We show in Section \ref{subsec:discussion_alg_sol} that the \texttt{P-FedAvg} algorithm converges to the performative stable solution and has a $\mathcal{O}(1/T)$ convergence rate with both the full and partial participation schemes under mild assumptions similar to those in prior works. 
    \item In doing so we also  introduce some novel proof techniques: we prove  convergence without a bounded gradient assumption. This technique can be directly applied to conventional FL, which is a special case of the performative setting.
\end{itemize}

\subsection{Related Works} 

\textbf{Federated Learning.}
Our work is strongly related to the literature on federated learning (FL). Although many studies have tried
to address client heterogeneity in FL through constrained gradient optimization and knowledge
distillation \cite{li2020federated,karimireddy2020scaffold,wang2020tackling,haddadpour2021federated,zhu2021data,li2019fedmd,lin2020ensemble}, most of them still assume the data is static without considering the distribution shifts. To the best of our knowledge, only a few recent works consider distribution shifts in FL \cite{guo2021towards,casado2022concept,DBLP:journals/corr/abs-2002-08782,hosseinalipour2022parallel,zhu2021diurnal,eichner2019semi,ding2020distributed}. For example, \cite{guo2021towards} considered FL with time-evolving clients where the  time-drift of each client is modeled as a \textit{time-independent} additive noise with \textit{zero-mean} and \textit{bounded variance}. \cite{casado2022concept} proposed an FL algorithm adaptable to distribution drifts; it monitors the confidence scores of the model prediction throughout the learning process and assumes the drift happens whenever there is a  substantial drop in confidence scores. \cite{DBLP:journals/corr/abs-2002-08782} also studied dynamic FL and assumed the true model under time-evolving data follows a \textit{random walk}. \cite{hosseinalipour2022parallel} considered FL with dynamic clients and modeled the drift using the variation in \textit{local loss} over two consecutive time steps. \cite{zhu2021diurnal,eichner2019semi,ding2020distributed} considered the \textit{periodical} distribution shift of client population in FL; they assume the
\textit{block-cyclic} structure where the clients from two different time zones alternately participate in training.  

\textbf{Performative Prediction.} In addition to the one we discussed in the introduction that focuses on the centralized setting for performative prediction, more recently, \cite{ArXiv_2022_Li_MPP} formalize the multi-agent/player performative predictions where agents try to learn a common decision rule but have heterogeneous distribution shifts (responses) to the model, and study the convergence of decentralized algorithms to the PS solution. The decentralized performative predictions capture the heterogeneity in agents'/clients' responses to the decision model and avoid centralized data collection for training. This work provides inspiration for our formulation of the performative federated learning framework, and our proposed \texttt{P-FedAvg} can be viewed as a substantial algorithmic extension that supports unbalanced data, much less frequent synchronizations, and partial device participation.
\cite{Narang_2022_ArXiv} propose a decentralized multi-player performative prediction framework where the players react to competing institutions’ actions. \cite{raab2021unintended} proposes a replicator dynamics model with label shift.

\textbf{Strategic Classification and Regression.}
As discussed in \cite{ICML_2020_Perdomo_PP}, performative prediction can be used to solve repeated strategic classification and regression problems. We can use Stackelberg games to model these problems, where the decision maker moves in the first stage by designing, publishing, and committing to a decision rule, then the agents move in the second stage, best responding to the decision rule by manipulating their features to get more desirable decision outcomes, and such manipulation can be modeled by the distribution shift mappings. Conventional strategic learning literature focus on finding the Stackelberg equilibrium \cite{Hardt2016,Kleinberg2020,shavit2020causal,Haghtalab2020}, i.e., the PO solution where the decision maker and the agents know each others' utilities, whereas performative prediction can find the PS solution in repeated strategic learning problems regardless of the knowledge on the utilities.

\section{Problem Formulation} \label{sec:formulation}

In this section, we formulate the performative federated learning problem, define the learning objective, and introduce our performative federated learning algorithm to optimize the objective function. 

To help with the understanding of performative federated learning, we first recall the performative prediction problem \cite{ICML_2020_Perdomo_PP}. Consider a typical loss minimization problem where the data distribution experiences a shift induced by the model parameter, expressed as a mapping $\mathcal{D}(\boldsymbol{\theta})$. The objective function is thus given by
\begin{equation*}
    f(\boldsymbol{\theta}) := \mathbb{E}_{Z \sim \mathcal{D}(\boldsymbol{\theta})}[\ell(\boldsymbol{\theta};Z)],
\end{equation*}
where $\ell$ denotes the loss function. Then the performative optimal (PO) solution is
 $  \boldsymbol{\theta}^{PO} := \arg \min_{\boldsymbol{\theta}} f(\boldsymbol{\theta})$.
\cite{ICML_2020_Perdomo_PP} also introduces a second,  decoupled objective function, also called the performatively stable (PS) model, which separates decision parameters ($\boldsymbol{\theta}$) from deployed parameters ($\Tilde{\boldsymbol{\theta}}$): 
\begin{equation*}    f(\boldsymbol{\theta};\Tilde{\boldsymbol{\theta}}) := \mathbb{E}_{Z \sim \mathcal{D}(\Tilde{\boldsymbol{\theta}})}[\ell(\boldsymbol{\theta};Z)]. 
\end{equation*}
Minimizing this objective 
 achieves minimal risk for the distribution  induced by the deployed parameters, eliminating  the need for
retraining, which makes it more practical. 
The PS solution is defined as
$    \boldsymbol{\theta}^{PS} := \arg \min_{\boldsymbol{\theta}} f(\boldsymbol{\theta};\boldsymbol{\theta}^{PS})$. 
\cite{ICML_2020_Perdomo_PP} showed that $\boldsymbol{\theta}^{PS} \neq \boldsymbol{\theta}^{PO}$ in general. We next consider a distributed setting and introduce performative federated learning. 

\subsection{System Settings and Objectives} \label{subsec:system_settings_objectives}
Consider a system with $N$ clients and a server, where the clients have feature distributions as $\mathcal{D}_i(\boldsymbol{\theta})$, supported on $\mathcal{Z} \subseteq \mathbb{R}^M$, and $\boldsymbol{\theta} \in \mathbb{R}^m$ denotes the decision (model) parameters deployed on the $i$-th client. We consider the general case where clients can have heterogeneous distributions $\mathcal{D}_i(\boldsymbol{\theta}) \neq \mathcal{D}_j(\boldsymbol{\theta})$, and each client represents a $p_i > 0$ fraction of the total data population, $\sum_{i=1}^N p_i = 1$. 

The system aims to minimize the weighted average loss across all agents, which is given by the performative optimal objective as follows 
\begin{equation} \label{eq:theta_PO}
    \boldsymbol{\theta}^{PO} := \arg \min_{\boldsymbol{\theta} \in \mathbb{R}^m} \sum_{i=1}^N p_i \mathbb{E}_{Z_i \sim \mathcal{D}_i(\boldsymbol{\theta})}[\ell(\boldsymbol{\theta};Z_i)].
\end{equation}

We note that our objective reduces to that in \cite{ArXiv_2022_Li_MPP} when $p_i = \frac{1}{N}$ for all $i$. This objective can typically model the strategic learning problem with different sub-populations in the system, where each client corresponds to a sub-population. Each sub-population may differ in some attributes so that they respond to the decision parameters differently, e.g., due to different action costs \cite{Milli2019, Hu2019, braverman2020role, Zhang_2022_ICML, EC_22_Jin}. The decision maker uses a common decision rule for the entire population and aims to minimize the expected loss, and $p_i$ represents the population fraction of each sub-population.
Correspondingly, the decoupled/performative stable objective is
\begin{equation*}
    f_{i}(\boldsymbol{\theta} ; \tilde{\boldsymbol{\theta}}) := \mathbb{E}_{Z_i \sim \mathcal{D}_i(\tilde{\boldsymbol{\theta}})}[\ell(\boldsymbol{\theta}; Z_i)], f(\boldsymbol{\theta};\tilde{\boldsymbol{\theta}}) := \sum_{i=1}^N p_i f_{i}(\boldsymbol{\theta}; \tilde{\boldsymbol{\theta}}),     
\end{equation*}
where the first argument denotes the client's decision parameter,
and the second argument the deployed parameters, which determine the distribution of the samples together with 
$\mathcal{D}_i(\cdot)$. The PS solution is 
\begin{equation}
    \begin{aligned}
        \boldsymbol{\theta}^{PS} := &~ \arg \min_{\boldsymbol{\theta}} \sum_{i=1}^N p_i \mathbb{E}_{Z_i \sim \mathcal{D}_i(\boldsymbol{\theta}^{PS})}[\ell(\boldsymbol{\theta};Z_i)]\\
        = &~ \arg \min_{\boldsymbol{\theta}} f(\boldsymbol{\theta};\boldsymbol{\theta}^{PS}).
    \end{aligned}
\end{equation}
Note that this is a fixed point equation with $\boldsymbol{\theta}^{PS}$ as a fixed point.

\subsection{Key Assumptions}

We make the following assumptions similar to \cite{ArXiv_2022_Li_MPP,ICML_2020_Perdomo_PP}:
\begin{assumption}[Strong Convexity] \label{asm:a1_obj_str_conv}
  \label{asm:convex} Given any $\tilde{\boldsymbol{\theta}} \in \mathbb{R}^m$, $f(\cdot, \tilde{\boldsymbol{\theta}})$ is $\mu$-strongly convex in $\boldsymbol{\theta}$, i.e., $
        f(\boldsymbol{\theta}';\tilde{\boldsymbol{\theta}}) \geq f(\boldsymbol{\theta};\tilde{\boldsymbol{\theta}}) + \langle \nabla f(\boldsymbol{\theta};\tilde{\boldsymbol{\theta}}) , \boldsymbol{\theta}' - \boldsymbol{\theta} \rangle + \frac{\mu}{2} \| \boldsymbol{\theta}' - \boldsymbol{\theta} \|_2^2, \forall \boldsymbol{\theta}', \boldsymbol{\theta} \in \mathbb{R}^K$.
\end{assumption}

In Assumption \ref{asm:a1_obj_str_conv}, we do not require strong convexity for every single $f_i$ but only the weighted average $f$.

\begin{assumption}[Smoothness] \label{asm:a2_obj_smooth}
     The loss function $\ell(\boldsymbol{\theta}; z)$ is $L$-smooth, i.e., 
    \begin{equation*}
        \|\nabla \ell(\boldsymbol{\theta}; \boldsymbol{z}) - \nabla \ell(\boldsymbol{\theta}'; \boldsymbol{z}')\|_2 \leq L (\| \boldsymbol{\theta}- \boldsymbol{\theta}'\|_2 + \|\boldsymbol{z}-\boldsymbol{z}'\|_2).
    \end{equation*}
\end{assumption}

\begin{assumption}[Distribution Mapping Sensitivity] \label{asm:a3_map_sensitivity} \label{asm:sensitive} For any $i = 1,\dots, n$ there exists $\epsilon_i > 0$ such that
    \begin{equation*}
    \mathcal{W}_1(\mathcal{D}_i(\boldsymbol{\theta}), \mathcal{D}_i(\boldsymbol{\theta}')) \leq \epsilon_i \|\boldsymbol{\theta} - \boldsymbol{\theta}'\|_2, ~~ \forall \boldsymbol{\theta}', \boldsymbol{\theta} \in \mathbb{R}^m,
    \end{equation*}
    where $\mathcal{W}_1(\mathcal{D}, \mathcal{D}'))$ is the 1-Wasserstein distance under $L_2$ norm between the distributions $\mathcal{D}, \mathcal{D}'$.
\end{assumption}

Assumption \ref{asm:a2_obj_smooth} and \ref{asm:a3_map_sensitivity} together induce the smoothness of $f_i(\cdot, \cdot)$, which is a result of Lemma 2.1 in \cite{drusvyatskiy2022stochastic} and will be used in the later proofs. 
\begin{lemma}[Continuity of $\triangledown f_i$] \label{lemma:fi_grad_continuity}
    Under Assumption \ref{asm:a2_obj_smooth} and \ref{asm:a3_map_sensitivity}, for any $\boldsymbol{\theta}_0, \boldsymbol{\theta}_1, \boldsymbol{\theta}, \hat{\boldsymbol{\theta}} \in \mathbb{R}^m$,
    \begin{equation*}
        \|\nabla f_i(\boldsymbol{\theta}_0; \boldsymbol{\theta}) 
        - \nabla f_i(\boldsymbol{\theta}_1; \hat{\boldsymbol{\theta}})\|_2 \leq
        L \|\boldsymbol{\theta}_0 - \boldsymbol{\theta}_1 \|_2 + L \epsilon_i \|\boldsymbol{\theta} - \hat{\boldsymbol{\theta}} \|_2.
    \end{equation*}
\end{lemma}

We introduce the following assumptions specifically made in decentralized performative predictions \cite{ArXiv_2022_Li_MPP}.

\begin{assumption}[Stochastic Gradient Variance Bound] \label{asm:a4_bound_var_sto_grad}
    For any $i=1, \dots, N$ and $\boldsymbol{\theta} \in \mathbb{R}^m$, there exists $\sigma \geq 0$ such that 
    \begin{equation*}
        \mathbb{E}_{Z_i \sim \mathcal{D}_i(\boldsymbol{\theta})}\| \nabla \ell (\boldsymbol{\theta}; Z_i) - \nabla f_i(\boldsymbol{\theta}; \boldsymbol{\theta}) \|_2^2 \leq \sigma^2 (1 + \|\boldsymbol{\theta} - \boldsymbol{\theta}^{PS}\|_2^2).
    \end{equation*}
\end{assumption}

\begin{assumption}[Local Gradient Bound] \label{asm:a5_bound_var_local_grad}
For any $i=1, \dots, N$ and $\boldsymbol{\theta} \in \mathbb{R}^m$, there exists $\varsigma \geq 0$ such that 
    \begin{equation*}
        \left\|\nabla f(\boldsymbol{\theta} ; \boldsymbol{\theta})-\nabla f_i(\boldsymbol{\theta} ; \boldsymbol{\theta})\right\|_2^2 \leq \varsigma^2(1+\|\boldsymbol{\theta}-\boldsymbol{\theta}^{PS}\|_2^2).
    \end{equation*}
\end{assumption}

Here we elaborate on Assumption \ref{asm:a5_bound_var_local_grad}, and explain reasons for using it over another commonly used assumption in federated learning \cite{ICLR_2020_Li_Convergence}, which is 
\begin{equation}\label{asm:uniform_gradnorm_bound}
    \mathbb{E}_{Z_i \sim \mathcal{D}_i(\boldsymbol{\theta})} [ \| \nabla \ell(\boldsymbol{\theta} ; Z_i) \|_2^2 ] \leq G^2.
\end{equation}
First, it can be shown that \eqref{asm:uniform_gradnorm_bound} implies Assumption \ref{asm:a5_bound_var_local_grad}, thus Assumption \ref{asm:a5_bound_var_local_grad} is weaker than \eqref{asm:uniform_gradnorm_bound}. To see this: when \eqref{asm:uniform_gradnorm_bound} holds, let $\varsigma^2 = 4G^2$, then $\left\|\nabla f(\boldsymbol{\theta} ; \boldsymbol{\theta})-\nabla f_i(\boldsymbol{\theta} ; \boldsymbol{\theta})\right\|_2^2 \leq 2 \left\|\nabla f(\boldsymbol{\theta} ; \boldsymbol{\theta})\|_2^2 + 2 \| \nabla f_i(\boldsymbol{\theta} ; \boldsymbol{\theta})\right\|_2^2 \leq 4 G^2 = \varsigma^2$.

We further give a concrete example where \eqref{asm:uniform_gradnorm_bound} does not hold but Assumption \ref{asm:a5_bound_var_local_grad} holds. 

\begin{example}
Suppose we have a two-client Gaussian mean estimation problem $\ell(\theta, Z) = \frac{1}{2}(\theta-Z)^2$ where $\theta, Z\in \mathbb{R}$, $\mathcal{D}_1(\theta) = \mathcal{N}(\frac{1}{2}\theta, \sigma^2)$, $\mathcal{D}_2(\theta) = \mathcal{N}(-\frac{1}{2}\theta, \sigma^2)$, and $p_1 = p_2 = \frac{1}{2}$. Then $\mathbb{E}_{Z_1 \sim \mathcal{D}_1(\theta)} [ \| \nabla \ell(\theta ; Z_1) \|_2^2 ] = \mathbb{E}_{Z_1 \sim \mathcal{D}_1(\theta)}[(\theta-Z_1)^2] = \sigma^2 + (\mathbb{E}_{Z_1 \sim \mathcal{D}_1(\theta)}[\theta-Z_1])^2 = \frac{1}{4} \theta^2 + \sigma^2$ and $\mathbb{E}_{Z_2 \sim \mathcal{D}_2(\theta)} [ \| \nabla \ell(\theta ; Z_2) \|_2^2 ] = \frac{9}{4} \theta^2 + \sigma^2$ which all go to infinity when $\theta$ goes to infinity. Thus \eqref{asm:uniform_gradnorm_bound} does not hold. On the other hand, $\nabla f_1(\theta; \theta)=\frac{1}{8} \theta$, $\nabla f_2(\theta; \theta)=\frac{9}{8} \theta$, and $\nabla f(\theta;\theta)= \frac{5}{8} \theta$, $\theta^{PS} = 0$, then by taking $\varsigma=\frac{1}{2}$, we can verify Assumption \ref{asm:a5_bound_var_local_grad} holds.
\end{example}

Secondly, \eqref{asm:uniform_gradnorm_bound} also implies Assumption \ref{asm:a4_bound_var_sto_grad}: when \eqref{asm:uniform_gradnorm_bound} holds, letting $\sigma^2 = G^2$ leads to $\mathbb{E}[\| \nabla l (\boldsymbol{\theta}; Z_i) - \nabla f_i(\boldsymbol{\theta}, \boldsymbol{\theta}) \|_2^2] \le \mathbb{E}[\| \nabla l (\boldsymbol{\theta}; Z_i) \|_2^2] \le G^2 = \sigma^2$. On the other hand, Assumption \ref{asm:a5_bound_var_local_grad} does not imply Assumption \ref{asm:a4_bound_var_sto_grad}.

It turns out that Assumption \ref{asm:a5_bound_var_local_grad} better characterizes the system heterogeneity, as we show how the heterogeneity impacts  convergence (more details are in Theorem \ref{thm:full}, \ref{thm:partial_w_rep_convergence}, and \ref{thm:partial_wo_rep_convergence}).

\subsection{Properties of the PS Solution}
Define the \textit{average sensitivity}  as $\overline{\epsilon} := \sum_{i=1}^N p_i \epsilon_i$, and the mapping $\Phi(\boldsymbol{\theta}) := \arg \min_{\boldsymbol{\theta}' \in \mathbb{R}^m} f(\boldsymbol{\theta}', \boldsymbol{\theta})$. Then we can establish the existence and uniqueness of the PS solution.
\begin{proposition}[Uniqueness of $\boldsymbol{\theta}^{PS}$]\label{prop:unique_PS}
     Under Assumptions \ref{asm:a1_obj_str_conv}, \ref{asm:a2_obj_smooth} and \ref{asm:a3_map_sensitivity}, 
    if $\overline{\epsilon} 
    < \mu/L$, then $\Phi(\cdot)$ is a contraction mapping with the unique fixed point $\boldsymbol{\theta}^{PS} = \Phi(\boldsymbol{\theta}^{PS})$; if $\overline{\epsilon} \ge \mu/L$, then there is an instance where any sequence generated by $\Phi(\cdot)$ will diverge.
\end{proposition}

Proposition \ref{prop:unique_PS} establishes a sufficient and necessary condition for the existence of $\boldsymbol{\theta}^{PS}$, similar to \cite{ArXiv_2022_Li_MPP}. This condition only depends on the average sensitivity $\overline{\epsilon}$, which implies that we may still have a unique performative stable solution $\boldsymbol{\theta}^{PS}$ for the whole system even if certain clients do not. The following proposition further validates the quality of $\boldsymbol{\theta}^{PS}$ in terms of its distance to $\boldsymbol{\theta}^{PO}$.

\begin{proposition}[Distance $\|\boldsymbol{\theta}^{PO}-\boldsymbol{\theta}^{PS} \|_2$ Bound]\label{prop:PS_PO_distance} 
Under Assumption \ref{asm:a1_obj_str_conv} and \ref{asm:a3_map_sensitivity}, suppose that the loss $\ell(\boldsymbol{\theta}; Z)$ is $L_z$-Lipschitz in $Z$, then for every performative stable solution $\boldsymbol{\theta}^{PS}$ and every performative optimal solution $\boldsymbol{\theta}^{PO}$, we have
$
\|\boldsymbol{\theta}^{PS} -\boldsymbol{\theta}^{PO}\|_2 \le \big(2L_z  \overline{\epsilon}\big)/\mu.
$
\end{proposition}
 The proofs of Proposition \ref{prop:unique_PS} and \ref{prop:PS_PO_distance} are in Appendix \ref{appensec:prop}.

\subsection{The \texttt{P-FedAvg} Algorithm}

In \texttt{P-FedAvg}, the clients communicate with the server  every $E$ local updates. Denote $\mathcal{I}_{E} := \{nE | n = 1, 2 , \dots\}$ as the set of aggregation steps. Next, we formalize the full and partial participation schemes of the proposed {\bf \texttt{P-FedAvg}}.

\textbf{Full client participation.} All clients communicate with the server at every aggregation step and update the local models $\boldsymbol{\theta}_i^{t+1}$ based on the following: let $Z_i^{t+1} \sim \mathcal{D}_i(\boldsymbol{\theta}_i^{t})$, then
\begin{align*}
    & \boldsymbol{w}_i^{t+1} = \boldsymbol{\theta}_i^{t} 
    - \eta_{t} \nabla \ell(\boldsymbol{\theta}_i^t; Z_i^{t+1}); \\
    & \boldsymbol{\theta}_i^{t+1} = \begin{cases}
        & \sum_{j=1}^N p_j \boldsymbol{w}_j^{t+1} ~~~~\text{if}~ t+1 \in \mathcal{I}_{E} \\
        & \boldsymbol{w}_i^{t+1} ~~~~~~~~~~~~~~~~~~\text{o.w.}
    \end{cases}
\end{align*}
\textbf{Partial client participation.} A more realistic setting that does not require the response of all clients' output at every aggregation step. In this case, the central server only collects the outputs of the first $K<N$ responded clients  at the aggregation step. Denote  the first $K<N$ responded clients in $t$-th step as a size-$K$ set $\mathcal{S}_t:=\{i_1,\ldots, i_K\}\in [N]$. Let $Z_i^{t+1} \sim \mathcal{D}_i(\boldsymbol{\theta}_i^{t})$, then
\begin{align*}
    & \boldsymbol{w}_i^{t+1} = \boldsymbol{\theta}_i^{t} 
    - \eta_{t} \nabla \ell(\boldsymbol{\theta}_i^t; Z_i^{t+1}); \nonumber \\
    & \boldsymbol{\theta}_i^{t+1} = \begin{cases}
        & \left(\begin{array}{l}
             \text {samples } \mathcal{S}_{t+1}, \text {and}   \\
             \text{average}\left\{\boldsymbol{w}_{t+1}^k\right\}_{k \in \mathcal{S}_{t+1}}
        \end{array}\right) ~~\text{if}~ t+1 \in \mathcal{I}_{E} \\
        & ~~\boldsymbol{w}_i^{t+1} ~~~~~~~~~~~~~~~~~~~~~~~~~~~~~~~~~~~~~~~~\text{o.w.}
    \end{cases}
\end{align*}
We further consider two schemes of partial participation:
\begin{enumerate}
    \item (\textbf{Scheme I}) The server establishes $\mathcal{S}_{t+1}$ by \textbf{\textit{i.i.d. with replacement}} sampling an index $k \in\{1, \cdots, N\}$ with probabilities $p_1, \cdots, p_N$ for $K$ times. Hence $\mathcal{S}_{t+1}$ is a multiset that allows an element to occur more than once. Then the server averages the parameters by $\boldsymbol{\theta}_i^{t+1}=\frac{1}{K} \sum_{k \in \mathcal{S}_{t+1}} \boldsymbol{w}^{t+1}_k$. This sampling scheme is first proposed in \cite{ArXiv_2018_Sahu} but the theoretical analysis was first done in \cite{ICLR_2020_Li_Convergence}.
    \item (\textbf{Scheme II}) The server samples $\mathcal{S}_{t+1}$ uniformly \textbf{\textit{without replacement}}. Hence each element in $\mathcal{S}_{t+1}$ only occurs once. Then the server averages the parameters by $\boldsymbol{\theta}_i^{t+1}=\sum_{k \in \mathcal{S}_{t+1}} p_k \frac{N}{K} \boldsymbol{w}^{t+1}_k$. Note that when the probabilities $\{p_k\}$ are not the same, one cannot ensure $\sum_{k \in \mathcal{S}_{t+1}} p_k \frac{N}{K}=1$ \cite{ICLR_2020_Li_Convergence}.
\end{enumerate}
Pseudo-codes of \texttt{P-FedAvg} are in Appendix \ref{appensec:pseudo_code}.

\textbf{Communication cost.} The \texttt{P-FedAvg} requires two rounds of communications, aggregation, and broadcast for every $E$ iterations. So at time step $T$, the system completes $2 \lfloor T/E \rfloor$ communications. We follow the setting in \cite{ICLR_2020_Li_Convergence} where the server aggregates based on the chosen scheme and broadcasts the aggregated parameters to all clients. 

Next, we will prove that \texttt{P-FedAvg} has $\mathcal{O}(1/T)$ convergence rate under the above assumption. As a supplement, we prove in Appendix \ref{appensec:alternative} that \texttt{P-FedAvg} also has $\mathcal{O}(1/T)$ convergence rate if we replace Assumption \ref{asm:a5_bound_var_local_grad} with \eqref{asm:uniform_gradnorm_bound}. 

\section{Convergence Analysis} \label{sec:convergence}

In this section, we show that the \texttt{P-FedAvg} converges to the unique $\boldsymbol{\theta}^{PS}$ at a rate of $\mathcal{O}(1/T)$ under the assumptions made in Section \ref{sec:formulation}, which holds for all above-introduced schemes. The key observation is that for sufficiently small and decaying learning rates, the effect of $E$ steps is similar to a one-step update with a larger learning rate in the static case, as stated in \cite{li2020federated} without the performative setting. Therefore, given appropriate sampling and updating schemes that satisfy the above assumptions, the global update behaves similarly to the repeated performative SGD in \cite{ICML_2020_Perdomo_PP}. We also show that partial device participation makes the averaged parameter sequence $\{\overline{\boldsymbol{\theta}}^{t}\}$ have the same mean as but a larger variance than the full participation, where the variance can be controlled with carefully chosen learning rates. It's worth noting that the heterogeneity of clients plays a key role in the convergence analysis, which we elaborate on below.

{\bf Quantifying the heterogeneity.} The client heterogeneity can be quantified by the consensus error $\sum_{i=1}^N p_i \| \boldsymbol{\theta}_i^t - \overline{\boldsymbol{\theta}}^{t} \|_2^2$, which is dynamic due to the nature of performative prediction. It depends on both the shift mappings $\mathcal{D}_i$ and the decision parameters. After every broadcast, the heterogeneity leads to heterogeneous distribution shifts, causing heterogeneous local updates and eventually resulting in the consensus error. The $\varsigma$ value in Assumption \ref{asm:a5_bound_var_local_grad} is also a good indicator for heterogeneity.

Next, we will first present the convergence analysis of the full participation scheme and later extend the analysis to partial participation schemes. Due to the complexity of analysis in the performative setting, we define the following constants for ease of analysis and clarity of presentation.

{\bf Constants independent of system design.} \\
$ \overline{\epsilon} := \sum_{i=1}^N p_i \epsilon_i, ~~~~ \epsilon_{max} :=  \max_i \epsilon_i$,\\ 
$\Tilde{\mu} :=  \mu - (1+\delta) \overline{\epsilon} L$, ~~~ $\delta >0$, \\
    $c_1 :=  \big(L (1+\epsilon_{max})^2\big)/(2 \delta \overline{\epsilon})$,\\
    $c_2 := 4 \big[\sigma^2 + L^2(1+\epsilon_{max})^2\big]$, 
    \\
    $c_3:= 6\big[2\sigma^2+ 3 L^2(1+\epsilon_{\max})^2\big]$,
    \\
    $c_4:= 16\sigma^2+12\varsigma^2+ (8\sigma^2+12\varsigma^2)/\mathbb{E}\|\overline{\boldsymbol{\theta}}^0-\boldsymbol{\theta}^{PS}\|_2^2$,
    \\
    $c_5:=  (48\sigma^2+36\varsigma^2)\mathbb{E}\|\overline{\boldsymbol{\theta}}^0-\boldsymbol{\theta}^{PS}\|_2^2 +(24\sigma^2+36\varsigma^2)$.
    
{\bf Constants related to system design (e.g., $E, K$).} \\
    $c_6 := (2E^2+3E+1)\log (E+1)$,
    \\
    $\tilde{\eta}_0:=\Tilde{\mu} / \big(2\sigma^2+( c_1c_3  + c_2/6 ) c_4c_6\big)$,
   \\ 
    $\hat{\eta}_0 :=\Tilde{\mu}/ \big(2\sigma^2+( c_1c_3  + c_2/6 ) c_4(2E^2-E)\log E\big)$,
    \\
    $B :=2\sigma^2+ ( 4 c_1 \hat{\eta}_0 + 4 c_2 \hat{\eta}_0^2 )c_5(2E^2-E)\log E$ ,
\\
    $B_1 := 2 \sigma^2  + (4c_1 \tilde{\eta}_0 + 4c_2 \tilde{\eta}_0^2 +1/K) c_5 c_6 $,
    
    $B_2 := 2 \sigma^2 + \big(4c_1 \tilde{\eta}_0 + 4c_2 \tilde{\eta}_0^2 + \frac{N-K}{K(N-1)}\big )c_5c_6$.

\subsection{Convergence of Full Participation}

\begin{theorem}[Full Participation]\label{thm:full}Consider \texttt{P-FedAvg} with full participation and diminishing step size $\eta_t = \frac{2}{\Tilde{\mu}(t+\gamma)}$, where $\gamma = \max\left\{ \frac{2}{\Tilde{\mu} \hat{\eta}_0}, E,\frac{2}{\Tilde{\mu}} \sqrt{(4E^2+2E)c_3}\right\}$. 
   Under Assumption \ref{asm:a1_obj_str_conv}, \ref{asm:a2_obj_smooth}, 
\ref{asm:a3_map_sensitivity}, \ref{asm:a4_bound_var_sto_grad}, \ref{asm:a5_bound_var_local_grad}, the following holds 
    \begin{align*}
        \mathbb{E} [\| \overline{\boldsymbol{\theta}}^{t} - \boldsymbol{\theta}^{PS} \|^2_2]  
        \leq & \frac{\upsilon}{\gamma+t},~\forall t
    \end{align*}
    where $\upsilon=\max \left\{\frac{4 B}{ \Tilde{\mu}^2},\gamma \mathbb{E}\|\overline{\boldsymbol{\theta}}^0-\boldsymbol{\theta}^{PS}\|_2^2\right\}$.
\end{theorem}

The proof of Theorem \ref{thm:full} can be found in Appendix \ref{appensec:full}. The key to the proof is that expected distance $\mathbb{E}\| \overline{\boldsymbol{\theta}}^t - \boldsymbol{\theta}^{PS} \|_2^2$ and expected consensus error $\sum_{i=1}^N p_i \mathbb{E}\| \boldsymbol{\theta}_i^{t-1} - \overline{\boldsymbol{\theta}}^{t} \|_2^2$ all depend on expected distance $\mathbb{E}\| \overline{\boldsymbol{\theta}}^{t-1} - \boldsymbol{\theta}^{PS} \|_2^2$ and expected consensus error $\sum_{i=1}^N p_i \mathbb{E}\| \boldsymbol{\theta}_i^{t-1} - \overline{\boldsymbol{\theta}}^{t-1} \|_2^2$ in the previous step. While we can establish a descent lemma for expected distance including the expected consensus error, it is impossible to establish one for expected consensus error, which makes it impossible to establish a joint descent lemma for expected distance and expected consensus error as in \cite{ArXiv_2022_Li_MPP}. Fortunately, consensus error will become zero at every aggregation step, which enables us to control expected consensus error at every step within a constant with a novel double-iteration technique under small enough step sizes. Then by relaxing the expected consensus error to the constant, we can establish a standard descent lemma in SGD analysis for expected distance. 

\subsection{Convergence of Partial Participation}

As mentioned in Section \ref{sec:formulation}, the partial participation scheme is more realistic  in federated learning \cite{ICLR_2020_Li_Convergence} and is of more interest since it reduces the stragglers' effect. 

We first present the convergence result of Scheme I.

\begin{theorem}[Partial Participation, Scheme I] \label{thm:partial_w_rep_convergence}Consider \texttt{P-FedAvg} with partial participation (scheme I) and a diminishing step size $\eta_t = \frac{2}{\Tilde{\mu}(t+\gamma)}$, where $\gamma = \max\left\{ \frac{2}{\Tilde{\mu} \tilde{\eta}_0}, E,\frac{2}{\Tilde{\mu}} \sqrt{(4E^2+10E+6)c_3}\right\}$.
   Under Assumption \ref{asm:a1_obj_str_conv}, \ref{asm:a2_obj_smooth}, \ref{asm:a3_map_sensitivity}, \ref{asm:a4_bound_var_sto_grad}, \ref{asm:a5_bound_var_local_grad}, the following holds 
    \begin{align*}
        \mathbb{E} [\| \overline{\boldsymbol{\theta}}^{t} - \boldsymbol{\theta}^{PS} \|^2_2]  
        \leq & \frac{\upsilon}{\gamma+t},~\forall t
    \end{align*}
    where $\upsilon=\max \left\{\frac{4 B_1}{ \Tilde{\mu}^2},\gamma \mathbb{E}\|\overline{\boldsymbol{\theta}}^0-\boldsymbol{\theta}^{PS}\|_2^2\right\}$.
\end{theorem}

Then we present the convergence result of Scheme II.
As discussed in Section \ref{sec:formulation}, we need probabilities $p_i = \frac{1}{N}, \forall i$ to ensure $\sum_{i \in \mathcal{S}_t} p_k \frac{N}{K}=1$.

\begin{theorem}[Partial Participation, Scheme II] \label{thm:partial_wo_rep_convergence}Consider \texttt{P-FedAvg} with partial participation (scheme II) and a diminishing step size $\frac{2}{\Tilde{\mu}(t+\gamma)}$, where $\gamma = \max\left\{ \frac{2}{\Tilde{\mu} \tilde{\eta}_0}, E,\frac{2}{\Tilde{\mu}} \sqrt{(4E^2+10E+6)c_5}\right\}$.
      Under Assumption \ref{asm:a1_obj_str_conv}, \ref{asm:a2_obj_smooth}, \ref{asm:a3_map_sensitivity}, \ref{asm:a4_bound_var_sto_grad}, \ref{asm:a5_bound_var_local_grad}, the following holds 
    \begin{align*}
        \mathbb{E} [\| \overline{\boldsymbol{\theta}}^{t} - \boldsymbol{\theta}^{PS} \|^2_2]  
        \leq & \frac{\upsilon}{\gamma+t},~\forall t
    \end{align*}
    where $\upsilon=\max \left\{\frac{4 B_2}{ \Tilde{\mu}^2},\gamma \mathbb{E}\|\overline{\boldsymbol{\theta}}^0-\boldsymbol{\theta}^{PS}\|_2^2\right\}$.
\end{theorem}

The proofs of Theorem \ref{thm:partial_w_rep_convergence} and \ref{thm:partial_wo_rep_convergence} can be found in Appendix \ref{appensec:partial}. Besides the technical difficulty as that of Theorem \ref{thm:full}, we also need to bound the variance of $\overline{\boldsymbol{\theta}}^t$ at the aggregation step. Fortunately, it can be bounded by the consensus error. Then by similar techniques as in the proof of Theorem \ref{thm:full}, we can establish a standard descent lemma in SGD analysis for the expected distance.

Scheme II requires $p_i = \frac{1}{N}, \forall i$, which violates the unbalanced nature of FL. One solution in \cite{ICLR_2020_Li_Convergence} is scaling the local objectives to $g_i(\boldsymbol{\theta}; \tilde{\boldsymbol{\theta}}) = p_i N f_i(\boldsymbol{\theta}; \tilde{\boldsymbol{\theta}})$, and then the global objective is a simple average of the scaled local objectives 
\begin{equation*}
    f(\boldsymbol{\theta}; \tilde{\boldsymbol{\theta}}) := \sum_{i=1}^N p_i f_{i}(\boldsymbol{\theta}; \tilde{\boldsymbol{\theta}}) = \frac{1}{N} \sum_{i=1}^N g_{i}(\boldsymbol{\theta}; \tilde{\boldsymbol{\theta}}).
\end{equation*}
We need to be careful with the Assumptions in Section \ref{sec:formulation} since scaling the objective will change those properties. The convergence theorems still hold if we replace $L, \mu, \sigma, \varsigma$ with $L' := q_{max} L, \mu' := q_{min} \mu, \sigma' := \sqrt{q_{max}} \sigma, \varsigma' := \sqrt{q_{max}} \varsigma$, where $q_{max} := N \cdot \max_i p_i, q_{min} := N \cdot \min_i p_i$.

\subsection{Discussions on the Algorithm and Solution} \label{subsec:discussion_alg_sol}
We will only discuss with respect to the aggregation step in this sub-section for convenience, denoted as $T \in \mathcal{I}_E$, then we can simply use $\frac{T}{E}$ when dividing $E$. Note for a general step $t$, we only need to use $\lfloor \frac{t}{E}\rfloor$ to obtain an integer.

\textbf{Choice of $E$.} We are interested in the total time we need to achieve an $\epsilon$ accuracy, and how this total time changes with $E$. We use our results in Theorem \ref{thm:full}, \ref{thm:partial_w_rep_convergence}, and \ref{thm:partial_wo_rep_convergence}, and denote $T_{\epsilon} := \frac{\upsilon}{\epsilon} - \gamma$ as the number of computation steps that is sufficient to guarantee an $\epsilon$-accuracy. To connect $T_{\epsilon}$ to the total time needed, suppose the expected time for each communication step is $C$ times the expected time of each computation step, then the total time required for $\epsilon$-accuracy is linear in $T_{\epsilon} + C \cdot \frac{T_{\epsilon}}{E}$. Below we separately analyze the influence of $E$ on $\frac{T_{\epsilon}}{E}$ and $T_{\epsilon}$, and then discuss how to choose the optimal $E$ for different $C$ values. 

Let $B_0:=B$ in Theorem \ref{thm:full} for full participation and $\gamma_i$ ($i=0,1,2$) denotes the $\gamma$ in Theorem \ref{thm:full}, \ref{thm:partial_w_rep_convergence}, and \ref{thm:partial_wo_rep_convergence} respectively. Then in Theorem \ref{thm:full}, \ref{thm:partial_w_rep_convergence}, and \ref{thm:partial_wo_rep_convergence}, $T_\epsilon$ is dominated by $\mathcal{O}\big(4 B_i/\Tilde{\mu}^2 +\gamma_i \mathbb{E} [\| \overline{\boldsymbol{\theta}}^{0} - \boldsymbol{\theta}^{PS} \|^2_2]\big)$ where $i=0,1,2$. From the definition, we know that $B_i$ ($i=0,1,2$) is almost a constant w.r.t. $E$ and $\gamma_i$ is of $\mathcal{O}(E^2\log E)$. This means that when $E$ grows, the total update steps to reach $\epsilon$-accuracy, $T_\epsilon$ will grow, while the number of aggregation steps needed, $\frac{T_\epsilon}{E}$ will first grow and then decrease. 

Now we consider $T_{\epsilon} + C \cdot \frac{T_{\epsilon}}{E}$, the total time needed to reach $\epsilon$-accuracy. From the above analysis, we know it is of order $\mathcal{O}(E^2\log E)+ C\cdot \mathcal{O}(E\log E)+C\cdot \mathcal{O}(\log E/E)$.
When communication is fast, i.e., $C$ is small, $\mathcal{O}(E^2\log E)$ is the dominating term, and we can focus more on the number of computation iterations $T_\epsilon$, and smaller $E$ values are preferable. However, when $C$ is large, $C\cdot \mathcal{O}(E\log E)+C\cdot \mathcal{O}(\log E/E)$ becomes the dominating term, and we should focus more on the number of communication rounds $\frac{T_\epsilon}{E}$ and some middle $E$ values are preferable.

\textbf{Choice of $K$.} Again $T_{\epsilon}$ is dominated by $\mathcal{O}\big(4 B_i/\Tilde{\mu}^2 +\gamma_i \mathbb{E} [\| \overline{\boldsymbol{\theta}}^{0} - \boldsymbol{\theta}^{PS} \|^2_2]\big)$ where $i=1,2$. Then by the formulae of $B_i$ ($i=1,2$), we know $T_\epsilon$ monotonically decreases with $K$, but the total communication time increases with $K$ due to more severe stragglers' effect. Generally, as we show in Theorem \ref{thm:partial_w_rep_convergence} and \ref{thm:partial_wo_rep_convergence}, the convergence rate has a weak dependence on $K$. We have empirically observed this phenomenon in Figure \ref{fig:impactofk:a}. Therefore, we can set $\frac{K}{N}$ to an appropriate small value to reduce the straggler's effect while keeping the convergence rate.

\textbf{Choice of sampling schemes.} We formalize the two sampling schemes in Section \ref{sec:formulation} and show their convergence properties in Theorem \ref{thm:partial_w_rep_convergence} and \ref{thm:partial_wo_rep_convergence}. We note that Scheme I has a desirable property that it naturally supports unbalanced clients, so if the server has control over the sampling, Scheme I should be chosen. 

But as discussed in \cite{ICLR_2020_Li_Convergence}, sometimes the server may have no control over the sampling and simply use the first $K$ received results for update. In this case, if the reception times from each client are IID random variables, we can treat this process as uniformly sampling $K$ out of $N$ at random without replacement. Theorem \ref{thm:partial_wo_rep_convergence} showed the convergence, and the discussion on scaling the objectives provides instructions on how to make the system work with arbitrary initial $p_1, \dots, p_N$ values. However, it's worth noting that when $p_1, \dots, p_N$ are highly non-uniform, the corresponding  $L', \sigma', \varsigma'$ values will be much larger than from $L, \sigma, \varsigma$, and $\mu'$ will be much smaller than $\mu$. Then by the formula of $\hat{\eta}_0$ and $\Tilde{\eta}_0$, we have to use much smaller starting learning rates and thus slower convergence. However, such a small learning rate may cause the model to fail to train at all. We also empirically show this in \cref{fig:highvarp}. 

However, an interesting observation is that when $p_i = \frac{1}{N}$, we empirically show in Figure \ref{fig:impactofh} and \ref{fig:credit}, Scheme II slightly outperforms Scheme I.

\textbf{Learning rate decay.} The learning rate decay is a necessity for stochastic gradient descent (SGD) to converge, even when clients have static, independent and individually distributed (IID) data. The decay is used in \cite{ArXiv_2022_Li_MPP} in decentralized performative prediction and the necessity for such decay is proved in \texttt{FedAvg} with static, non-IID clients. We also empirically show that constant learning rates fail to converge in \cref{fig:learningrate}.

\textbf{$\boldsymbol{\theta}^{PS}$ and $\boldsymbol{\theta}^{PO}$.} Here we discuss the relationship between the $\boldsymbol{\theta}^{PS}$ and $\boldsymbol{\theta}^{PO}$ solutions more in depth. In the strategic learning setting, \cite{ICML_2020_Perdomo_PP} showed that  $\boldsymbol{\theta}^{PO}$ is the Stackelberg equilibrium. It's worth noting that  $\boldsymbol{\theta}^{PS}$ is not merely an approximation to $\boldsymbol{\theta}^{PO}$, but a natural convergence point of the best response dynamics (BRD). More specifically, when the clients and the decision maker have no information about others' utilities, backward induction is unavailable, and playing the Stackelberg equilibrium is unrealistic. In this case, treating others' strategies in the previous time step as constants, and optimizing one's own strategy accordingly is a rational strategy. Such an optimization step is a best response, and in multi-round sequential strategic learning problems \cite{NEURIPS_2021_Zrnic}, the best responses can form the BRD, and $\boldsymbol{\theta}^{PS}$ is {\bf the convergence point of the BRD}. Although the decision maker's natural best response step is a risk minimization step, the gradient-based \texttt{P-FedAvg} can find the same $\boldsymbol{\theta}^{PS}$. Another interesting observation of $\boldsymbol{\theta}^{PS}$ is that if we remove the sequential decision nature, then $\{ \boldsymbol{\theta}^{PS}, \mathcal{D}_1(\boldsymbol{\theta}^{PS}), \dots, \mathcal{D}_N(\boldsymbol{\theta}^{PS}) \}$ is a {\bf Nash equilibrium} since no participant has an incentive to unilaterally deviate. 
\section{Numerical Experiments} \label{sec:numerical}

\subsection{Weighted Gaussian mean performative prediction}

As a numerical simulation, we perform \texttt{P-FedAvg} to estimate the mean of heterogeneous Gaussian data under performative effects and examine the impact of the hyperparameters, the sampling schemes, and client heterogeneity. We consider $N = 25$ clients, with the $i$-th client minimizing the loss function $\ell(\theta; Z_i) := (\theta - Z)^2 / 2$, $\theta, Z \in \mathbb{R}$ on data $Z_i\sim \mathcal{D}_i(\theta) := \mathcal{N}(m_i + \epsilon_i \theta, \sigma^2)$. For this loss function, we have $\mu = 1$, $L=1$. For $\overline{\epsilon} \in [0,1)$, the PS solution is $\theta^{PS} = \frac{\sum_{i=1}^N p_i m_i}{1 - \overline{\epsilon}}$; while $\theta^{PS}$ does not exist when $\overline{\epsilon} \geq 1$.  Denote the weighted average of $m_i$ as $\overline{m} = \sum_{i=1}^N p_i m_i$ and the variance as $\text{Var}(m) = \sum_{i=1}^N p_i (m_i - \overline{m})^2$. In experiment, we set $\overline{\epsilon} = 0.9$, $\overline{m} = 10$.

\begin{figure}[htbp]
    \centering
    \subfigure[Scheme I]{\label{fig:impactofe:a}\includegraphics[width=0.45\textwidth]{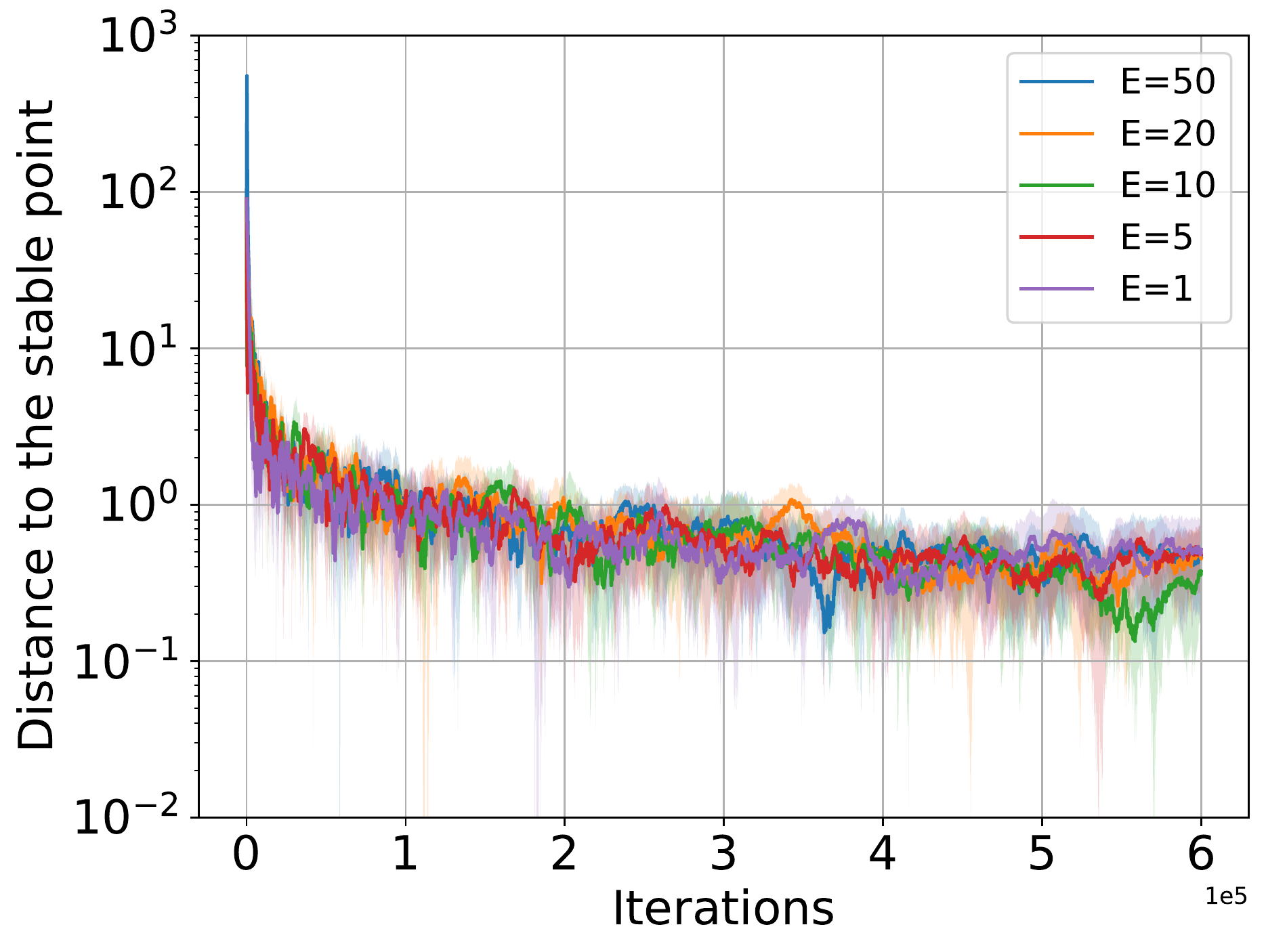}}
    \subfigure[Scheme II]{\label{fig:impactofe:b}\includegraphics[width=0.45\textwidth]{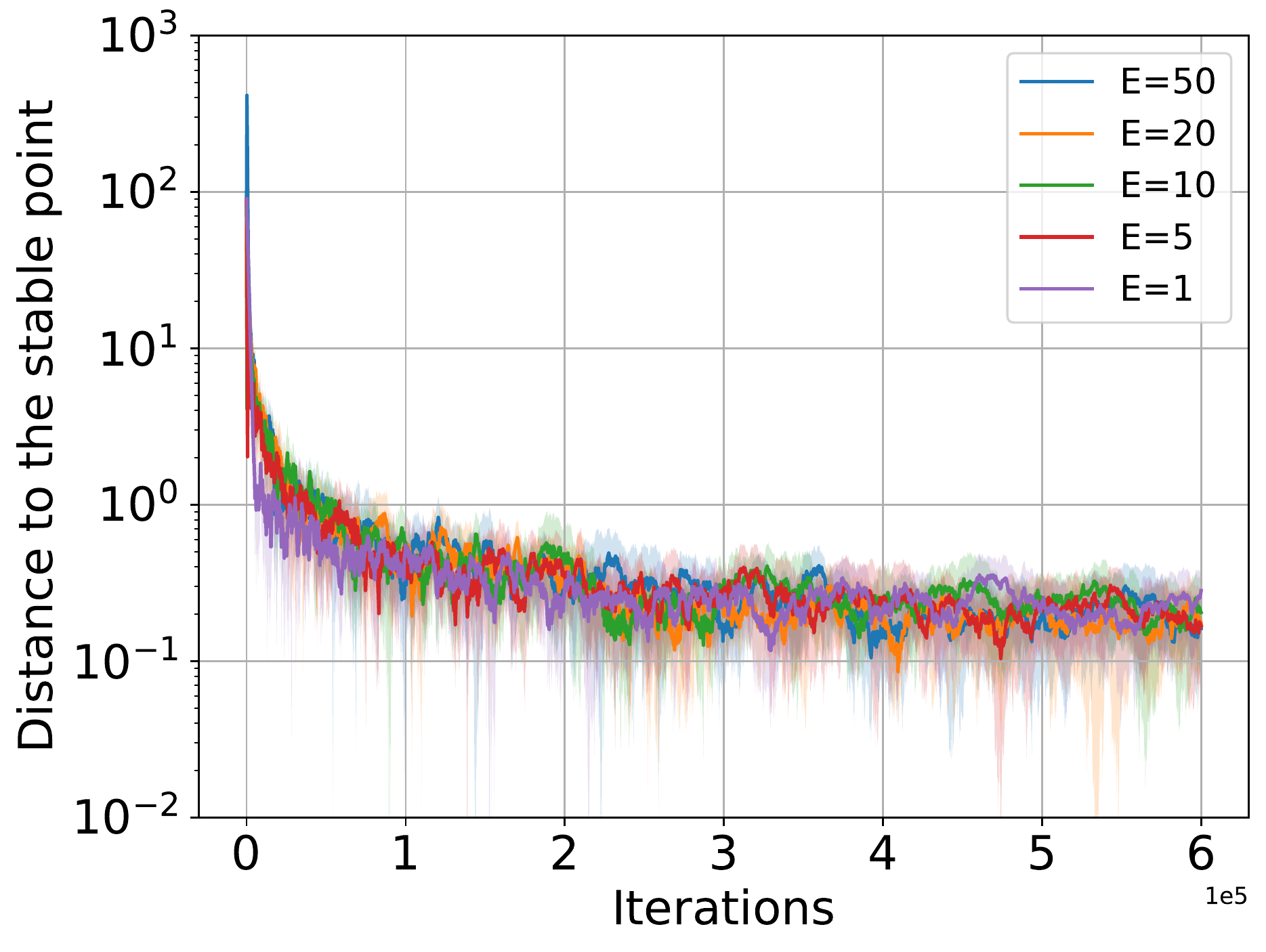}}
    \caption{Impact of E on Scheme I and Scheme II. 
    $K=20$, $\text{Var}(m) = 0.6$, $\text{Var}(\epsilon) = 0.1$ for both (a) and (b). For (b), $p_i = \frac{1}{25}$.}
    \label{fig:impactofe}
\end{figure}

\begingroup

 \begin{figure}
     \centering
     \includegraphics[width=0.45\textwidth]{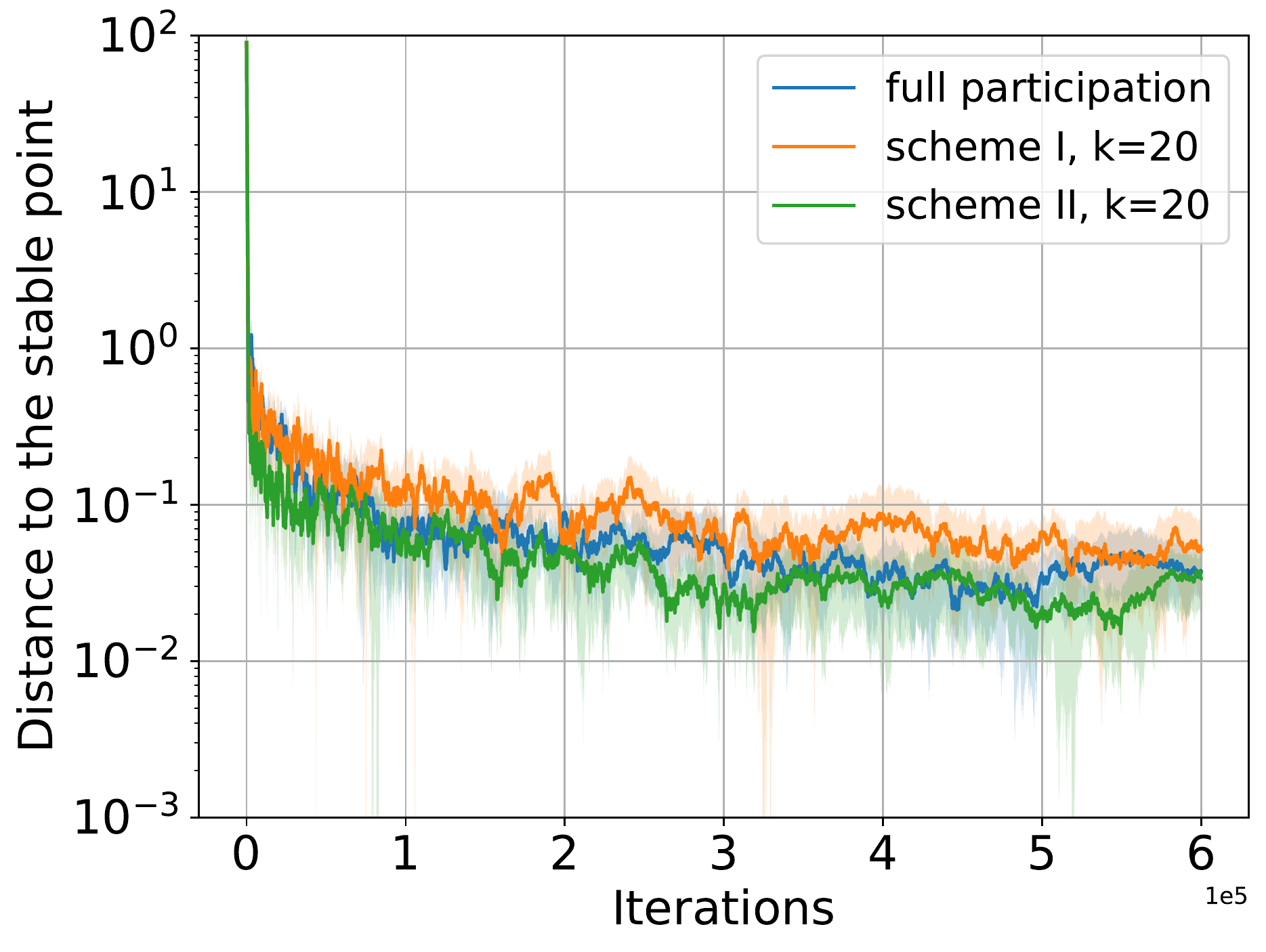}
     \caption{Distance to the performative stable solution vs. the number of iterations for full participation, Scheme I, and Scheme II.}
     \label{fig:convergence}
 \end{figure}
 
Figure \ref{fig:convergence} shows \texttt{P-FedAvg} converges to the performative stable solution in all three communication settings: full participation and the two schemes for the partial participation. Interestingly, partial participation with scheme II converges the fastest in this experiment. Despite the full participation scheme having the lowest upper bound on the number of iterations sufficient to convergence, our experimental results show that the actual convergence behaviors of all three schemes are very similar and weakly depend on $K$, especially when $p_i = \frac{1}{N}$.

{\bf Impact of $E$.} We conduct an experiment to compare the performance of our algorithm with a variety of $E$ values, under a homogeneous system. Figure \ref{fig:impactofe} shows the result on both sampling schemes, with $K=20$. A slightly larger $E$ leads to faster convergence. However, an extremely large $E$($E=50$ in the experiment) can also cause slower convergence. Since at this case, the clients deviate too much at each aggregation, which causes low efficiency issues. In real world scenarios, as the communication cost changes, $E$ should be carefully chosen.
 \endgroup
 
\begin{figure}[htbp]
    \centering
    \subfigure[Scheme I]{\label{fig:impactofk:a}\includegraphics[width=0.45\textwidth]{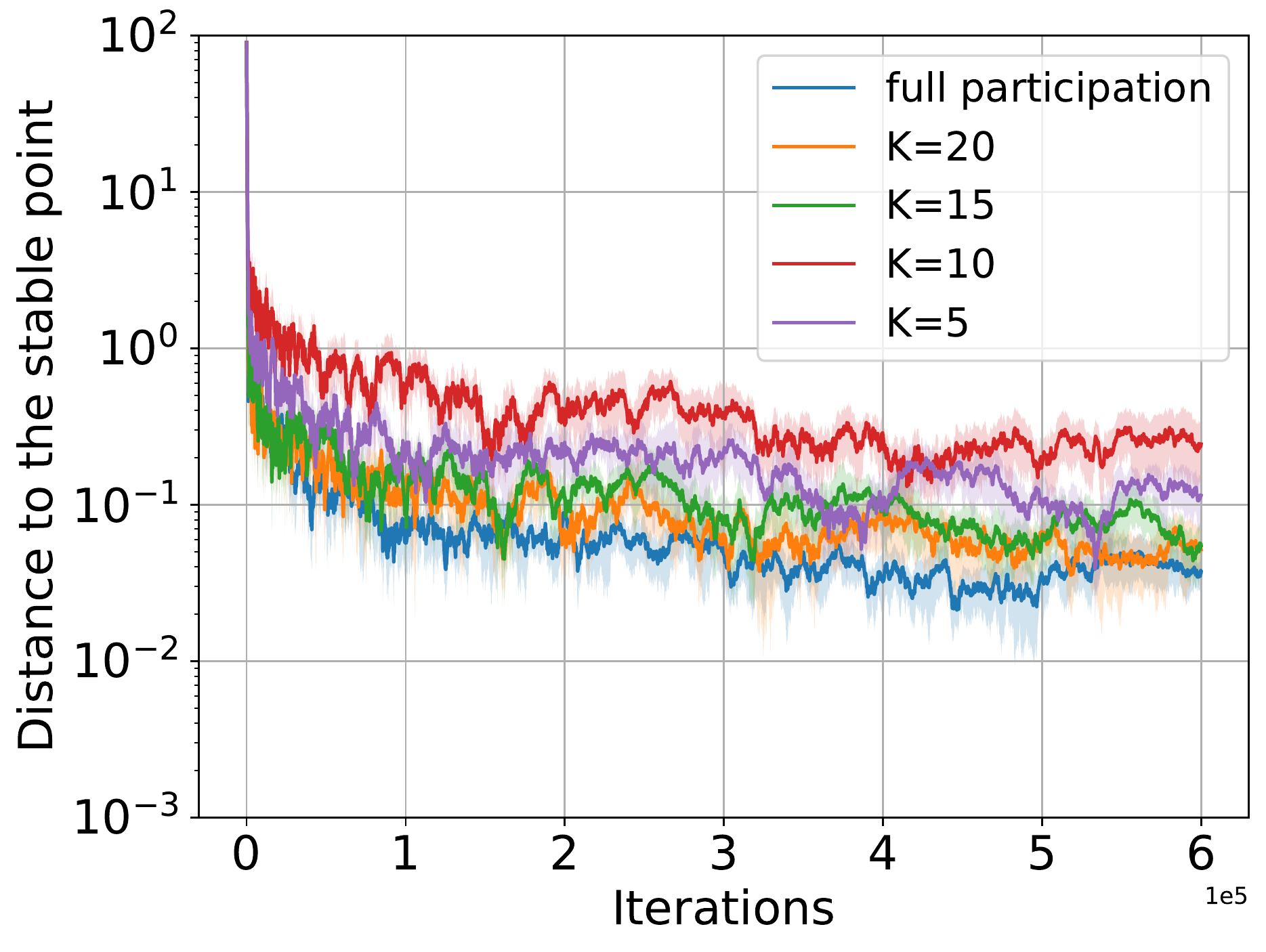}}
    \subfigure[Scheme II]{\label{fig:impactofk:b}\includegraphics[width=0.45\textwidth]{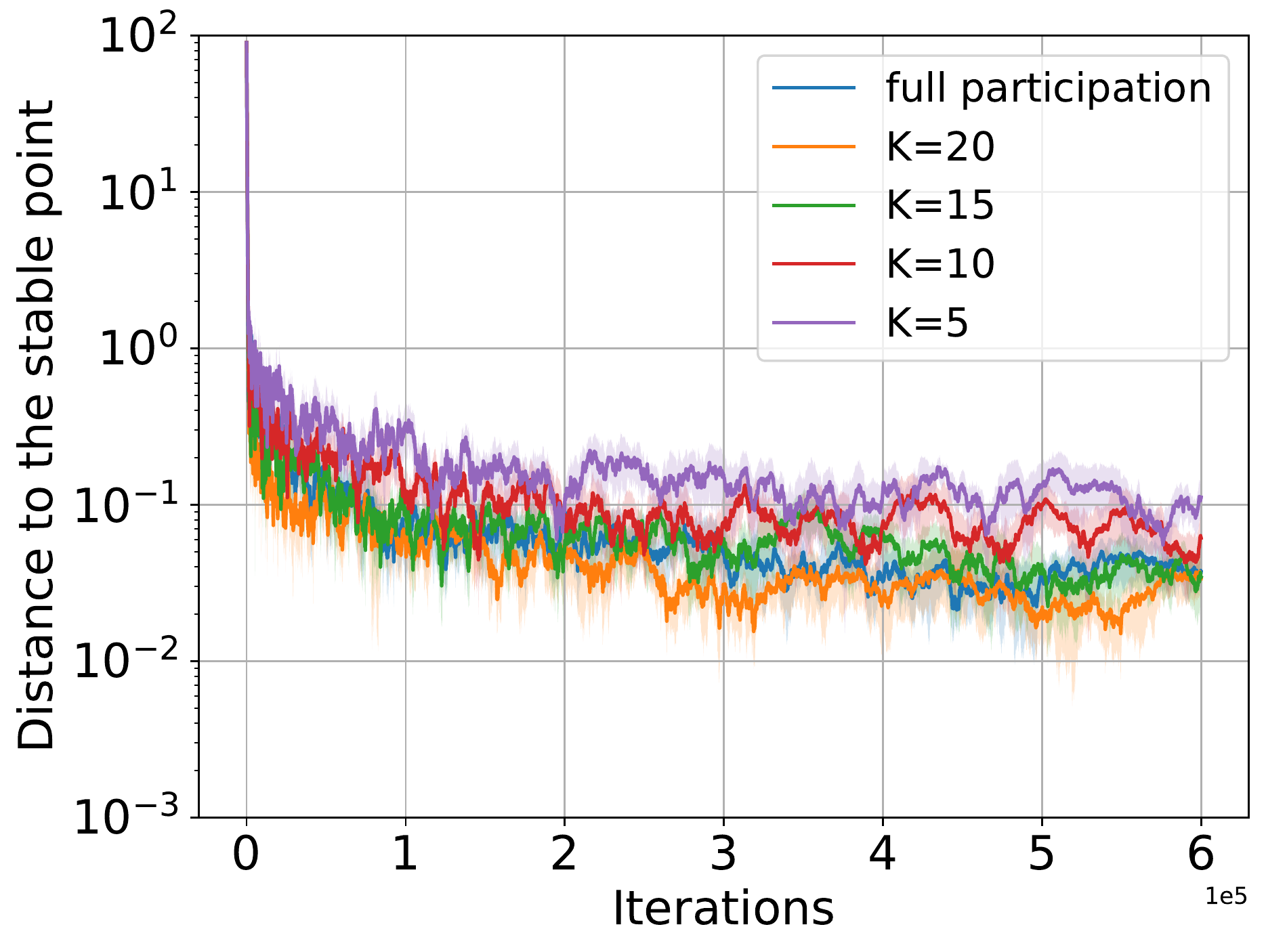}}
    \caption{Impact of K on Scheme I and Scheme II . 
    $E=5$, $\text{Var}(m) = 0.6$, $\text{Var}(\epsilon) = 0.1$ for both (a) and (b). For (b), $p_i = \frac{1}{25}$.}
    \label{fig:impactofk}
\end{figure}

{\bf Impact of $K$.} Figure \ref{fig:impactofk} shows the convergence of FedAvg under different k values, For scheme I, larger k leads to faster convergence. While for scheme II, as $k$ increasing, the convergence rate will first increase and then decrease. 

{\bf Impact of sampling schemes.} Figure \ref{fig:convergence} also compares different schemes. We can see if the clients' data are uniformly sampled ($p_i = \frac{1}{N}$), then scheme II achieves a better convergence rate, which conforms to our theoretical result because $B_1>B_2$.

\begin{figure}[htbp]  
\centering
    \subfigure[Heterogeneity in $m_i$]{\label{fig:impactofh:a}\includegraphics[width=0.45\textwidth]{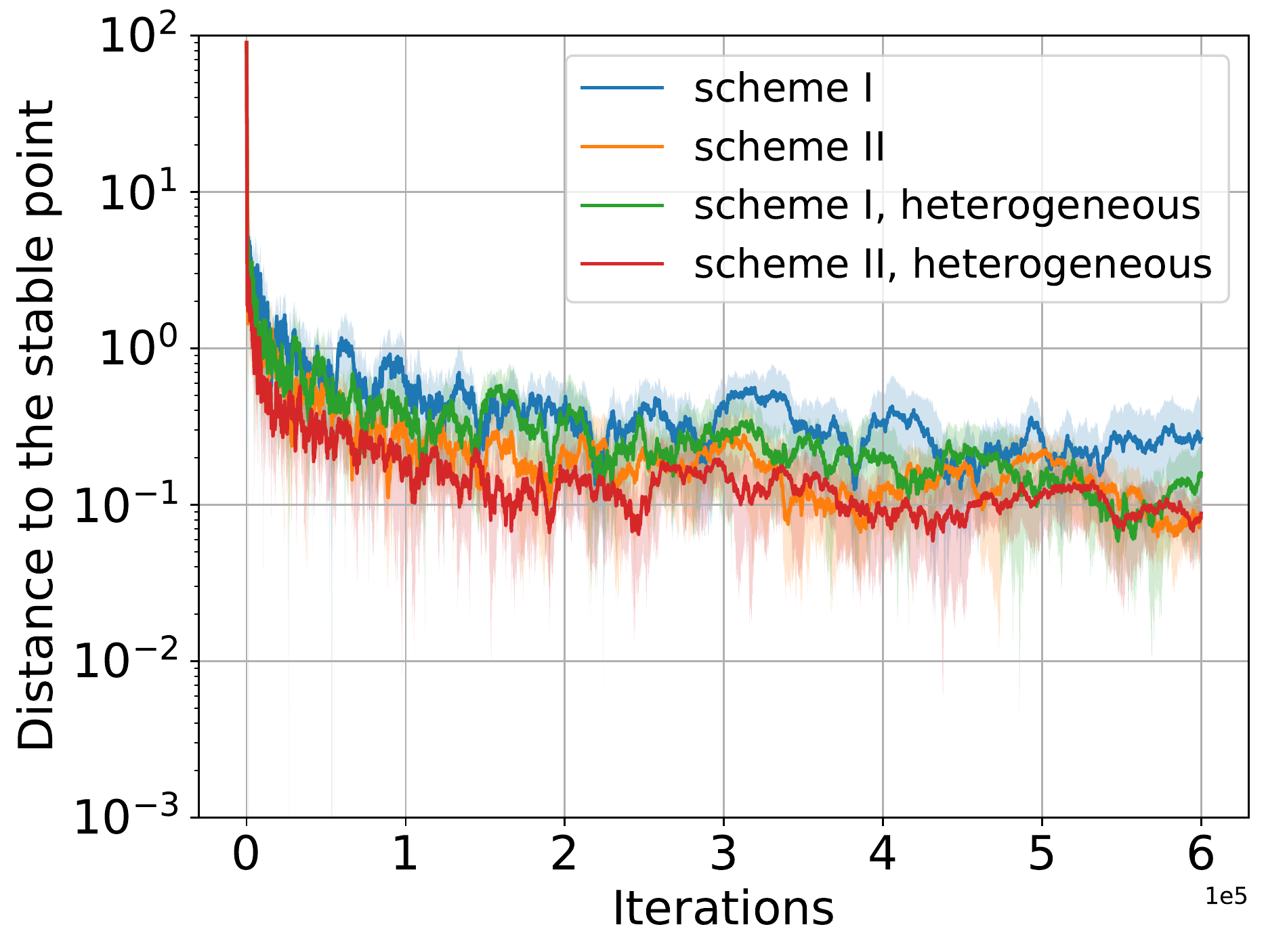}}
    \subfigure[Heterogeneity in $\epsilon_i$]{\label{fig:impactofh:b}\includegraphics[width=0.45\textwidth]{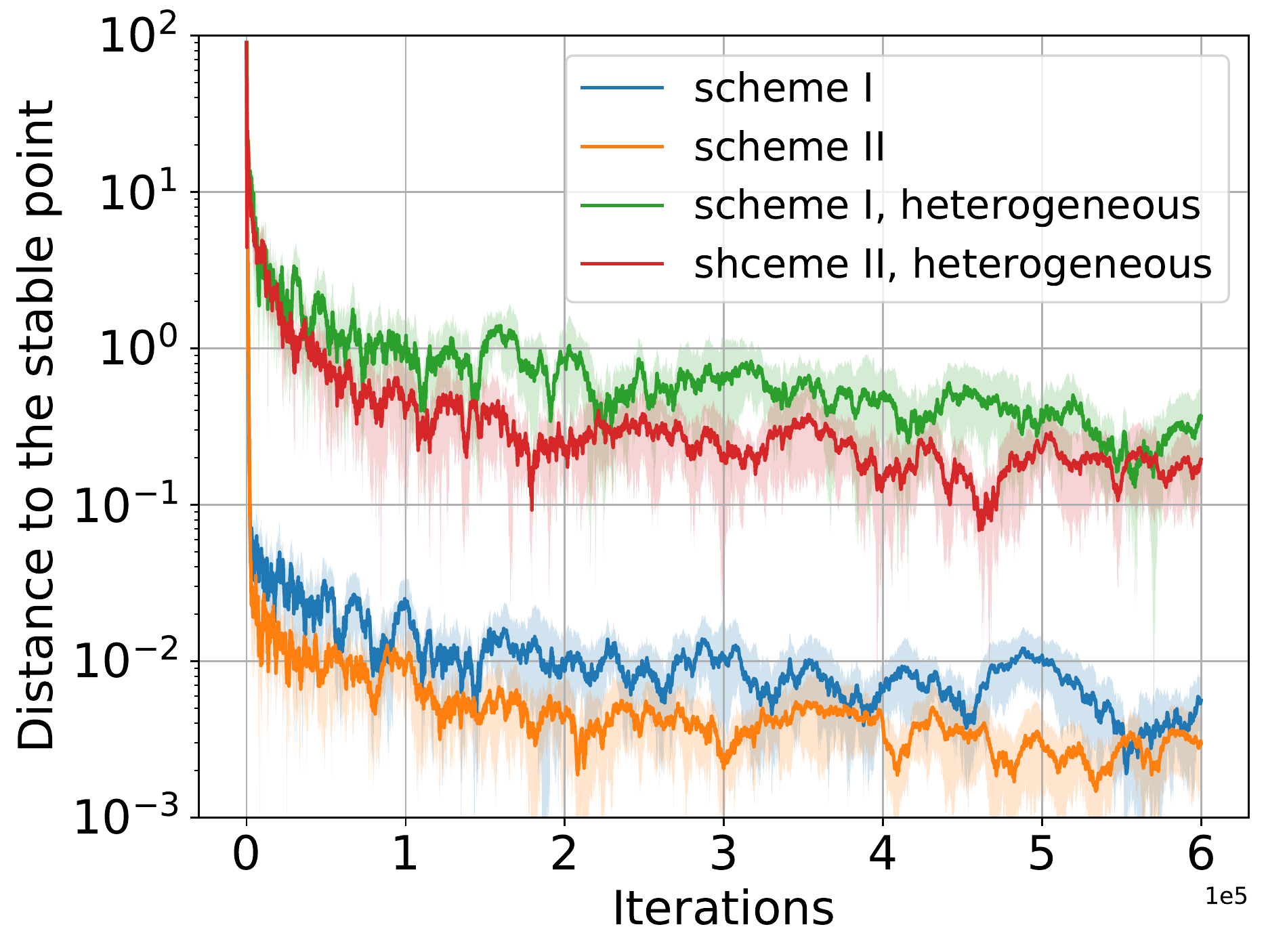}}
    \caption{Impact of heterogeneity on the two schemes of partial participation. The impact of $m_i$ is shown in (a) and the impact of $\epsilon_i$ is shown in (b). 
    $K=20, p_i = \frac{1}{25}$ and. In (a), $\text{Var}(m) = 6$ for hetergeneous case and $0$ for homogeneous case, $\text{Var}(\epsilon) = 0.1$. In (b), $\text{Var}(\epsilon) = 0.6$ for hetergeneous case and $0.1$ for homogeneous case, $\text{Var}(m) = 0.6$.}
    \label{fig:impactofh}
\end{figure}

{\bf Data heterogeneity and shifting heterogeneity.}
In Figure \ref{fig:impactofh} we test our algorithm under data heterogeneity. Specifically, we set $m$ and $\epsilon$ to have large variances, respectively. In this example, $m$ mainly captures the data heterogeneity and $\epsilon$ capture the shifting heterogeneity. This experiment shows our algorithm still converges under a certain amount of heterogeneity. Comparing the performance of our algorithm on both figures, we can see shifting heterogeneity is the main factor in performative federated learning. 

\subsection{Credit Score Strategic Classification}

To show the performance of \texttt{P-FedAvg} on a real world dataset, we follow \cite{ICML_2020_Perdomo_PP} and use the same Kaggle dataset \footnote{www.kaggle.com/competitions/GiveMeSomeCredit/data}, where a bank predicts whether loan applicants are creditworthy. 
The features consist of the information about an individual, and the target is 1 if the individual defaulted on a loan, and 0 otherwise. We use the same strategic setting as in \cite{ICML_2020_Perdomo_PP} where the applicants can manipulate their features in (1) revolving utilization of unsecured lines, (2) number of open credit lines and loans, and (3) number real estate loans or lines. The strength of manipulation for the $i$-th population is controlled by $\epsilon_i$. We equally partition the training set into 10 subsets and distributed it to 10 clients, and thus $p_i = 0.1, \forall i$. The sensitivities $\epsilon_i$ for the 10 clients are independently and uniformly sampled from $[0.9, 1.1]$. We set $K=5$ in partial participation. We train a logistic regression binary classifier. In each round of \texttt{P-FedAvg}, we perform $E=5$ gradient descent steps on a random minibatch of size 4. A discussion on the effect of the batch size can be found in Appendix~\ref{appensec:exp_credit}.

Figure~\ref{fig:credit} shows the loss function and the distance to the PS solution as the number of deployment rounds increases. The mean and 1 standard deviation error bar are generated from 5 experiments with different random seeds. Similar to the numerical simulation, the actual convergence behaviors of all three schemes are very similar.

\begin{figure} [htbp]
    \subfigure[Training losses]{\label{fig:credit_loss}\includegraphics[width=0.45\textwidth]{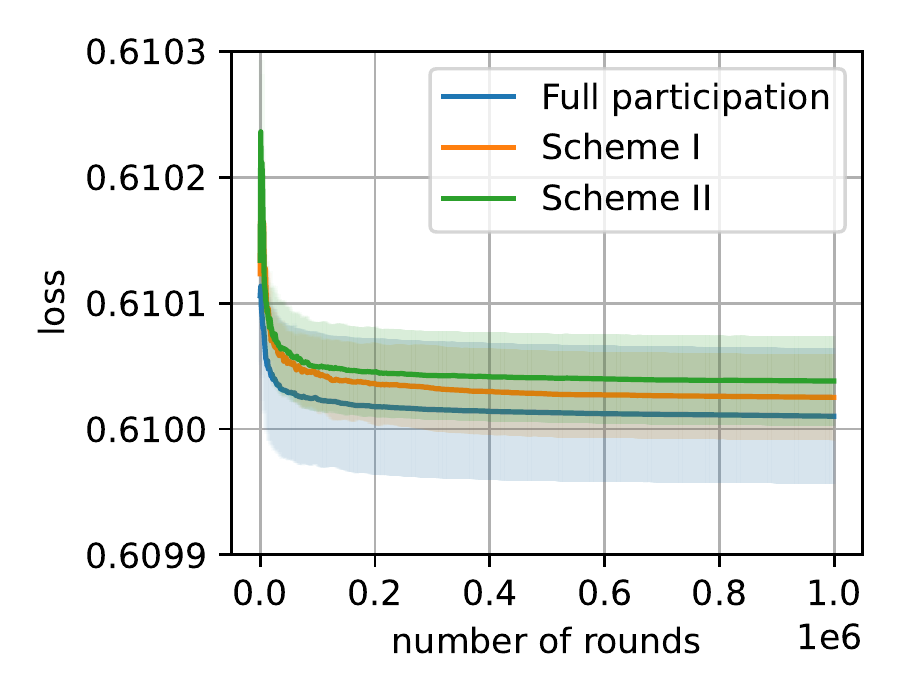}}
    \subfigure[Distances to $\boldsymbol{\theta}^{PS}$]{\label{fig:credit_dtheta}\includegraphics[width=0.45\textwidth]{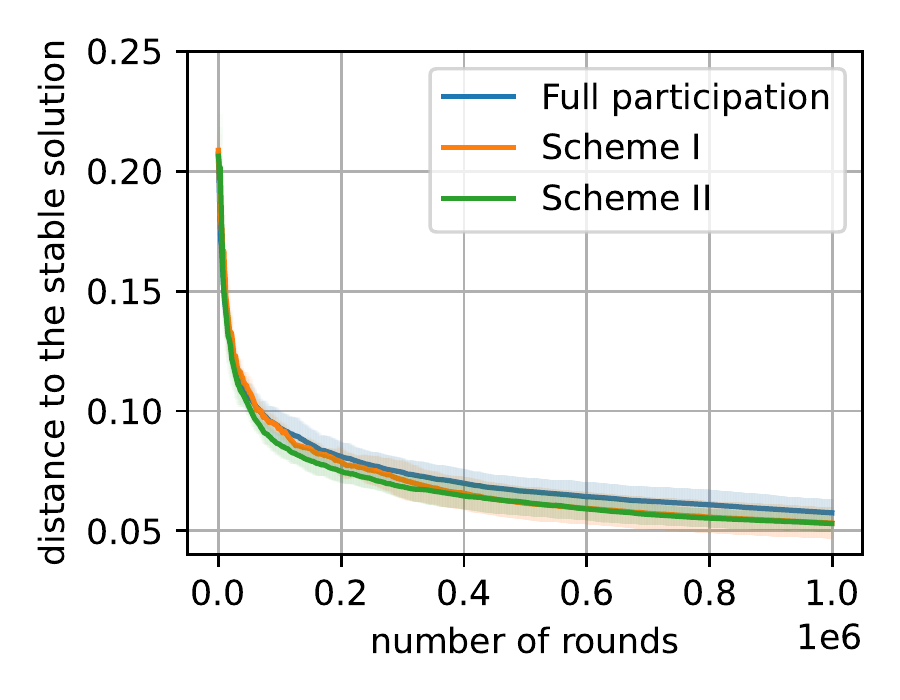}}
    \caption{The losses (a) and the distances to the PS solution (b) for the full participation, Scheme I and Scheme II.}
    \label{fig:credit}
\end{figure}

\section{Conclusion}
In this work, we leveraged the idea of distribution shift mappings in performative predictions to study federated learning problems where data shifts exists and such shifts are induced by the decision parameters. We formulated the performative federated learning problem and showed that a unique performative stable solution exists, which is a natural equilibrium of the iterative updating process between the server and the clients. Then we formalized the \texttt{P-FedAvg} algorithm and proved that both the full device participation and the partial device participation schemes have $\mathcal{O}(1/T)$ convergence rate to the performative stable solution. We also thoroughly discussed the impact of how some of the key system parameters influence the convergence, including the aggregation interval size, the number of sampled devices in partial participation, the sampling schemes, and the heterogeneity among clients. Our numerical results validate our theory and discussion, and provide valuable insights into the real-world applications of performative federated learning.

\bibliography{example_paper}
\bibliographystyle{plainnat}

\newpage
\appendix
\onecolumn
\section{Proof of Proposition \ref{prop:unique_PS} and \ref{prop:PS_PO_distance}} \label{appensec:prop}

\textbf{Proposition 2.5.} (Uniqueness of $\boldsymbol{\theta}^{PS}$)
    Under Assumptions \ref{asm:a1_obj_str_conv}, \ref{asm:a2_obj_smooth} and \ref{asm:a3_map_sensitivity}, define the map $\Phi : \mathbb{R}^m \mapsto \mathbb{R}^m$
    \begin{equation*}
        \Phi(\boldsymbol{\theta}) := arg \min_{\boldsymbol{\theta}' \in \mathbb{R}^M} f(\boldsymbol{\theta}', \boldsymbol{\theta})
    \end{equation*}
    If $\overline{\epsilon} := \sum_{i=1}^N p_i \epsilon_i < \mu/L$, then $\Phi(\cdot)$ is a contraction mapping with the unique fixed point $\boldsymbol{\theta}^{PS} = \Phi(\boldsymbol{\theta}^{PS})$. On the contrary, if $\overline{\epsilon} \ge \mu/L$, then there is an instance where any sequence generated by $\Phi(\cdot)$ will diverge.

\begin{proof}
This proof simulates the proof of Proposition 1 in \cite{ArXiv_2022_Li_MPP}.
   
    Fix $\boldsymbol{\theta}', \boldsymbol{\theta} \in \mathbb{R}^m$, the optimality condition implies that 
    \begin{equation*}
        \sum_{i=1}^N p_i \nabla f_i(\Phi(\boldsymbol{\theta}); \boldsymbol{\theta}) = \boldsymbol{0}, ~~ \sum_{i=1}^N p_i \nabla f_i(\Phi(\boldsymbol{\theta}'); \boldsymbol{\theta}') = \boldsymbol{0}
    \end{equation*}
    where the gradients are taken w.r.t the first argument in $f_i$. Then we have
    \begin{align*}
        0 = & \langle \boldsymbol{0}, \Phi(\boldsymbol{\theta})- \Phi(\boldsymbol{\theta}') \rangle \\
        = & \langle \sum_{i=1}^N p_i 
        \big(\nabla f_i(\Phi(\boldsymbol{\theta}); \boldsymbol{\theta}) 
        - \nabla f_i(\Phi(\boldsymbol{\theta}'); \boldsymbol{\theta}')\big), 
        \Phi(\boldsymbol{\theta})- \Phi(\boldsymbol{\theta}') \rangle.
    \end{align*}
    Rearranging the above equation and adding $\sum_{i=1}^N p_i f_i(\Phi(\boldsymbol{\theta}),\boldsymbol{\theta}')$ to both hand sides leads to
    \begin{align*}
        &  \langle \sum_{i=1}^N p_i 
        \big(\nabla f_i(\Phi(\boldsymbol{\theta}); \boldsymbol{\theta}') 
        - \nabla f_i(\Phi(\boldsymbol{\theta});\boldsymbol{\theta})\big), 
        \Phi(\boldsymbol{\theta})- \Phi(\boldsymbol{\theta}') \rangle \\
        = & \langle \sum_{i=1}^N p_i 
        \big(\nabla f_i(\Phi(\boldsymbol{\theta}); \boldsymbol{\theta}') 
        - \nabla f_i(\Phi(\boldsymbol{\theta}'); \boldsymbol{\theta}')\big), 
        \Phi(\boldsymbol{\theta})- \Phi(\boldsymbol{\theta}') \rangle.
    \end{align*}
    By strong convexity in assumption \ref{asm:a1_obj_str_conv}, we have
    \begin{equation*}
         f(\Phi(\boldsymbol{\theta});\boldsymbol{\theta}') \geq f(\Phi(\boldsymbol{\theta}');\boldsymbol{\theta}') + 
         \langle \nabla f(\Phi(\boldsymbol{\theta}');\boldsymbol{\theta}') ,
         \Phi(\boldsymbol{\theta})- \Phi(\boldsymbol{\theta}') \rangle 
         + \frac{\mu}{2} \| \Phi(\boldsymbol{\theta})- \Phi(\boldsymbol{\theta}') \|_2^2,
    \end{equation*}
    \begin{equation*}
         f(\Phi(\boldsymbol{\theta})'; \boldsymbol{\theta}') 
         \geq f(\Phi(\boldsymbol{\theta}); \boldsymbol{\theta}') + 
         \langle \nabla f(\Phi(\boldsymbol{\theta});\boldsymbol{\theta}'), \Phi(\boldsymbol{\theta})- \Phi(\boldsymbol{\theta}') \rangle 
         + \frac{\mu}{2} \| \Phi(\boldsymbol{\theta})- \Phi(\boldsymbol{\theta}' )\|_2^2,
    \end{equation*}
    and thus
    \begin{align}
       &~ \langle \nabla f(\Phi(\boldsymbol{\theta});\boldsymbol{\theta}')-\nabla f(\Phi(\boldsymbol{\theta}');\boldsymbol{\theta}'), \Phi(\boldsymbol{\theta})- \Phi(\boldsymbol{\theta}') \rangle 
        \geq \mu \|\Phi(\boldsymbol{\theta})- \Phi(\boldsymbol{\theta}') \|_2^2 \label{eqn:uniq_eqn1}.
    \end{align} 
    Applying Lemma \ref{lemma:fi_grad_continuity}, we have
    \begin{align}
        \sum_{i=1}^N p_i  \langle \nabla f_i(\Phi(\boldsymbol{\theta}); \boldsymbol{\theta}')-\nabla f_i(\Phi(\boldsymbol{\theta}');\boldsymbol{\theta}'),
        \Phi(\boldsymbol{\theta})- \Phi(\boldsymbol{\theta}') ) \rangle 
        \le  \sum_{i=1}^N p_i  L \epsilon_i \|\boldsymbol{\theta} - \boldsymbol{\theta}'\|_2 \cdot \|\Phi(\boldsymbol{\theta}) - \Phi(\boldsymbol{\theta})\|_2.\label{eqn:uniq_eqn2}
    \end{align}
    Combine \eqref{eqn:uniq_eqn1} and \eqref{eqn:uniq_eqn2}, we have
    \begin{equation}
        \|\Phi(\boldsymbol{\theta}) - \Phi(\boldsymbol{\theta}')\|_2 \leq \frac{\sum_{i=1}^N p_i \epsilon_i L}{\mu} \|\boldsymbol{\theta} - \boldsymbol{\theta}'\|_2 = \frac{\overline{\epsilon} L}{\mu} \|\boldsymbol{\theta} - \boldsymbol{\theta}'\|_2.
    \end{equation}
    Therefore, if $\overline{\epsilon} < \frac{L}{\mu}$, $\Phi(\cdot)$ is a contraction mapping by Banach fixed point theorem and admits a unique fixed point $\boldsymbol{\theta}^{PS}$.

    To show the divergence when $\overline{\epsilon} \geq \frac{L}{\mu}$, we consider the following example where $\theta \in \mathbb{R}, \frac{L}{\mu}=1$, $\overline{\gamma} := \sum_{i=1}^N p_i \gamma_i \neq 0$, and 
    \begin{equation*}
        \ell(\theta; Z) = \frac{1}{2}(\theta - Z)^2, Z \sim \mathcal{D}_i(\theta) = \mathcal{N}(\gamma_i + \epsilon_i \theta, 1)
    \end{equation*}
    we observe 
    \begin{align*}
        f_i(\theta' ; \theta) 
        = & \mathbb{E}_{Z \sim \mathcal{D}_i(\theta)}[\frac{1}{2}(\theta - Z)^2] \\
        = & \mathbb{E}_{\Tilde{Z} \sim \mathcal{N}(0,1)}[\frac{1}{2}(\theta' - \gamma_i - \epsilon_i \theta- \Tilde{Z})^2] \\
        = & \frac{1}{2} (\theta' - \gamma_i - \epsilon_i \theta)^2 + \frac{1}{2},
    \end{align*}
    \begin{equation*}
        \Phi(\theta) = \operatorname{argmin}\limits_{\theta \in \mathbb{R}} \sum_{i=1}^N p_i (\theta' - \gamma_i - \epsilon_i \theta)^2 = \overline{\epsilon} \theta + \overline{\gamma},
    \end{equation*}
     so by applying $\Phi(\cdot)$ $t$ times, we obtain
    \begin{equation*}
        \Phi^t(\theta) = \overline{\epsilon}^t \theta + (1+\overline{\epsilon} + \dots + \overline{\epsilon}^{(t-1)}) \overline{\gamma},
    \end{equation*}
    and since $\overline{\epsilon} \geq \frac{L}{\mu} = 1, \overline{\gamma} \neq 0$, we have $\lim_{t \rightarrow \infty} \|\Phi^t(\theta)\|_2 = \infty$.
\end{proof}

\textbf{Proposition 2.6.} 
Under Assumption \ref{asm:a1_obj_str_conv} and \ref{asm:a3_map_sensitivity}, suppose that the loss $\ell(\boldsymbol{\theta}; Z)$ is $L_z$-Lipschitz in $Z$, let $\overline{\epsilon} := \sum_{i=1}^N p_i \epsilon_i$, we have for every performative stable solution and every performative optimal solution $\boldsymbol{\theta}^{PO}$ that
$$
\|\boldsymbol{\theta}^{PS} -\boldsymbol{\theta}^{PO}\|_2 \leq \frac{2L_z  \overline{\epsilon}}{\mu}.
$$
\begin{proof}
This proof simulates the proof of Theorem 4.3 in \cite{ICML_2020_Perdomo_PP}.

First by the optimality of $\boldsymbol{\theta}^{PO}$, we have $f(\boldsymbol{\theta}^{PO}; \boldsymbol{\theta}^{PO}) \le f(\boldsymbol{\theta}^{PS}; \boldsymbol{\theta}^{PS})$. 
By strong convexity in Assumption \ref{asm:a1_obj_str_conv}, we have
\begin{align*}
    f(\boldsymbol{\theta}^{PO};\boldsymbol{\theta}^{PS}) 
    \geq f(\boldsymbol{\theta}^{PS};\boldsymbol{\theta}^{PS}) + \langle \nabla f(\boldsymbol{\theta}^{PS};\boldsymbol{\theta}^{PS}), ~ \boldsymbol{\theta}^{PO} - \boldsymbol{\theta}^{PS} \rangle + \frac{\mu}{2} \| \boldsymbol{\theta}^{PO} - \boldsymbol{\theta}^{PS} \|_2^2 \geq \frac{\mu}{2} \| \boldsymbol{\theta}^{PO} - \boldsymbol{\theta}^{PS} \|_2^2.
\end{align*}
Further by Assmption \ref{asm:a3_map_sensitivity}, the the loss $\ell(\boldsymbol{\theta}; Z)$ is $L_z$-Lipschitz in $Z$, and Kantorovich-Rubinstein duality, we have
\begin{align}
&~f(\boldsymbol{\theta}^{PO}; \boldsymbol{\theta}^{PS}) - f(\boldsymbol{\theta}^{PO}; \boldsymbol{\theta}^{PO})\nonumber\\
=& ~\sum_{i=1}^N p_i \Big(\mathbb{E}_{Z_i \sim \mathcal{D}_i(\boldsymbol{\theta}^{PS})}[\ell(\boldsymbol{\theta}^{PO};Z_i)] - \mathbb{E}_{Z_i \sim \mathcal{D}_i(\boldsymbol{\theta}^{PO})}[ \ell(\boldsymbol{\theta}^{PO};Z_i)]\Big)\nonumber\\
\le & ~ \sum_{i=1}^N p_i L_z\mathcal{W}_1(\mathcal{D}_i(\boldsymbol{\theta}^{PS}), \mathcal{D}_i(\boldsymbol{\theta}^{PO}) ) \nonumber\\
=& ~\sum_{i=1}^N p_i L_{z} \epsilon_i \| \boldsymbol{\theta}^{PO} - \boldsymbol{\theta}^{PS} \|_2 = L_z \overline{\epsilon} \| \boldsymbol{\theta}^{PO} - \boldsymbol{\theta}^{PS} \|_2. \label{eq:pspo-popo}
\end{align}


where the inequality is a well-know conclusion in optimal tranport theory. Equation \ref{eq:pspo-popo}, we have $L_z \overline{\epsilon} \epsilon_i \| \boldsymbol{\theta}^{PO} - \boldsymbol{\theta}^{PS} \|_2 \geq f(\boldsymbol{\theta}^{PO}, \boldsymbol{\theta}^{PS}) - f(\boldsymbol{\theta}^{PO}; \boldsymbol{\theta}^{PO}) \geq f(\boldsymbol{\theta}^{PO}; \boldsymbol{\theta}^{PS}) - f(\boldsymbol{\theta}^{PS}; \boldsymbol{\theta}^{PS}) \geq \frac{\mu}{2} \| \boldsymbol{\theta}^{PO} - \boldsymbol{\theta}^{PS} \|_2^2$, implying that $\| \boldsymbol{\theta}^{PO} - \boldsymbol{\theta}^{PS} \|_2 \leq \frac{2 L_z \overline{\epsilon}}{\mu}$.

\end{proof}

\section{Proof of Theorem \ref{thm:full}} \label{appensec:full}

\subsection{Additional Notation}
In our analysis, for the sake of convenience, we will define two additional sequences as $\overline{\boldsymbol{w}}^{t}:= \sum_{i=1}^N p_i \boldsymbol{w}_i^t$ and $\overline{\boldsymbol{\theta}}^{t}:= \sum_{i=1}^N p_i \boldsymbol{\theta}_i^t$, following that of \cite{ICLR_2020_Li_Convergence}. We note that $\overline{\boldsymbol{w}}^t$ results from a single step of SGD from $\overline{\boldsymbol{\theta}}^t$. When $t+1\notin \mathcal{I}_E$, both $\overline{\boldsymbol{w}}^{t}$ and $\overline{\boldsymbol{\theta}}^{t}$ are unaccessible. When $t+1\in \mathcal{I}_E$, we can obtain $\overline{\boldsymbol{\theta}}^t$. In addition, we also define $ \overline{\boldsymbol{g}}_t := \sum_{i=1}^N p_i \nabla f_i(\boldsymbol{\theta}_i^t; \boldsymbol{\theta}_i^t),~~ \boldsymbol{g}_t := \sum_{i=1}^N p_i \nabla \ell(\boldsymbol{\theta}_i^t; Z_i^{t+1})$ where $Z_i^{t+1} \sim \mathcal{D}_i(\boldsymbol{\theta}_i^{t})$. It is clear that in full participation, $\overline{\boldsymbol{w}}^{t+1} = \overline{\boldsymbol{\theta}}^{t} - \eta_t \boldsymbol{g}_t$ and $\mathbb{E} \boldsymbol{g}_t = \overline{\boldsymbol{g}}_t$. Clearly we have $\overline{\boldsymbol{\theta
}}^{t} = \overline{\boldsymbol{w}}^{t}$ for any $t$.

\subsection{Key Lemmas}
For clarity, we will present several lemmas for establishing our main theorem. In particular, we will present a descent lemma for $\mathbb{E} [\| \overline{\boldsymbol{w}}^{t} - \boldsymbol{\theta}^{PS} \|^2_2]$ and an upper bound for 
$\sum_{i=1}^N p_i \mathbb{E}\| \boldsymbol{\theta}_i^{t} - \overline{\boldsymbol{\theta}}^{t} \|_2^2 $, which together gives a standard descent lemma for $\mathbb{E} [\| \overline{\boldsymbol{\theta}}^{t+1} - \boldsymbol{\theta}^{PS} \|^2_2]$ in SGD analysis and leads to $\mathcal{O}(\frac{1}{t})$ convergence. The proof of the Lemmas will be deferred. 

In the following lemma, we aim to establish an upper bound for $\mathbb{E} [\| \overline{\boldsymbol{\theta}}^{t+1} - \boldsymbol{\theta}^{PS} \|^2_2]$. Because $\overline{\boldsymbol{\theta}}^{t+1} = \overline{\boldsymbol{w}}^{t+1}$ in full participation, this is equivalent to establishing an upper bound for $\mathbb{E} [\| \overline{\boldsymbol{w}}^{t+1} - \boldsymbol{\theta}^{PS} \|^2_2]$.
\begin{lemma}(Descent Lemma) \label{lem:descent}
    Under Assumptions \ref{asm:a1_obj_str_conv}, \ref{asm:a2_obj_smooth}, \ref{asm:a3_map_sensitivity}, \ref{asm:a4_bound_var_sto_grad}, in full participation
    \begin{align*}\label{eq:onestep}
         \mathbb{E} [\| \overline{\boldsymbol{\theta}}^{t+1} - \boldsymbol{\theta}^{PS} \|^2_2] = \mathbb{E} [\| \overline{\boldsymbol{w}}^{t+1} - \boldsymbol{\theta}^{PS} \|^2_2]  
        \leq  (1 - \Tilde{\mu} \eta_t) \mathbb{E}\|\overline{\boldsymbol{\theta}}^{t} - \boldsymbol{\theta}^{PS}\|_2^2 + 2 \sigma^2 \eta_t^2  + ( c_1 \eta_t + c_2 \eta_t^2 ) \sum_{i=1}^N p_i \mathbb{E}\| \boldsymbol{\theta}_i^t - \overline{\boldsymbol{\theta}}^{t} \|_2^2 
    \end{align*}
    for any $t$, where $\epsilon_{max} := \max_i \epsilon_i, \overline{\epsilon}: = \sum_{i=1}^N p_i \epsilon_i, 
    c_1 := \frac{L (1+\epsilon_{max})^2}{2 \delta \overline{\epsilon}},
    c_2 := 4 [\sigma^2 + L^2(1+\epsilon_{max})^2], 
    \Tilde{\mu} := \mu - (1+\delta) \overline{\epsilon} L$.
\end{lemma}

Now we are going to establish an upper bound for $\sum_{i=1}^N p_i \mathbb{E}\| \boldsymbol{\theta}_i^t - \overline{\boldsymbol{\theta}}^{t} \|_2^2 $. Note that if $t \in \mathcal{I}_E$, the synchronization step, we have $\boldsymbol{\theta}_i^t = \overline{\boldsymbol{\theta}}^{t}$ for any $i\in [N]$, which implies that $\sum_{i=1}^N p_i \mathbb{E}\| \boldsymbol{\theta}_i^t - \overline{\boldsymbol{\theta}}^{t} \|_2^2 =0$. If $t \notin \mathcal{I}_E$, the following lemma gives an upper bound for $\sum_{i=1}^N p_i \mathbb{E}\| \boldsymbol{\theta}_i^t - \overline{\boldsymbol{\theta}}^{t} \|_2^2 $. 

\begin{lemma}(Consensus Error) \label{lem:consensuserr}
    Under Assumption \ref{asm:a1_obj_str_conv}, \ref{asm:a2_obj_smooth}, \ref{asm:a3_map_sensitivity}, \ref{asm:a4_bound_var_sto_grad}, \ref{asm:a5_bound_var_local_grad}, if $\{\eta_t\}$ is non-increasing, $\eta_{t}\le 2\eta_{t+E}$, $t \notin  \mathcal{I}_E$, $ \eta_t^2 \le 1/\big(2c_3 (t+1-t_0)(1+2(t+1-t_0))\big) $, and
    \begin{align*}
        \eta_0\le \hat{\eta}_0:= \frac{\Tilde{\mu} \mathbb{E}\|\overline{\boldsymbol{\theta}}^0-\boldsymbol{\theta}^{PS}\|_2^2}{2\sigma^2+( c_1c_3  + c_2/6 ) (2E^2-E)\log E\big( (16\sigma^2+12\varsigma^2)\mathbb{E}\|\overline{\boldsymbol{\theta}}^0-\boldsymbol{\theta}^{PS}\|_2^2+ (8\sigma^2+12\varsigma^2)\big) },
    \end{align*}
    then in full participation, we have
    \begin{align*}
    \sum_{i=1}^N p_i\mathbb{E}\|\boldsymbol{\theta}^{t}_i- \overline{\boldsymbol{\theta}}^{t}\|^2_2 
    \le 4\eta_{t}^2 (2E^2-E)\log E (48\sigma^2+36\varsigma^2)\mathbb{E}\|\overline{\boldsymbol{\theta}}^0-\boldsymbol{\theta}^{PS}\|_2^2 + 4\eta_{t}^2 (2E^2-E)\log E (24\sigma^2+36\varsigma^2).
\end{align*}
where for any $t$, where $\epsilon_{max} := \max_i \epsilon_i, \overline{\epsilon}: = \sum_{i=1}^N p_i \epsilon_i, 
    c_1 := \frac{L (1+\epsilon_{max})^2}{2 \delta \overline{\epsilon}},
    c_2 := 4 [\sigma^2 + L^2(1+\epsilon_{max})^2], 
    \Tilde{\mu} := \mu - (1+\delta) \overline{\epsilon} L$, $c_3:= 12\sigma^2+18 L^2(1+\epsilon_{\max})^2$.
    
(One should note that $4\eta_t^2$, $(48\sigma^2+36\varsigma^2)$, and $(24\sigma^2+36\varsigma^2)$ comes from several times of applying $\eta_{t-1}\le 2\eta_t$ and the real constants could be much smaller by choosing stepsizes carefully.)
\end{lemma}

The following lemma gives us a standard descent lemma in SGD analysis under technical conditions for establishing the $\mathcal{O}(\frac{1}{T})$ convergence in Theorem \ref{thm:full}.
\begin{lemma}\label{lem:cleandescent}
Under Assumption \ref{asm:a1_obj_str_conv}, \ref{asm:a2_obj_smooth}, \ref{asm:a3_map_sensitivity}, \ref{asm:a4_bound_var_sto_grad}, \ref{asm:a5_bound_var_local_grad}, if $\{\eta_t\}$ is non-increasing, $\eta_{t}\le 2\eta_{t+E}$, $ \eta_t^2 \le 1/\big(2c_3 (t+1-t_0)(1+2(t+1-t_0))\big) $, and
    \begin{align*}
        \eta_0\le\hat{\eta}_0:= \frac{\Tilde{\mu} \mathbb{E}\|\overline{\boldsymbol{\theta}}^0-\boldsymbol{\theta}^{PS}\|_2^2}{2\sigma^2+( c_1c_3  + c_2/6 ) (2E^2-E)\log E\big( (16\sigma^2+12\varsigma^2)\mathbb{E}\|\overline{\boldsymbol{\theta}}^0-\boldsymbol{\theta}^{PS}\|_2^2+ (8\sigma^2+12\varsigma^2)\big) },
    \end{align*}
    then in full participation, we have
    \begin{align*}
        \mathbb{E} [\| \overline{\boldsymbol{\theta}}^{t+1} - \boldsymbol{\theta}^{PS} \|^2_2]  
        \leq & (1 - \Tilde{\mu} \eta_t) \mathbb{E}\|\overline{\boldsymbol{\theta}}^{t} - \boldsymbol{\theta}^{PS}\|_2^2 + B \eta_t^2 
    \end{align*}
where for any $t$, where $\epsilon_{max} := \max_i \epsilon_i, \overline{\epsilon}: = \sum_{i=1}^N p_i \epsilon_i, 
    c_1 := \frac{L (1+\epsilon_{max})^2}{2 \delta \overline{\epsilon}},
    c_2 := 4 [\sigma^2 + L^2(1+\epsilon_{max})^2], 
    \Tilde{\mu} := \mu - (1+\delta) \overline{\epsilon} L$, $c_3:= 12\sigma^2+18 L^2(1+\epsilon_{\max})^2$, and
        $B:= 2\sigma^2+ ( 4 c_1 \hat{\eta}_0 + 4 c_2 \hat{\eta}_0^2 )(2E^2-E)\log E \big((48\sigma^2+36\varsigma^2)\mathbb{E}\|\overline{\boldsymbol{\theta}}^0-\boldsymbol{\theta}^{PS}\|_2^2 +(24\sigma^2+36\varsigma^2)\big)$.
\end{lemma}
\subsection{Completing the Proof of Theorem \ref{thm:full}}\label{sec:proofoftheorem3.1}
We restate the definitions of all the constants here:

{\bf Constants independent of system design.} \\
$\epsilon_{max} :=  \max_i \epsilon_i$,

    $\overline{\epsilon} := \sum_{i=1}^N p_i \epsilon_i$,
    
    $\Tilde{\mu} :=  \mu - (1+\delta) \overline{\epsilon} L$,
    
    $c_1 :=  \big(L (1+\epsilon_{max})^2\big)/(2 \delta \overline{\epsilon})$,
    
    $c_2 := 4 \big[\sigma^2 + L^2(1+\epsilon_{max})^2\big]$, 
    
    $c_3:= 6\big[2\sigma^2+ 3 L^2(1+\epsilon_{\max})^2\big]$,
    
    $c_4:= 16\sigma^2+12\varsigma^2+ (8\sigma^2+12\varsigma^2)/\mathbb{E}\|\overline{\boldsymbol{\theta}}^0-\boldsymbol{\theta}^{PS}\|_2^2$,
    
    $c_5:=  (48\sigma^2+36\varsigma^2)\mathbb{E}\|\overline{\boldsymbol{\theta}}^0-\boldsymbol{\theta}^{PS}\|_2^2 +(24\sigma^2+36\varsigma^2)$.

{\bf Constants related to system design (e.g., $E, K$).} \\
    $\hat{\eta}_0 :=\Tilde{\mu}/ \big(2\sigma^2+( c_1c_3  + c_2/6 ) c_4(2E^2-E)\log E\big)$,
    
    $B :=2\sigma^2+ ( 4 c_1 \hat{\eta}_0 + 4 c_2 \hat{\eta}_0^2 )c_5(2E^2-E)\log E$ ,

    $c_6 := (2E^2+3E+1)\log (E+1)$,

    $\tilde{\eta}_0:=\Tilde{\mu} / \big(2\sigma^2+( c_1c_3  + c_2/6 ) c_4c_6\big)$,
    
    $B_1 := 2 \sigma^2  + (4c_1 \tilde{\eta}_0 + 4c_2 \tilde{\eta}_0^2 +1/K) c_5 c_6 $,
    
    $B_2 := 2 \sigma^2 + \big(4c_1 \tilde{\eta}_0 + 4c_2 \tilde{\eta}_0^2 + \frac{N-K}{KN(N-1)}\big )c_5c_6$.

Instead of proving Theorem \ref{thm:full} directly, we prove a more general version of convergence results suppose that some conditions about the stepsize are satisfied. Then we will show that the stepsizes given in Theorem \ref{thm:full} satisfy the conditions. 
\begin{theorem}
Under Assumption \ref{asm:a1_obj_str_conv}, \ref{asm:a2_obj_smooth}, \ref{asm:a3_map_sensitivity}, \ref{asm:a4_bound_var_sto_grad}, \ref{asm:a5_bound_var_local_grad}, for a diminishing stepsize $\eta_t = \frac{\beta}{t+\gamma}$ where  $\beta > \frac{1}{\Tilde{\mu}}$, $\gamma>0$ such that $\eta_0\le \hat{\eta}_0$, $\eta_{t}\le 2\eta_{t+E}$, and $ \eta_t^2 \le 1/\big(2c_3 (t+1-t_0)(1+2(t+1-t_0))\big) $,
    then in full participation, we have for any $t$
    \begin{align*}
        \mathbb{E} [\| \overline{\boldsymbol{\theta}}^{t} - \boldsymbol{\theta}^{PS} \|^2_2]  
        \leq & \frac{\upsilon}{\gamma+t}
    \end{align*}
    where $\upsilon=\max \left\{\frac{4 B}{ \Tilde{\mu}^2},\gamma \mathbb{E} [\| \overline{\boldsymbol{\theta}}^{0} - \boldsymbol{\theta}^{PS} \|^2_2]\right\}$.
\end{theorem}
\begin{proof}
    Let $\Delta_t:=\mathbb{E} [\| \overline{\boldsymbol{\theta}}^{t} - \boldsymbol{\theta}^{PS} \|^2_2]$, then from Lemma \ref{lem:cleandescent}, we have
    \begin{align*}
        \Delta_{t+1}\le (1 - \Tilde{\mu} \eta_t) \mathbb{E}\|\overline{\boldsymbol{\theta}}^{t} - \boldsymbol{\theta}^{PS}\|_2^2 + B \eta_t^2 .
    \end{align*}
    For a diminishing stepsize $\eta_t = \frac{\beta}{t+\gamma}$ where  $\beta > \frac{1}{\Tilde{\mu}}$, $\gamma>0$ such that $ \eta_t^2 \le 1/\big(2c_3 (t+1-t_0)(1+2(t+1-t_0))\big) $, $\eta_0 \le \hat{\eta}_0$, and $\eta_t\le 2\eta_{t+E}$, we will prove that $\Delta_t\le \frac{\upsilon}{\gamma+t}$ where $\upsilon=\max \left\{\frac{\beta^2 B}{\beta \Tilde{\mu}-1},\gamma \Delta_0\right\}=\max \left\{\frac{4 B}{ \Tilde{\mu}^2},\gamma \Delta_0\right\}$ by induction.

    Firstly, $\Delta_0 \le \frac{\upsilon}{\gamma} $ by the definition of $\upsilon$. Assume that for some $0\le t$, $\Delta_t \le \frac{\upsilon}{\gamma+t} $, then
    \begin{align*}
\Delta_{t+1} \leq &~ \left(1-\eta_t \Tilde{\mu}\right) \Delta_t+\eta_t^2 B \\
\leq& ~\left(1-\frac{\beta \Tilde{\mu}}{t+\gamma}\right) \frac{v}{t+\gamma}+\frac{\beta^2 B}{(t+\gamma)^2} \\
=&~ \frac{t+\gamma-1}{(t+\gamma)^2} v+\left[\frac{\beta^2 B}{(t+\gamma)^2}-\frac{\beta \Tilde{\mu}-1}{(t+\gamma)^2} v\right] \\
\leq& ~ \frac{v}{t+\gamma+1} .
    \end{align*}
    Specifically, if we choose $\beta = \frac{2}{\Tilde{\mu}}$, $\gamma = \max\{ \frac{2}{\Tilde{\mu} \hat{\eta}_0}, E,\frac{2}{\Tilde{\mu}} \sqrt{2E(2E+1)(12\sigma^2+18 L^2(1+\epsilon_{\max})^2)}\}$, then we have 
    \begin{align*}
        \eta_0  = \frac{\beta}{\gamma}  \le \frac{2}{\Tilde{\mu} \frac{2}{\Tilde{\mu} \hat{\eta}_0}} = \hat{\eta}_0
        \end{align*}
    and 
    \begin{align*}
        \eta_t - 2\eta_{t+E} = \frac{\beta}{\gamma+t} - \frac{2\beta}{\gamma+t+E} = \frac{\beta (E-\gamma-t)}{(\gamma+t)(\gamma+t+E)}\le \frac{\beta (E-\gamma)} {(\gamma+t)(\gamma+t+E)} \le 0. 
    \end{align*}
    To prove that $ \eta_t^2 \le 1/\big(2c_3 (t+1-t_0)(1+2(t+1-t_0))\big)$ for any $t$, it suffices to prove that for $0\le t\le E-1$ because $\{\eta_t\}$, i.e., $t_0=0$, is non-increasing and $t+1-t_0$ is periodic with period $E$. When $t_0=0$, we need to prove $ \eta_t^2 \le 1/\big(2c_3 (t+1)(1+2(t+1))\big)$ for $0\le t\le E-1$, which is satisfied if
    \begin{align*}
        &~ \max_{0\le t\le E-1} \eta_t \le \min_{0\le t\le E-1} \sqrt{1/\big(2c_3 (t+1)(1+2(t+1))\big)}\\
       \Longleftrightarrow &~ \eta_0 \le \sqrt{\frac{1}{2E(2E+1)c_3}} \\
       \Longleftrightarrow &~ \gamma \ge \beta\sqrt{2E(2E+1)c_3} = \frac{2}{\Tilde{\mu}} \sqrt{2E(2E+1)c_3}= \frac{2}{\Tilde{\mu}} \sqrt{2E(2E+1)(12\sigma^2+18 L^2(1+\epsilon_{\max})^2)}.
    \end{align*}
\end{proof}

\subsection{Deferred Proofs of Key Lemmas}
\begin{proof}[Proof of Lemma \ref{lem:descent}]
This proof follows from Lemma 3 in \cite{ArXiv_2022_Li_MPP}. We first decompose $\| \overline{\boldsymbol{w}}^{t+1} - \boldsymbol{\theta}^{PS} \|^2_2$ as
\begin{equation} \label{eqn:1step_bound}
       \mathbb{E} \| \overline{\boldsymbol{w}}^{t+1} - \boldsymbol{\theta}^{PS} \|^2_2 = \mathbb{E}\| \overline{\boldsymbol{\theta}}^{t} - \eta_{t} \boldsymbol{g}_t - \boldsymbol{\theta}^{PS} \|^2_2=  \mathbb{E}\| \overline{\boldsymbol{\theta}}^{t} - \boldsymbol{\theta}^{PS} \|^2_2 - 2 \eta_t   \mathbb{E}\langle \overline{\boldsymbol{\theta}}^{t} - \boldsymbol{\theta}^{PS} ,~ \boldsymbol{g}_t \rangle  + \eta_t^2  \mathbb{E}\| \boldsymbol{g}_t\|_2^2 .
    \end{equation}

Next we present an upper bound for $\mathbb{E}\| \boldsymbol{g}_t\|_2^2 $. By the definition of $\boldsymbol{\theta}^{PS}$, we have $\sum_{i=1}^N p_i \nabla f_i(\boldsymbol{\theta}^{PS}; \boldsymbol{\theta}^{PS}) = \boldsymbol{0}$, and thus
    \begin{align*} 
        \mathbb{E}\| \boldsymbol{g}_t\|_2^2 
        = &~ \mathbb{E}\| \sum_{i=1}^N p_i [\nabla \ell(\boldsymbol{\theta}_i^t; Z_i^{t+1}) - \nabla f_i(\boldsymbol{\theta}_i^t; \boldsymbol{\theta}_i^t) 
        + \nabla f_i(\boldsymbol{\theta}_i^t; \boldsymbol{\theta}_i^t) - \nabla f_i(\boldsymbol{\theta}^{PS}; \boldsymbol{\theta}^{PS}) ]\|_2^2  \nonumber \\
        \leq &~ 2  \mathbb{E}\|  \sum_{i=1}^N \nabla \ell(\boldsymbol{\theta}_i^t; Z_i^{t+1}) 
        - \nabla f_i(\boldsymbol{\theta}_i^t; \boldsymbol{\theta}_i^t) \|_2^2
        + 2 \mathbb{E}\| \sum_{i=1}^N p_i [\nabla f_i(\boldsymbol{\theta}_i^t; \boldsymbol{\theta}_i^t) - \nabla f_i(\boldsymbol{\theta}^{PS}; \boldsymbol{\theta}^{PS}) ]\|_2^2\\
        \le &~2 \sum_{i=1}^N p_i \mathbb{E} [\| \nabla \ell(\boldsymbol{\theta}_i^t; Z_i^{t+1}) 
        - \nabla f_i(\boldsymbol{\theta}_i^t; \boldsymbol{\theta}_i^t) \|_2^2 ] \nonumber + 2 \sum_{i=1}^N p_i \mathbb{E} [\| \nabla f_i(\boldsymbol{\theta}_i^t; \boldsymbol{\theta}_i^t) - \nabla f_i(\boldsymbol{\theta}^{PS}; \boldsymbol{\theta}^{PS}) \|_2^2 ] \nonumber \\
        \le &~ 2 \sum_{i=1}^N p_i \sigma^2 \left(1 + \mathbb{E}\|\boldsymbol{\theta}_i^t - \boldsymbol{\theta}^{PS}\|_2^2 \right) 
        +  2 \sum_{i=1}^N p_i L^2 (1 + \epsilon_i)^2  \mathbb{E}\|\boldsymbol{\theta}_i^t - \boldsymbol{\theta}^{PS}\|_2^2
    \end{align*}
    where the second inequality is due to the convexity of 2-norm and the last inequality is due to Assumption \ref{asm:a4_bound_var_sto_grad} and Lemma \ref{lemma:fi_grad_continuity}. Since $\|\boldsymbol{\theta}_i^t - \boldsymbol{\theta}^{PS}\|_2^2 \leq 2\|\boldsymbol{\theta}_i^t - \overline{\boldsymbol{\theta}}^{t}\|_2^2 + 2 \|\overline{\boldsymbol{\theta}}^{t}- \boldsymbol{\theta}^{PS}\|_2^2$ and $\epsilon_i \le \epsilon_{max}$, we have
    \begin{align} 
        \mathbb{E} [ \| \boldsymbol{g}_t\|_2^2 ] 
        \leq  &~ 2 \sigma^2 + 4[\sigma^2 + L^2(1+\epsilon_{max})^2] \mathbb{E}\| \overline{\boldsymbol{\theta}}^{t} - \boldsymbol{\theta}^{PS} \|^2_2  + 4 [\sigma^2 + L^2(1+\epsilon_{max})^2] ~ \sum_{i=1}^N p_i  \mathbb{E}\|\boldsymbol{\theta}_i^t - \overline{\boldsymbol{\theta}}^{t} \|_2^2\nonumber\\
        =&~  2\sigma^2 + c_2 \mathbb{E}\|\overline{\boldsymbol{\theta}}^{t} - \boldsymbol{\theta}^{PS}\|_2^2 + c_2 \sum_{i=1}^N p_i \mathbb{E}\| \boldsymbol{\theta}_i^t - \overline{\boldsymbol{\theta}}^{t} \|_2^2. \label{eqn:1step_bound_p2}
    \end{align}
    Next, we focus on establishing a lower bound for $\mathbb{E} [ \langle \overline{\boldsymbol{\theta}}^{t} - \boldsymbol{\theta}^{PS} ,~ \boldsymbol{g}_t \rangle ]$.  By the law of total expectation and $\sum_{i=1}^N p_i \nabla f_i(\boldsymbol{\theta}^{PS}; \boldsymbol{\theta}^{PS}) = \boldsymbol{0}$, we have
    \begin{align*}
        \mathbb{E}  \langle \overline{\boldsymbol{\theta}}^{t} - \boldsymbol{\theta}^{PS} ,~ \boldsymbol{g}_t \rangle 
        = &~ \mathbb{E} \big[ \mathbb{E}_t\langle \overline{\boldsymbol{\theta}}^{t} - \boldsymbol{\theta}^{PS} ,~ \boldsymbol{g}_t \rangle \big]\\
    = &~ \mathbb{E} [ \langle \overline{\boldsymbol{\theta}}^{t} - \boldsymbol{\theta}^{PS} ,~ \overline{\boldsymbol{g}}_t \rangle ]\\
       = &~ \mathbb{E} \bigg[ \underbrace{\sum_{i=1}^N p_i \big\langle \overline{\boldsymbol{\theta}}^{t} - \boldsymbol{\theta}^{PS},~
        \nabla f_i(\boldsymbol{\theta}_i^t; \boldsymbol{\theta}_i^t) 
        - \nabla f_i(\overline{\boldsymbol{\theta}}^{t} ; \boldsymbol{\theta}^{PS}) \big\rangle}_{A} \nonumber \\
        & + \underbrace{\sum_{i=1}^N p_i \big\langle \overline{\boldsymbol{\theta}}^{t} - \boldsymbol{\theta}^{PS},~
        \nabla f_i(\overline{\boldsymbol{\theta}}^{t} ; \boldsymbol{\theta}^{PS})
        - \nabla f_i(\boldsymbol{\theta}^{PS} ; \boldsymbol{\theta}^{PS})  \big\rangle}_{B} \bigg].
    \end{align*}
    On the one hand, applying Cauchy-Schwarz inequality and Lemma \ref{lemma:fi_grad_continuity}, we have
    \begin{align*}
        A \geq &~ - \|\overline{\boldsymbol{\theta}}^{t} - \boldsymbol{\theta}^{PS}\|_2 \sum_{i=1}^N p_i \big( L \| \boldsymbol{\theta}_i^t - \overline{\boldsymbol{\theta}}^{t} \|_2 + L \epsilon_i \|\boldsymbol{\theta}_i^t - \boldsymbol{\theta}^{PS}\|_2 \big) \nonumber \\
        \geq &~ - \|\overline{\boldsymbol{\theta}}^{t} - \boldsymbol{\theta}^{PS}\|_2 \sum_{i=1}^N p_i \big( L(1+\epsilon_i) \| \boldsymbol{\theta}_i^t - \overline{\boldsymbol{\theta}}^{t} \|_2 + L \epsilon_i \|\overline{\boldsymbol{\theta}}^{t} - \boldsymbol{\theta}^{PS}\|_2 \big)\\
    \ge &~- L\overline{\epsilon}\|\overline{\boldsymbol{\theta}}^{t} - \boldsymbol{\theta}^{PS}\|_2^2-L(1 + \epsilon_{max}) \sum_{i=1}^N p_i 
        \|\overline{\boldsymbol{\theta}}^{t} - \boldsymbol{\theta}^{PS}\|_2 
        \| \boldsymbol{\theta}_i^t - \overline{\boldsymbol{\theta}}^{t} \|_2. 
    \end{align*}
    On the other hand, with the strong convexity in Assumption \ref{asm:a1_obj_str_conv}, we have $B \geq \mu \|\overline{\boldsymbol{\theta}}^{t} - \boldsymbol{\theta}^{PS}\|_2^2$.
    Therefore, for any $\alpha > 0$, using the lower bounds on $A, B$, and the Young's inequality shows that
    \begin{align} 
        &~ \mathbb{E} [\langle \overline{\boldsymbol{\theta}}^{t} - \boldsymbol{\theta}^{PS} ,~ \boldsymbol{g}_t  \rangle ]  \nonumber \\
        \geq &~ (\mu - L \overline{\epsilon}) \mathbb{E}\|\overline{\boldsymbol{\theta}}^{t} - \boldsymbol{\theta}^{PS}\|_2^2 
        - L(1 + \epsilon_{max}) \sum_{i=1}^N p_i 
        \mathbb{E}\big[\|\overline{\boldsymbol{\theta}}^{t} - \boldsymbol{\theta}^{PS}\|_2 
        \| \boldsymbol{\theta}_i^t - \overline{\boldsymbol{\theta}}^{t} \|_2\big] \nonumber \\
        \geq &~ \big( \mu - L \overline{\epsilon} - \frac{\alpha}{2} L (1 + \epsilon_{max}) \big)\mathbb{E}\|\overline{\boldsymbol{\theta}}^{t} - \boldsymbol{\theta}^{PS}\|_2^2 
        - \frac{L(1 + \epsilon_{max})}{2 \alpha} \sum_{i=1}^N p_i \mathbb{E}\| \boldsymbol{\theta}_i^t - \overline{\boldsymbol{\theta}}^{t} \|_2^2  \nonumber \\
        \geq &~ (\mu - (1 + \delta) L \overline{\epsilon})\mathbb{E} \|\overline{\boldsymbol{\theta}}^{t} - \boldsymbol{\theta}^{PS}\|_2^2 
        - \frac{L (1 + \epsilon_{max})^2}{4 \delta \overline{\epsilon}} \sum_{i=1}^N p_i \mathbb{E}\| \boldsymbol{\theta}_i^t - \overline{\boldsymbol{\theta}}^{t} \|_2^2 \label{eqn:1step_bound_p3}
    \end{align}
    where we have set $\alpha := \frac{2 \delta \overline{\epsilon}}{1 + \epsilon_{max}}$ in the last line.

    Recall that we denote 
    \begin{equation*}
        c_1 := \frac{L (1+\epsilon_{max})^2}{2 \delta \overline{\epsilon}},~~
        c_2 := 4 [\sigma^2 + L^2(1+\epsilon_{max})^2], ~~
        \Tilde{\mu} := \mu - (1+\delta) \overline{\epsilon} L.
    \end{equation*}
    Combining \eqref{eqn:1step_bound}, \eqref{eqn:1step_bound_p2}, \eqref{eqn:1step_bound_p3}, we have
    \begin{align*}
        &~ \mathbb{E} \big[\| \overline{\boldsymbol{w}}^{t+1} - \boldsymbol{\theta}^{PS} \|^2_2\big] \nonumber \\
        \leq &~ \mathbb{E}\| \overline{\boldsymbol{\theta}}^{t} - \boldsymbol{\theta}^{PS} \|^2_2 \nonumber \\
        & - 2 \eta_t \bigg[(\mu - (1 + \delta) L \overline{\epsilon}) \mathbb{E}\|\overline{\boldsymbol{\theta}}^{t} - \boldsymbol{\theta}^{PS}\|_2^2 
        - \frac{L (1 + \epsilon_{max})}{4 \delta \overline{\epsilon}} \sum_{i=1}^N p_i \mathbb{E}\| \boldsymbol{\theta}_i^t - \overline{\boldsymbol{\theta}}^{t} \|_2^2 \bigg] \nonumber \\
        & + \eta_t^2 \bigg[ 2\sigma^2 + c_2 \mathbb{E}\|\overline{\boldsymbol{\theta}}^{t} - \boldsymbol{\theta}^{PS}\|_2^2 + c_2 \sum_{i=1}^N p_i \mathbb{E}\| \boldsymbol{\theta}_i^t - \overline{\boldsymbol{\theta}}^{t} \|_2^2 \bigg] \nonumber \\
        = &~ (1 - 2 \Tilde{\mu} \eta_t + c_2 \eta_t^2) \mathbb{E}\|\overline{\boldsymbol{\theta}}^{t} - \boldsymbol{\theta}^{PS}\|_2^2 + ( c_1 \eta_t + c_2 \eta_t^2 ) \sum_{i=1}^N p_i \mathbb{E}\| \boldsymbol{\theta}_i^t - \overline{\boldsymbol{\theta}}^{t} \|_2^2 + 2 \sigma^2 \eta_t^2 \nonumber \\
        \leq &~ (1 - \Tilde{\mu} \eta_t) \mathbb{E}\|\overline{\boldsymbol{\theta}}^{t} - \boldsymbol{\theta}^{PS}\|_2^2 + ( c_1 \eta_t + c_2 \eta_t^2 ) \sum_{i=1}^N p_i \mathbb{E}\| \boldsymbol{\theta}_i^t - \overline{\boldsymbol{\theta}}^{t} \|_2^2 + 2 \sigma^2 \eta_t^2
    \end{align*}
    where the last inequality is obtained by observing the condition $\eta_t \leq \Tilde{\mu}/c_2$.
\end{proof}

\begin{proof}[Proof of Lemma \ref{lem:consensuserr}]
In this proof, for convenience, we will discuss with respect to $t+1$ where we assume $t+1\notin \mathcal{I}_E$ and transfer back to $t$ in the last. First by the update rule, we have
    \begin{align*}
        \boldsymbol{\theta}^{t+1}_i- \overline{\boldsymbol{\theta}}^{t+1} = \boldsymbol{\theta}_i^{t} 
     - \overline{\boldsymbol{\theta}}^{t} - \eta_{t} (\nabla \ell(\boldsymbol{\theta}_i^t; Z_i^{t+1})
    -{g}_t).
    \end{align*}
    Using Young's inequality, we have
    \begin{align}
        \sum_{i=1}^N p_i\mathbb{E}\|\boldsymbol{\theta}^{t+1}_i- \overline{\boldsymbol{\theta}}^{t+1}\|^2_2=&~ \sum_{i=1}^N p_i\mathbb{E}\|\boldsymbol{\theta}_i^{t} 
     - \overline{\boldsymbol{\theta}}^{t} - \eta_{t} (\nabla \ell(\boldsymbol{\theta}_i^t; Z_i^{t+1})
    - {g}_t)\|^2_2\nonumber\\
    \le &~ (1+\alpha_t) \sum_{i=1}^N p_i\mathbb{E}\|\boldsymbol{\theta}_i^{t} 
     - \overline{\boldsymbol{\theta}}^{t}\|_2^2 + \eta_{t}^2(1+\alpha_t^{-1})\underbrace{\sum_{i=1}^N p_i\mathbb{E}\| \nabla \ell(\boldsymbol{\theta}_i^t; Z_i^{t+1})
    - {g}_t\|^2_2}_{B} \label{eqn:consensuserr_1}
    \end{align}
    where $\alpha_t>0$ is a free chosen parameter. Next, we are going to establish an upper bound for $B$. Notice that
\begin{align*}
B=&~ \sum_{i=1}^N p_i\mathbb{E}\| \nabla \ell(\boldsymbol{\theta}_i^t; Z_i^{t+1})
    - {g}_t\|^2_2\\
    =&~ \mathbb{E} \bigg[\sum_{i=1}^N p_i \| \nabla \ell(\boldsymbol{\theta}_i^{t}; Z_i^{t+1})- \sum_{j=1}^N p_j  \nabla \ell(\boldsymbol{\theta}_j^{t}; Z_j^{t+1})  \|_2^2\bigg]\\
    = &~ \mathbb{E} \bigg[\sum_{i=1}^N p_i \| \nabla \ell(\boldsymbol{\theta}_i^{t}; Z_i^{t+1}) -\nabla f_i\left(\boldsymbol{\theta}_i^t, \boldsymbol{\theta}_i^t\right)+ \nabla f_i\left(\boldsymbol{\theta}_i^t, \boldsymbol{\theta}_i^t\right)- \sum_{j=1}^N p_j  \nabla f_j(\boldsymbol{\theta}_j^{t}; \boldsymbol{\theta}_j^{t}) \\
    &~+ \sum_{j=1}^N p_j  \nabla f_j(\boldsymbol{\theta}_j^{t}; \boldsymbol{\theta}_j^{t}) -\sum_{j=1}^N p_j  \nabla \ell(\boldsymbol{\theta}_j^{t}; Z_j^{t+1}) \|_2^2\bigg]\\
    \le &~ 3 \mathbb{E} \bigg[\sum_{i=1}^N p_i \|\nabla \ell\left(\boldsymbol{\theta}_i^t ; Z_i^{t+1}\right)-\nabla f_i\left(\boldsymbol{\theta}_i^t, \boldsymbol{\theta}_i^t\right)\|^2_2\bigg] +3 \mathbb{E} \bigg[\sum_{i=1}^N p_i\|\nabla f_i\left(\boldsymbol{\theta}_i^t, \boldsymbol{\theta}_i^t\right)- \sum_{j=1}^N p_j\nabla f_j\left(\boldsymbol{\theta}_j^t, \boldsymbol{\theta}_j^t\right)\|^2_2\bigg]\\
    &~ +3 \mathbb{E} \bigg[\sum_{i=1}^N p_i \|\sum_{j=1}^N p_j  \nabla f_j(\boldsymbol{\theta}_j^{t}; \boldsymbol{\theta}_j^{t}) -\sum_{j=1}^N p_j  \nabla \ell(\boldsymbol{\theta}_j^{t}; Z_j^{t+1})\|^2_2 \bigg]\\
    \le &~ 3 \mathbb{E} \bigg[\sum_{i=1}^N p_i \|\nabla \ell\left(\boldsymbol{\theta}_i^t ; Z_i^{t+1}\right)-\nabla f_i\left(\boldsymbol{\theta}_i^t, \boldsymbol{\theta}_i^t\right)\|^2_2\bigg] +3 \mathbb{E} \bigg[\sum_{i=1}^N p_i\|\nabla f_i\left(\boldsymbol{\theta}_i^t, \boldsymbol{\theta}_i^t\right)- \sum_{j=1}^N p_j\nabla f_j\left(\boldsymbol{\theta}_j^t, \boldsymbol{\theta}_j^t\right)\|^2_2\bigg]\\
    &~ +3 \mathbb{E} \bigg[ \sum_{i=1}^N p_i\sum_{j=1}^N p_j\|  \nabla f_j(\boldsymbol{\theta}_j^{t}; \boldsymbol{\theta}_j^{t}) - \nabla \ell(\boldsymbol{\theta}_j^{t}; Z_j^{t+1})\|^2_2 \bigg]\\
    = &~ 3 \mathbb{E} \bigg[\sum_{i=1}^N p_i \|\nabla \ell\left(\boldsymbol{\theta}_i^t ; Z_i^{t+1}\right)-\nabla f_i\left(\boldsymbol{\theta}_i^t, \boldsymbol{\theta}_i^t\right)\|^2_2\bigg] +3 \mathbb{E} \bigg[\sum_{i=1}^N p_i\|\nabla f_i\left(\boldsymbol{\theta}_i^t, \boldsymbol{\theta}_i^t\right)- \sum_{j=1}^N p_j\nabla f_j\left(\boldsymbol{\theta}_j^t, \boldsymbol{\theta}_j^t\right)\|^2_2\bigg]\\
    &~ +3 \mathbb{E} \bigg[ \sum_{j=1}^N p_j\|  \nabla f_j(\boldsymbol{\theta}_j^{t}; \boldsymbol{\theta}_j^{t}) - \nabla \ell(\boldsymbol{\theta}_j^{t}; Z_j^{t+1})\|^2_2 \bigg]\\
    \le &~ 6 \sigma^2 \bigg( 1+ \mathbb{E} \bigg[\sum_{i=1}^N p_i \|\boldsymbol{\theta}^t_i - \boldsymbol{\theta}^{PS} \|_2^2\bigg]\bigg)\\
    &~+ 3 \mathbb{E} \bigg[\sum_{i=1}^N p_i\|\nabla f_i\left(\boldsymbol{\theta}_i^t, \boldsymbol{\theta}_i^t\right)- \sum_{j=1}^N p_j\nabla f_j\left(\boldsymbol{\theta}_j^t, \boldsymbol{\theta}_j^t\right)\|^2_2\bigg]
\end{align*}
where the last inequality is by Assumption \ref{asm:a4_bound_var_sto_grad}. On the other hand, we have
\begin{align*}
    &~3 \mathbb{E} \bigg[\sum_{i=1}^N p_i\big\|\nabla f_i(\boldsymbol{\theta}_i^t, \boldsymbol{\theta}_i^t)- \sum_{j=1}^N p_j\nabla f_j(\boldsymbol{\theta}_j^t, \boldsymbol{\theta}_j^t)\big\|^2_2\bigg]\\
    =&~ 3 \sum_{i=1}^N p_i\mathbb{E} \big\|\nabla f_i(\boldsymbol{\theta}_i^t, \boldsymbol{\theta}_i^t)-\nabla f_i(\overline{\boldsymbol{\theta}}^t, \overline{\boldsymbol{\theta}}^t)+\nabla f_i(\overline{\boldsymbol{\theta}}^t, \overline{\boldsymbol{\theta}}^t)- \sum_{j=1}^N p_j\nabla f_j(\overline{\boldsymbol{\theta}}^t, \overline{\boldsymbol{\theta}}^t)- \sum_{j=1}^N p_j\big(\nabla f_j(\boldsymbol{\theta}_j^t, \boldsymbol{\theta}_j^t)-\nabla f_j(\overline{\boldsymbol{\theta}}^t, \overline{\boldsymbol{\theta}}^t)\big)\big\|_2^2\\
    \le&~  \underbrace{ 9\sum_{i=1}^N p_i\mathbb{E} \|\nabla f_i(\boldsymbol{\theta}_i^t, \boldsymbol{\theta}_i^t)-\nabla f_i(\overline{\boldsymbol{\theta}}^t, \overline{\boldsymbol{\theta}}^t)\|_2^2}_{B_1} +  \underbrace{9\sum_{i=1}^N p_i\mathbb{E} \|\nabla f_i(\overline{\boldsymbol{\theta}}^t, \overline{\boldsymbol{\theta}}^t)- \sum_{j=1}^N p_j \nabla f_j(\overline{\boldsymbol{\theta}}^t, \overline{\boldsymbol{\theta}}^t)\|_2^2}_{B_2} \\
    &~+ \underbrace{9\sum_{i=1}^N p_i\mathbb{E} \| \sum_{j=1}^N p_j\big(\nabla f_j(\boldsymbol{\theta}_j^t, \boldsymbol{\theta}_j^t)-\nabla f_j(\overline{\boldsymbol{\theta}}^t, \overline{\boldsymbol{\theta}}^t)\big)\|_2^2}_{B_3}.
\end{align*}
Using Lemma \ref{lemma:fi_grad_continuity}, we have
\begin{align*}
    B_1\le &~ 9 \sum_{i=1}^N p_i L^2(1+\epsilon_i)^2 \mathbb{E}\|\boldsymbol{\theta}_i^t-\overline{\boldsymbol{\theta}}^t\|_2^2\le 9 \sum_{i=1}^N p_i L^2(1+\epsilon_{max})^2 \mathbb{E}\|\boldsymbol{\theta}_i^t-\overline{\boldsymbol{\theta}}^t\|_2^2.
\end{align*}
Using Assumption \ref{asm:a5_bound_var_local_grad}, we have
\begin{align*}
    B_2=&~ 9\sum_{i=1}^N p_i\mathbb{E} \big\|\nabla f_i(\overline{\boldsymbol{\theta}}^t, \overline{\boldsymbol{\theta}}^t)- \nabla f(\overline{\boldsymbol{\theta}}^t, \overline{\boldsymbol{\theta}}^t)\big\|_2^2\\
    \le &~ 9\sum_{i=1}^N p_i\varsigma^2(1+\mathbb{E}\|\overline{\boldsymbol{\theta}}^t-\boldsymbol{\theta}^{PS}\|_2^2)\\
    =&~ 9\varsigma^2 + 9\varsigma^2\mathbb{E}\|\overline{\boldsymbol{\theta}}^t-\boldsymbol{\theta}^{PS}\|_2^2.
\end{align*}
Using Lemma \ref{lemma:fi_grad_continuity}, we have
\begin{align*}
    B_3 \le &~ 9\sum_{i=1}^N p_i \sum_{j=1}^N p_j\mathbb{E} \| \nabla f_j(\boldsymbol{\theta}_j^t, \boldsymbol{\theta}_j^t)-\nabla f_j(\overline{\boldsymbol{\theta}}^t, \overline{\boldsymbol{\theta}}^t)\|_2^2\le 9 \sum_{i=1}^N p_i L^2(1+\epsilon_{max})^2 \mathbb{E}\|\boldsymbol{\theta}_i^t-\overline{\boldsymbol{\theta}}^t\|_2^2.
\end{align*}
Therefore,
\begin{align*}
    B_1+B_2+B_3 \le 18 \sum_{i=1}^N p_i L^2(1+\epsilon_{max})^2 \mathbb{E}\|\boldsymbol{\theta}_i^t-\overline{\boldsymbol{\theta}}^t\|_2^2+9\varsigma^2 + 9\varsigma^2\sum_{i=1}^N p_i\mathbb{E}\|\overline{\boldsymbol{\theta}}^t-\boldsymbol{\theta}^{PS}\|_2^2,
\end{align*}
which results in that
\begin{align*}
    B \le &~ 6 \sigma^2 \bigg( 1+ \mathbb{E} \bigg[\sum_{i=1}^N p_i \|\boldsymbol{\theta}^t_i - \boldsymbol{\theta}^{PS} \|_2^2\bigg]\bigg)+ 18 \sum_{i=1}^N p_i L^2(1+\epsilon_{max})^2 \mathbb{E}\|\boldsymbol{\theta}_i^t-\overline{\boldsymbol{\theta}}^t\|_2^2+\\
    &~9\varsigma^2 + 9\varsigma^2\mathbb{E}\|\overline{\boldsymbol{\theta}}^t-\boldsymbol{\theta}^{PS}\|_2^2\\
    \le  &~ 6 \sigma^2 \bigg( 1+ 2\sum_{i=1}^N p_i\mathbb{E} \|\boldsymbol{\theta}^t_i - \overline{\boldsymbol{\theta}}^{t}  \|_2^2 + 2\mathbb{E}\|\overline{\boldsymbol{\theta}}^{t}-\boldsymbol{\theta}^{PS}\|_2^2+ 18 \sum_{i=1}^N p_i L^2(1+\epsilon_{max})^2 \mathbb{E}\|\boldsymbol{\theta}_i^t-\overline{\boldsymbol{\theta}}^t\|_2^2+\\
    &~9\varsigma^2 + 9\varsigma^2\mathbb{E}\|\overline{\boldsymbol{\theta}}^t-\boldsymbol{\theta}^{PS}\|_2^2\bigg)\\
    =&~6\sigma^2+9\varsigma^2 +\big(12\sigma^2+18 L^2(1+\epsilon_{\max})^2 \big)\sum_{i=1}^N p_i\mathbb{E} \|\boldsymbol{\theta}^t_i - \overline{\boldsymbol{\theta}}^{t}  \|_2^2+(12\sigma^2+9\varsigma^2)\mathbb{E}\|\overline{\boldsymbol{\theta}}^t-\boldsymbol{\theta}^{PS}\|_2^2 .
\end{align*}
Inserting this formula into \eqref{eqn:consensuserr_1}, we obtain
\begin{align*}
     &~\sum_{i=1}^N p_i\mathbb{E}\|\boldsymbol{\theta}^{t+1}_i- \overline{\boldsymbol{\theta}}^{t+1}\|^2_2 \\
     \le &~ (1+\alpha_t) \sum_{i=1}^N p_i\mathbb{E}\|\boldsymbol{\theta}_i^{t} 
     - \overline{\boldsymbol{\theta}}^{t}\|_2^2 \\
     &~ + \eta_{t}^2(1+\alpha_t^{-1})\bigg(6\sigma^2+9\varsigma^2 +\big(12\sigma^2+18 L^2(1+\epsilon_{\max})^2 \big)\sum_{i=1}^N p_i\mathbb{E} \|\boldsymbol{\theta}^t_i - \overline{\boldsymbol{\theta}}^{t}  \|_2^2+(12\sigma^2+9\varsigma^2)\mathbb{E}\|\overline{\boldsymbol{\theta}}^t-\boldsymbol{\theta}^{PS}\|_2^2\bigg)\\
     =&~ \big( 1+\alpha_t+  \eta_{t}^2(1+\alpha_t^{-1})\big(12\sigma^2+18 L^2(1+\epsilon_{\max})^2 \big) \big) \sum_{i=1}^N p_i\mathbb{E}\|\boldsymbol{\theta}_i^{t}- \overline{\boldsymbol{\theta}}^{t}\|_2^2+ \eta_{t}^2(1+\alpha_t^{-1})(12\sigma^2+9\varsigma^2)\mathbb{E}\|\overline{\boldsymbol{\theta}}^t-\boldsymbol{\theta}^{PS}\|_2^2\\
     &~+ \eta_{t}^2(1+\alpha_t^{-1})(6\sigma^2+9\varsigma^2)
\end{align*}
where $\alpha_t >0$ is a free chosen parameter. Let $t_0 := \max\{s \mid s < t+1, s\in \mathcal{I}_E\} $ and $ c_3:= 12\sigma^2+18 L^2(1+\epsilon_{\max})^2$. Then we choose $\alpha_t = \frac{1}{2(t+1-t_0)}$, if we have
\begin{align}
    &~ \eta_{t}^2(1+\alpha_t^{-1})\big(12\sigma^2+18 L^2(1+\epsilon_{\max})^2 \big) = \eta_{t}^2(1+\alpha_t^{-1})c_3 \le \frac{1}{2(t+1-t_0)} \nonumber \\
    \Longleftrightarrow&~  \eta_t^2 \le \frac{1}{2c_3 (t+1-t_0)\big(1+2(t+1-t_0)\big)} \label{eqn:consensuserr_cond1},
\end{align}
then note that $1+\alpha_t^{-1} = 1+ 2(t+1-t_0)\le 2E-1$, 
\begin{align*}
    &~\sum_{i=1}^N p_i\mathbb{E}\|\boldsymbol{\theta}^{t+1}_i- \overline{\boldsymbol{\theta}}^{t+1}\|^2_2 \\
     \le &~ \frac{t+2-t_0}{t+1-t_0} \sum_{i=1}^N p_i\mathbb{E}\|\boldsymbol{\theta}_i^{t}- \overline{\boldsymbol{\theta}}^{t}\|_2^2 + \eta_t^2 (2E-1) (12\sigma^2+9\varsigma^2)\mathbb{E}\|\overline{\boldsymbol{\theta}}^t-\boldsymbol{\theta}^{PS}\|_2^2 + \eta_{t}^2(2E-1)(6\sigma^2+9\varsigma^2).
\end{align*}
Continuing the above expansion until $t_0$ and leveraging $\eta_s\le \eta_{t_0}\le 2\eta_{t_0+E}\le 2\eta_t$ gives us
\begin{align}
    &~\sum_{i=1}^N p_i\mathbb{E}\|\boldsymbol{\theta}^{t+1}_i- \overline{\boldsymbol{\theta}}^{t+1}\|^2_2 \nonumber\\
     \le &~ \frac{t+2-t_0}{t_0+1-t_0} \sum_{i=1}^N p_i\mathbb{E}\|\boldsymbol{\theta}^{t_0}_i- \overline{\boldsymbol{\theta}}^{t_0}\|^2_2 + \sum_{s=t_0}^{t} \frac{t+2-t_0}{s+2-t_0}  \eta_s^2 (2E-1) (12\sigma^2+9\varsigma^2)\mathbb{E}\|\overline{\boldsymbol{\theta}}^s-\boldsymbol{\theta}^{PS}\|_2^2\nonumber\\
     &~ + \sum_{s=t_0}^{t} \frac{t+2-t_0}{s+2-t_0}  \eta_s^2 (2E-1) (6\sigma^2+9\varsigma^2)\nonumber\\
     =&~ \sum_{s=0}^{t-t_0} \frac{t+2-t_0}{s+2}  \eta_s^2 (2E-1) (12\sigma^2+9\varsigma^2)\mathbb{E}\|\overline{\boldsymbol{\theta}}^s-\boldsymbol{\theta}^{PS}\|_2^2 + \sum_{s=0}^{t-t_0} \frac{t+2-t_0}{s+2}   \eta_s^2 (2E-1) (6\sigma^2+9\varsigma^2)\nonumber\\
     \le & \sum_{s=0}^{t-t_0} \frac{t+2-t_0}{s+2}  \eta_t^2 (2E-1) (48\sigma^2+36\varsigma^2)\mathbb{E}\|\overline{\boldsymbol{\theta}}^s-\boldsymbol{\theta}^{PS}\|_2^2 + \sum_{s=0}^{t-t_0} \frac{t+2-t_0}{s+2}   \eta_t^2 (2E-1) (24\sigma^2+36\varsigma^2).\label{eqn:consensuserr_2}
\end{align}

With the above formula and Lemma \ref{lem:descent}, we now prove that if $\eta_0$ is sufficiently small, the for any $t$, we have $\mathbb{E}\|\overline{\boldsymbol{\theta}}^t-\boldsymbol{\theta}^{PS}\|_2^2 \le \mathbb{E}\|\overline{\boldsymbol{\theta}}^0-\boldsymbol{\theta}^{PS}\|_2^2$. We first derive the following inequality, which we will use later. Note that for any $t$ where $t_0 := \max\{s \mid s < t+1, s\in \mathcal{I}_E\} $, we have
\begin{align}
    \sum_{s=0}^{t-t_0} \frac{t+2-t_0}{s+2}  = (t+2-t_0)(\frac{1}{2}+ \ldots+\frac{1}{t-t_0+2})\le (t+2-t_0)\log (t+2-t_0)\le E\log E.\label{eqn:consensuserr_ineq}
\end{align}

We prove $\mathbb{E}\|\overline{\boldsymbol{\theta}}^t-\boldsymbol{\theta}^{PS}\|_2^2 \le \mathbb{E}\|\overline{\boldsymbol{\theta}}^0-\boldsymbol{\theta}^{PS}\|_2^2$ by induction. 

First, this inequality clearly holds for $t=0$. Suppose it holds for $0\le s \le t$ where $t\le E-1$. Then by Lemma \ref{lem:descent} and \eqref{eqn:consensuserr_2}, we have
\begin{align*}
        &~\mathbb{E} [\| \overline{\boldsymbol{\theta}}^{t+1} - \boldsymbol{\theta}^{PS} \|^2_2] \\
        =&~ \mathbb{E} [\| \overline{\boldsymbol{w}}^{t+1} - \boldsymbol{\theta}^{PS} \|^2_2]  \\
        \leq  &~ (1 - \Tilde{\mu} \eta_t) \mathbb{E}\|\overline{\boldsymbol{\theta}}^{0} - \boldsymbol{\theta}^{PS}\|_2^2 + 2 \sigma^2 \eta_t^2  + ( c_1 \eta_t + c_2 \eta_t^2 ) \bigg(\sum_{s=0}^{t-t_0-1} \frac{t+1-t_0}{s+2}  \eta_{t-1}^2 (2E-1) (48\sigma^2+36\varsigma^2)\mathbb{E}\|\overline{\boldsymbol{\theta}}^0-\boldsymbol{\theta}^{PS}\|_2^2 \\&~ + \sum_{s=0}^{t-t_0-1} \frac{t+1-t_0}{s+2}   \eta_{t-1}^2 (2E-1) (24\sigma^2+36\varsigma^2)\bigg)\\
        =&~ (1 - \Tilde{\mu} \eta_t) \mathbb{E}\|\overline{\boldsymbol{\theta}}^{0} - \boldsymbol{\theta}^{PS}\|_2^2 + 2 \sigma^2 \eta_t^2  \\
        &~ + ( c_1 \eta_t + c_2 \eta_t^2 ) \eta_{t-1}^2 (2E^2-E)\log E\bigg( (48\sigma^2+36\varsigma^2)\mathbb{E}\|\overline{\boldsymbol{\theta}}^0-\boldsymbol{\theta}^{PS}\|_2^2+ (24\sigma^2+36\varsigma^2)\bigg).
    \end{align*}
By \eqref{eqn:consensuserr_cond1}, we need $\eta_0\le \frac{1}{6c_3}$, together with $\eta_{t-1}\le 2\eta_t$ implies
\begin{align*}
    ( c_1 \eta_t + c_2 \eta_t^2 ) \eta_{t-1}^2\le \eta_t^2 ( c_1  + c_2(6c_3)^{-1} )\frac{c_3}{3}.
\end{align*}
Therefore, we have
\begin{align*}
    &~\mathbb{E} [\| \overline{\boldsymbol{\theta}}^{t+1} - \boldsymbol{\theta}^{PS} \|^2_2] \\
        \le &~ (1 - \Tilde{\mu} \eta_t) \mathbb{E}\|\overline{\boldsymbol{\theta}}^{0} - \boldsymbol{\theta}^{PS}\|_2^2 + 2 \sigma^2 \eta_t^2  \\
        &~ + \eta_t^2 ( c_1c_3  + c_2/6 ) (2E^2-E)\log E\bigg( (16\sigma^2+12\varsigma^2)\mathbb{E}\|\overline{\boldsymbol{\theta}}^0-\boldsymbol{\theta}^{PS}\|_2^2+ (8\sigma^2+12\varsigma^2)\bigg)
\end{align*}
whose right-hand side is no larger than $\mathbb{E}\|\overline{\boldsymbol{\theta}}^0-\boldsymbol{\theta}^{PS}\|_2^2$ if 
\begin{align}
    \eta_t\le&~ \eta_0 \nonumber\\
    \le &~ \frac{\Tilde{\mu} \mathbb{E}\|\overline{\boldsymbol{\theta}}^0-\boldsymbol{\theta}^{PS}\|_2^2}{2\sigma^2+( c_1c_3  + c_2/6 ) (2E^2-E)\log E\big( (16\sigma^2+12\varsigma^2)\mathbb{E}\|\overline{\boldsymbol{\theta}}^0-\boldsymbol{\theta}^{PS}\|_2^2+ (8\sigma^2+12\varsigma^2)\big) }= \hat{\eta}_0.\label{eqn:consensuserr_cond2}
\end{align}
Thus we have proved that for any $0\le t\le E$, if \eqref{eqn:consensuserr_cond2} holds, then $\mathbb{E}\|\overline{\boldsymbol{\theta}}^t-\boldsymbol{\theta}^{PS}\|_2^2 \le \mathbb{E}\|\overline{\boldsymbol{\theta}}^0-\boldsymbol{\theta}^{PS}\|_2^2$. The same proof technique can be extended to any $n E\le t\le (n+1)E$ where $n\in \mathbb{N}_+$ and thus for any $t$, if \eqref{eqn:consensuserr_cond2} holds, then $\mathbb{E}\|\overline{\boldsymbol{\theta}}^t-\boldsymbol{\theta}^{PS}\|_2^2 \le \mathbb{E}\|\overline{\boldsymbol{\theta}}^0-\boldsymbol{\theta}^{PS}\|_2^2$. 

Therefore, under \eqref{eqn:consensuserr_cond1}, \eqref{eqn:consensuserr_cond2} and $\eta_{t-1}\le 2\eta_t$, by \eqref{eqn:consensuserr_1} and \eqref{eqn:consensuserr_2}, if $t\notin \mathcal{I}_E$, we have
\begin{align*}
    \sum_{i=1}^N p_i\mathbb{E}\|\boldsymbol{\theta}^{t}_i- \overline{\boldsymbol{\theta}}^{t}\|^2_2 \le \eta_{t-1}^2 (2E^2-E)\log E (48\sigma^2+36\varsigma^2)\mathbb{E}\|\overline{\boldsymbol{\theta}}^0-\boldsymbol{\theta}^{PS}\|_2^2 + \eta_{t-1}^2 (2E^2-E)\log E (24\sigma^2+36\varsigma^2)\\
    \le 4\eta_{t}^2 (2E^2-E)\log E (48\sigma^2+36\varsigma^2)\mathbb{E}\|\overline{\boldsymbol{\theta}}^0-\boldsymbol{\theta}^{PS}\|_2^2 + 4\eta_{t}^2 (2E^2-E)\log E (24\sigma^2+36\varsigma^2).
\end{align*}
\end{proof}

\begin{proof}[Proof of Lemma \ref{lem:cleandescent}]
We discuss in two cases, $t\in \mathcal{I}_E$ and $t\notin \mathcal{I}_E$. If $t\in \mathcal{I}_E$, then we have $\boldsymbol{\theta}_i^t =\overline{\boldsymbol{\theta}}^t$, and by Lemma \ref{lem:descent},
\begin{align*}
    &~\mathbb{E} [\| \overline{\boldsymbol{\theta}}^{t+1} - \boldsymbol{\theta}^{PS} \|^2_2]  \\
        \leq &~ (1 - \Tilde{\mu} \eta_t) \mathbb{E}\|\overline{\boldsymbol{\theta}}^{t} - \boldsymbol{\theta}^{PS}\|_2^2 + 2 \sigma^2 \eta_t^2 + ( c_1 \eta_t + c_2 \eta_t^2 ) \sum_{i=1}^N p_i \mathbb{E}\| \boldsymbol{\theta}_i^t - \overline{\boldsymbol{\theta}}^{t} \|_2^2\\
        =&~ (1 - \Tilde{\mu} \eta_t) \mathbb{E}\|\overline{\boldsymbol{\theta}}^{t} - \boldsymbol{\theta}^{PS}\|_2^2 + 2 \sigma^2 \eta_t^2\\
        \le &~ (1 - \Tilde{\mu} \eta_t) \mathbb{E}\|\overline{\boldsymbol{\theta}}^{t} - \boldsymbol{\theta}^{PS}\|_2^2 + B \eta_t^2.
\end{align*}

If $t\notin \mathcal{I}_E$, combining Lemma \ref{lem:descent} and Lemma \ref{lem:consensuserr}, we have
\begin{align*}
         &~\mathbb{E} [\| \overline{\boldsymbol{\theta}}^{t+1} - \boldsymbol{\theta}^{PS} \|^2_2]  \\
        \leq &~ (1 - \Tilde{\mu} \eta_t) \mathbb{E}\|\overline{\boldsymbol{\theta}}^{t} - \boldsymbol{\theta}^{PS}\|_2^2 + 2 \sigma^2 \eta_t^2  + ( c_1 \eta_t + c_2 \eta_t^2 ) \sum_{i=1}^N p_i \mathbb{E}\| \boldsymbol{\theta}_i^t - \overline{\boldsymbol{\theta}}^{t} \|_2^2 \\
        \le & ~ (1 - \Tilde{\mu} \eta_t) \mathbb{E}\|\overline{\boldsymbol{\theta}}^{t} - \boldsymbol{\theta}^{PS}\|_2^2 + 2 \sigma^2 \eta_t^2\\
        &~ + ( 4 c_1 \eta_t + 4 c_2 \eta_t^2 )\bigg(\eta_{t}^2 (2E^2-E)\log E (48\sigma^2+36\varsigma^2)\mathbb{E}\|\overline{\boldsymbol{\theta}}^0-\boldsymbol{\theta}^{PS}\|_2^2 + \eta_{t}^2 (2E^2-E)\log E (24\sigma^2+36\varsigma^2)\bigg) \\
        \le &~  (1 - \Tilde{\mu} \eta_t) \mathbb{E}\|\overline{\boldsymbol{\theta}}^{t} - \boldsymbol{\theta}^{PS}\|_2^2 + 2B \eta_t^2.
    \end{align*}
\end{proof}

\section{Proof of Theorem \ref{thm:partial_w_rep_convergence} and Theorem \ref{thm:partial_wo_rep_convergence}}\label{appensec:partial}
\subsection{Additional Notations}
Similar to Appendix \ref{appensec:full},  in our analysis, for the sake of convenience, we will define two additional sequences as $\overline{\boldsymbol{w}}^{t}:= \sum_{i=1}^N p_i \boldsymbol{w}_i^t$ and $\overline{\boldsymbol{\theta}}^{t}:= \sum_{i=1}^N p_i \boldsymbol{\theta}_i^t$, following that of \cite{ICLR_2020_Li_Convergence}. We note that $\overline{\boldsymbol{w}}^t$ results from a single step of SGD from $\overline{\boldsymbol{\theta}}^t$. When $t+1\notin \mathcal{I}_E$, both $\overline{\boldsymbol{w}}^{t}$ and $\overline{\boldsymbol{\theta}}^{t}$ are unaccessible. When $t+1\in \mathcal{I}_E$, we can obtain $\overline{\boldsymbol{\theta}}^t$. In addition, we also define $ \overline{\boldsymbol{g}}_t := \sum_{i=1}^N p_i \nabla f_i(\boldsymbol{\theta}_i^t; \boldsymbol{\theta}_i^t),~~ \boldsymbol{g}_t := \sum_{i=1}^N p_i \nabla \ell(\boldsymbol{\theta}_i^t; Z_i^{t+1})$ where $Z_i^{t+1} \sim \mathcal{D}_i(\boldsymbol{\theta}_i^{t})$. It is clear that in full participation, $\overline{\boldsymbol{w}}^{t+1} = \overline{\boldsymbol{w}}^{t} - \eta_t \boldsymbol{g}_t$ and $\mathbb{E} \boldsymbol{g}_t = \overline{\boldsymbol{g}}_t$. Notice now we do not have $\overline{\boldsymbol{\theta
}}^{t} = \overline{\boldsymbol{w}}^{t}$ for any $t$. But we will show later that they are equal with expectation to the choice of $\mathcal{S}_t$. 

In particular, in our analysis, there would be two types of randomness, one from the stochastic gradients and one from the random sampling of the devices. All analysis in Appendix \ref{appensec:full} only involves the former. To make a distinguishment, we use $\mathbb{E}_{\mathcal{S}_t}$ to denote the latter.
\subsection{Key Lemmas}
We show in this subsection the key lemmas for proving the convergence and defer proofs to later parts. We first show that the sampling schemes I \& II are unbiased.
\begin{lemma}{\cite{ICLR_2020_Li_Convergence}} \label{lem:partial_unbiased} (Unbiased sampling scheme). If $t+1 \in \mathcal{I}_E$, for Scheme I and Scheme II, we have
$$
\mathbb{E}_{\mathcal{S}_t}\big[\overline{\boldsymbol{\theta}}^{t+1}\big]=\overline{\boldsymbol{w}}^{t+1} .
$$
\end{lemma}

Similar to Lemma \ref{lem:descent}, we are going to establish an upper bound for $\mathbb{E} [\| \overline{\boldsymbol{\theta}}^{t+1} - \boldsymbol{\theta}^{PS} \|^2_2]$. When $t+1\notin \mathcal{I}_E$, we have $\overline{\boldsymbol{\theta}}^{t+1} = \overline{\boldsymbol{w}}^{t+1}$ for both schemes, and therefore this is equivalent to establishing an upper bound for $\mathbb{E} [\| \overline{\boldsymbol{w}}^{t+1} - \boldsymbol{\theta}^{PS} \|^2_2]$. However, when $t+1\in \mathcal{I}_E$, we only have $\mathbb{E}_{\mathcal{S}_t}\big[\overline{\boldsymbol{\theta}}^{t+1}\big]=\overline{\boldsymbol{w}}^{t+1} $ and we need other upper-bounding strategies.

\begin{lemma}\label{lem:descent_partial}
Under Assumptions \ref{asm:a1_obj_str_conv}, \ref{asm:a2_obj_smooth}, \ref{asm:a3_map_sensitivity}, \ref{asm:a4_bound_var_sto_grad}, for scheme I \& II:
\begin{enumerate}
    \item if $t+1\notin \mathcal{I}_E$, 
    \begin{align*}
        \mathbb{E} [\| \overline{\boldsymbol{\theta}}^{t+1} - \boldsymbol{\theta}^{PS} \|^2_2] = \mathbb{E} [\| \overline{\boldsymbol{w}}^{t+1} - \boldsymbol{\theta}^{PS} \|^2_2]  
        \leq  (1 - \Tilde{\mu} \eta_t) \mathbb{E}\|\overline{\boldsymbol{\theta}}^{t} - \boldsymbol{\theta}^{PS}\|_2^2 + 2 \sigma^2 \eta_t^2  + ( c_1 \eta_t + c_2 \eta_t^2 ) \sum_{i=1}^N p_i \mathbb{E}\| \boldsymbol{\theta}_i^t - \overline{\boldsymbol{\theta}}^{t} \|_2^2.
    \end{align*}
    \item if $t+1\in \mathcal{I}_E$:
    for scheme I,
        \begin{align*}
        &~\mathbb{E} [\| \overline{\boldsymbol{\theta}}^{t+1} - \boldsymbol{\theta}^{PS} \|^2_2] \\
        \leq &~ \frac{1}{K} \sum_{k=1}^N p_k \mathbb{E}\| \boldsymbol{w}_{k}^{t+1} - \overline{\boldsymbol{w}}^{t+1} \|_2^2 + (1 - \Tilde{\mu} \eta_t) \mathbb{E}\|\overline{\boldsymbol{\theta}}^{t} - \boldsymbol{\theta}^{PS}\|_2^2 + 2 \sigma^2 \eta_t^2  + ( c_1 \eta_t + c_2 \eta_t^2 ) \sum_{i=1}^N p_i \mathbb{E}\| \boldsymbol{\theta}_i^t - \overline{\boldsymbol{\theta}}^{t} \|_2^2,
    \end{align*}
       while for scheme II, 
        \begin{align*}
        &~\mathbb{E} [\| \overline{\boldsymbol{\theta}}^{t+1} - \boldsymbol{\theta}^{PS} \|^2_2] \\
        \leq &~ \frac{N}{K(N-1)} \bigg(1 - \frac{K}{N}\bigg) \sum_{k=1}^N p_k \mathbb{E}\| \boldsymbol{w}_{k}^{t+1} - \overline{\boldsymbol{w}}^{t+1} \|_2^2 + (1 - \Tilde{\mu} \eta_t) \mathbb{E}\|\overline{\boldsymbol{\theta}}^{t} - \boldsymbol{\theta}^{PS}\|_2^2 + 2 \sigma^2 \eta_t^2  + ( c_1 \eta_t + c_2 \eta_t^2 ) \sum_{i=1}^N p_i \mathbb{E}\| \boldsymbol{\theta}_i^t - \overline{\boldsymbol{\theta}}^{t} \|_2^2.
        \end{align*}
    
\end{enumerate}
    where $\epsilon_{max} := \max_i \epsilon_i, \overline{\epsilon}: = \sum_{i=1}^N p_i \epsilon_i, 
    c_1 := \frac{L (1+\epsilon_{max})^2}{2 \delta \overline{\epsilon}},
    c_2 := 4 [\sigma^2 + L^2(1+\epsilon_{max})^2], 
    \Tilde{\mu} := \mu - (1+\delta) \overline{\epsilon} L$.
\end{lemma}

To really give a descent lemma as in SGD analysis, we have to bound $\sum_{i=1}^N p_i \mathbb{E}\| \boldsymbol{\theta}_i^t - \overline{\boldsymbol{\theta}}^{t} \|_2^2$ for $t\notin \mathcal{I}_E$ and $\sum_{i=1}^N p_i \mathbb{E}\| \boldsymbol{w}_i^t - \overline{\boldsymbol{w}}^{t} \|_2^2$ for $t\in \mathcal{I}_E$, given by the following lemma.

\begin{lemma}\label{lem:consensuserr_partial}
    Under Assumption \ref{asm:a1_obj_str_conv}, \ref{asm:a2_obj_smooth}, \ref{asm:a3_map_sensitivity}, \ref{asm:a4_bound_var_sto_grad}, \ref{asm:a5_bound_var_local_grad}, if $\{\eta_t\}$ is non-increasing, $\eta_{t}\le 2\eta_{t+E}$, $t \notin  \mathcal{I}_E$, $ \eta_t^2 \le 1/\big(2c_3 (t+1-t_0)(1+2(t+1-t_0))\big) $, and
    \begin{align*}
        \eta_0\le \tilde{\eta}_0:= \frac{\Tilde{\mu} \mathbb{E}\|\overline{\boldsymbol{\theta}}^0-\boldsymbol{\theta}^{PS}\|_2^2}{2\sigma^2+( c_1c_3  + c_2/6 ) (2E^2+3E+1)\log (E+1)\big( (16\sigma^2+12\varsigma^2)\mathbb{E}\|\overline{\boldsymbol{\theta}}^0-\boldsymbol{\theta}^{PS}\|_2^2+ (8\sigma^2+12\varsigma^2)\big) },
    \end{align*}
    then \begin{enumerate}
   \item for scheme I,
        \begin{align*}
        &~\mathbb{E} [\| \overline{\boldsymbol{\theta}}^{t+1} - \boldsymbol{\theta}^{PS} \|^2_2] \\
        \le &~ (1 - \Tilde{\mu} \eta_t) \mathbb{E}\|\overline{\boldsymbol{\theta}}^{t} - \boldsymbol{\theta}^{PS}\|_2^2 + 2 \sigma^2 \eta_t^2 \\
        &~+ (4c_1 \eta_t + 4c_2 \eta_t^2 +K^{-1}) (2E^2+3E+1)\log (E+1) \eta_t^2 \bigg(  (48\sigma^2+36\varsigma^2)\mathbb{E}\|\overline{\boldsymbol{\theta}}^0-\boldsymbol{\theta}^{PS}\|_2^2 +(24\sigma^2+36\varsigma^2)\bigg),
    \end{align*}
      \item for scheme II, 
        \begin{align*}
        &~\mathbb{E} [\| \overline{\boldsymbol{\theta}}^{t+1} - \boldsymbol{\theta}^{PS} \|^2_2] \\
        \le &~ (1 - \Tilde{\mu} \eta_t) \mathbb{E}\|\overline{\boldsymbol{\theta}}^{t} - \boldsymbol{\theta}^{PS}\|_2^2 + 2 \sigma^2 \eta_t^2 \\
        &~+ \bigg(4c_1 \eta_t + 4c_2 \eta_t^2 +\frac{N-K}{K(N-1)}\bigg ) (2E^2+3E+1)\log (E+1) \eta_t^2 \bigg(  (48\sigma^2+36\varsigma^2)\mathbb{E}\|\overline{\boldsymbol{\theta}}^0-\boldsymbol{\theta}^{PS}\|_2^2 +(24\sigma^2+36\varsigma^2)\bigg).
        \end{align*}
\end{enumerate}
where for any $t$, where $\epsilon_{max} := \max_i \epsilon_i, \overline{\epsilon}: = \sum_{i=1}^N p_i \epsilon_i, 
    c_1 := \frac{L (1+\epsilon_{max})^2}{2 \delta \overline{\epsilon}},
    c_2 := 4 [\sigma^2 + L^2(1+\epsilon_{max})^2], 
    \Tilde{\mu} := \mu - (1+\delta) \overline{\epsilon} L$, $c_3:= 12\sigma^2+18 L^2(1+\epsilon_{\max})^2$.
    
(One should note that $4c_1 \eta_t + 4c_2 \eta_t^2$, $(48\sigma^2+36\varsigma^2)$, and $(24\sigma^2+36\varsigma^2)$ comes from several times of applying $\eta_{t-1}\le 2\eta_t$ and the real constants could be much smaller by choosing stepsizes carefully.)    
\end{lemma}

The following lemma gives us a standard descent lemma in SGD analysis under technical conditions for establishing the $\mathcal{O}(\frac{1}{T})$ convergence in Theorem \ref{thm:partial_w_rep_convergence} and \ref{thm:partial_wo_rep_convergence}.
\begin{lemma}\label{lem:cleandescent_partial}
    Under Assumption \ref{asm:a1_obj_str_conv}, \ref{asm:a2_obj_smooth}, \ref{asm:a3_map_sensitivity}, \ref{asm:a4_bound_var_sto_grad}, \ref{asm:a5_bound_var_local_grad}, if $\{\eta_t\}$ is non-increasing, $\eta_{t}\le 2\eta_{t+E}$, $t \notin  \mathcal{I}_E$, $ \eta_t^2 \le 1/\big(2c_3 (t+1-t_0)(1+2(t+1-t_0))\big) $, and
    \begin{align*}
        \eta_0\le \tilde{\eta}_0:= \frac{\Tilde{\mu} \mathbb{E}\|\overline{\boldsymbol{\theta}}^0-\boldsymbol{\theta}^{PS}\|_2^2}{2\sigma^2+( c_1c_3  + c_2/6 ) (2E^2+3E+1)\log (E+1)\big( (16\sigma^2+12\varsigma^2)\mathbb{E}\|\overline{\boldsymbol{\theta}}^0-\boldsymbol{\theta}^{PS}\|_2^2+ (8\sigma^2+12\varsigma^2)\big) },
    \end{align*}
    then \begin{enumerate}
   \item for scheme I, 
   \begin{align*}
        \mathbb{E} [\| \overline{\boldsymbol{\theta}}^{t+1} - \boldsymbol{\theta}^{PS} \|^2_2] \le  (1 - \Tilde{\mu} \eta_t) \mathbb{E}\|\overline{\boldsymbol{\theta}}^{t} - \boldsymbol{\theta}^{PS}\|_2^2 + B_1 \eta_t^2,
    \end{align*}
    with 
   \begin{align*}
       B_1:= 2 \sigma^2  + (4c_1 \eta_t + 4c_2 \eta_t^2 +K^{-1}) (2E^2+3E+1)\log (E+1)  \bigg(  (48\sigma^2+36\varsigma^2)\mathbb{E}\|\overline{\boldsymbol{\theta}}^0-\boldsymbol{\theta}^{PS}\|_2^2 +(24\sigma^2+36\varsigma^2)\bigg),
   \end{align*}
    \item for scheme II, 
    \begin{align*}
        \mathbb{E} [\| \overline{\boldsymbol{\theta}}^{t+1} - \boldsymbol{\theta}^{PS} \|^2_2] \le  (1 - \Tilde{\mu} \eta_t) \mathbb{E}\|\overline{\boldsymbol{\theta}}^{t} - \boldsymbol{\theta}^{PS}\|_2^2 + B_2 \eta_t^2
    \end{align*}
    with
   \begin{align*}
       B_2:= 2 \sigma^2  + \bigg(4c_1 \eta_t + 4c_2 \eta_t^2 +\frac{N-K}{K(N-1)}\bigg )(2E^2+3E+1)\log (E+1)  \bigg(  (48\sigma^2+36\varsigma^2)\mathbb{E}\|\overline{\boldsymbol{\theta}}^0-\boldsymbol{\theta}^{PS}\|_2^2 +(24\sigma^2+36\varsigma^2)\bigg),
   \end{align*}  
\end{enumerate}
where for any $t$, where $\epsilon_{max} := \max_i \epsilon_i, \overline{\epsilon}: = \sum_{i=1}^N p_i \epsilon_i, 
    c_1 := \frac{L (1+\epsilon_{max})^2}{2 \delta \overline{\epsilon}},
    c_2 := 4 [\sigma^2 + L^2(1+\epsilon_{max})^2], 
    \Tilde{\mu} := \mu - (1+\delta) \overline{\epsilon} L$, $c_3:= 12\sigma^2+18 L^2(1+\epsilon_{\max})^2$.
\end{lemma}

\subsection{Completing the proof of Theorem \ref{thm:partial_w_rep_convergence} and \ref{thm:partial_wo_rep_convergence}} \label{sec:proofoftheorem3.2-3.3}
We restate the definitions of all the constants here:

{\bf Constants independent of system design.} \\
$\epsilon_{max} :=  \max_i \epsilon_i$,

    $\overline{\epsilon} := \sum_{i=1}^N p_i \epsilon_i$,
    
    $\Tilde{\mu} :=  \mu - (1+\delta) \overline{\epsilon} L$,
    
    $c_1 :=  \big(L (1+\epsilon_{max})^2\big)/(2 \delta \overline{\epsilon})$,
    
    $c_2 := 4 \big[\sigma^2 + L^2(1+\epsilon_{max})^2\big]$, 
    
    $c_3:= 6\big[2\sigma^2+ 3 L^2(1+\epsilon_{\max})^2\big]$,
    
    $c_4:= 16\sigma^2+12\varsigma^2+ (8\sigma^2+12\varsigma^2)/\mathbb{E}\|\overline{\boldsymbol{\theta}}^0-\boldsymbol{\theta}^{PS}\|_2^2$,
    
    $c_5:=  (48\sigma^2+36\varsigma^2)\mathbb{E}\|\overline{\boldsymbol{\theta}}^0-\boldsymbol{\theta}^{PS}\|_2^2 +(24\sigma^2+36\varsigma^2)$.

{\bf Constants related to system design (e.g., $E, K$).} \\
    $\hat{\eta}_0 :=\Tilde{\mu}/ \big(2\sigma^2+( c_1c_3  + c_2/6 ) c_4(2E^2-E)\log E\big)$,
    
    $B :=2\sigma^2+ ( 4 c_1 \hat{\eta}_0 + 4 c_2 \hat{\eta}_0^2 )c_5(2E^2-E)\log E$ ,

    $c_6 := (2E^2+3E+1)\log (E+1)$,

    $\tilde{\eta}_0:=\Tilde{\mu} / \big(2\sigma^2+( c_1c_3  + c_2/6 ) c_4c_6\big)$,
    
    $B_1 := 2 \sigma^2  + (4c_1 \tilde{\eta}_0 + 4c_2 \tilde{\eta}_0^2 +1/K) c_5 c_6 $,
    
    $B_2 := 2 \sigma^2 + \big(4c_1 \tilde{\eta}_0 + 4c_2 \tilde{\eta}_0^2 + \frac{N-K}{K(N-1)}\big )c_5c_6$.

Instead of proving Theorem \ref{thm:partial_w_rep_convergence} and \ref{thm:partial_wo_rep_convergence} directly, we prove a more general version of convergence results suppose that some conditions about the stepsize are satisfied. Then we will show that the stepsizes given in Theorem \ref{thm:partial_w_rep_convergence} and \ref{thm:partial_wo_rep_convergence} satisfy the conditions. The proof of lemmas will be deferred 

\begin{theorem}
Under Assumption \ref{asm:a1_obj_str_conv}, \ref{asm:a2_obj_smooth}, \ref{asm:a3_map_sensitivity}, \ref{asm:a4_bound_var_sto_grad}, \ref{asm:a5_bound_var_local_grad}, for a diminishing stepsize $\eta_t = \frac{\beta}{t+\gamma}$ where  $\beta > \frac{1}{\Tilde{\mu}}$, $\gamma>0$ such that $\eta_0\le \tilde{\eta}_0$, $\eta_{t}\le 2\eta_{t+E}$, and $ \eta_t^2 \le 1/\big(2c_3 (t+1-t_0)(1+2(t+1-t_0))\big) $,
    then \begin{enumerate}
   \item for scheme I, 
   \begin{align*}
        \mathbb{E} [\| \overline{\boldsymbol{\theta}}^{t+1} - \boldsymbol{\theta}^{PS} \|^2_2] \le   \frac{\upsilon}{\gamma+t},
    \end{align*}
    where $\upsilon=\max \left\{\frac{4 B_1}{ \Tilde{\mu}^2},\gamma \mathbb{E} [\| \overline{\boldsymbol{\theta}}^{0} - \boldsymbol{\theta}^{PS} \|^2_2]\right\}$;
    \item for scheme II, 
    \begin{align*}
        \mathbb{E} [\| \overline{\boldsymbol{\theta}}^{t+1} - \boldsymbol{\theta}^{PS} \|^2_2] \le   \frac{\upsilon}{\gamma+t}
    \end{align*}
    where $\upsilon=\max \left\{\frac{4 B_2}{ \Tilde{\mu}^2},\gamma \mathbb{E} [\| \overline{\boldsymbol{\theta}}^{0} - \boldsymbol{\theta}^{PS} \|^2_2]\right\}$.
\end{enumerate}
\end{theorem}
\begin{proof}
We give a proof for scheme I and the proof for scheme II follows exactly the same way. 

Let $\Delta_t:=\mathbb{E} [\| \overline{\boldsymbol{\theta}}^{t} - \boldsymbol{\theta}^{PS} \|^2_2]$, then from Lemma \ref{lem:cleandescent_partial}, we have
    \begin{align*}
        \Delta_{t+1}\le (1 - \Tilde{\mu} \eta_t) \mathbb{E}\|\overline{\boldsymbol{\theta}}^{t} - \boldsymbol{\theta}^{PS}\|_2^2 + B_1 \eta_t^2 .
    \end{align*}
    For a diminishing stepsize $\eta_t = \frac{\beta}{t+\gamma}$ where  $\beta > \frac{1}{\Tilde{\mu}}$, $\gamma>0$ such that $ \eta_t^2 \le 1/\big(2c_3 (t+1-t_0)(1+2(t+1-t_0))\big) $, $\eta_0 \le \hat{\eta}_0$, and $\eta_t\le 2\eta_{t+E}$, we will prove that $\Delta_t\le \frac{\upsilon}{\gamma+t}$ where $\upsilon=\max \left\{\frac{\beta^2 B_1}{\beta \Tilde{\mu}-1},\gamma \Delta_0\right\}=\max \left\{\frac{4 B_1}{\Tilde{\mu}^2},\gamma \Delta_0\right\}$ by induction.

    Firstly, $\Delta_0 \le \frac{\upsilon}{\gamma} $ by the definition of $\upsilon$. Assume that for some $0\le t$, $\Delta_t \le \frac{\upsilon}{\gamma+t} $, then
    \begin{align*}
\Delta_{t+1} \leq &~ \left(1-\eta_t \Tilde{\mu}\right) \Delta_t+\eta_t^2 B_1 \\
\leq& ~\left(1-\frac{\beta \Tilde{\mu}}{t+\gamma}\right) \frac{v}{t+\gamma}+\frac{\beta^2 B_1}{(t+\gamma)^2} \\
=&~ \frac{t+\gamma-1}{(t+\gamma)^2} v+\left[\frac{\beta^2 B_1}{(t+\gamma)^2}-\frac{\beta \Tilde{\mu}-1}{(t+\gamma)^2} v\right] \\
\leq& ~ \frac{v}{t+\gamma+1} .
    \end{align*}

    Specifically, if we choose $\beta = \frac{2}{\Tilde{\mu}}$, $\gamma = \max\{ \frac{2}{\Tilde{\mu} \tilde{\eta}_0}, E,\frac{2}{\Tilde{\mu}} \sqrt{(4E^2+10E+6)(12\sigma^2+18 L^2(1+\epsilon_{\max})^2)}\}$, then we have 
    \begin{align*}
        \eta_0  = \frac{\beta}{\gamma}  \le \frac{2}{\Tilde{\mu} \frac{2}{\Tilde{\mu} \tilde{\eta}_0}} = \hat{\eta}_0
        \end{align*}
    and 
    \begin{align*}
        \eta_t - 2\eta_{t+E} = \frac{\beta}{\gamma+t} - \frac{2\beta}{\gamma+t+E} = \frac{\beta (E-\gamma-t)}{(\gamma+t)(\gamma+t+E)}\le \frac{\beta (E-\gamma)} {(\gamma+t)(\gamma+t+E)} \le 0. 
    \end{align*}
    To prove that $ \eta_t^2 \le 1/\big(2c_3 (t+1-t_0)(1+2(t+1-t_0))\big)$ for any $t$, it suffices to prove that for $0\le t\le E$ because $\{\eta_t\}$, i.e., $t_0=0$, is non-increasing and $t+1-t_0$ is periodic with period $E$. When $t_0=0$, we need to prove $ \eta_t^2 \le 1/\big(2c_3 (t+1)(1+2(t+1))\big)$ for $0\le t\le E$, which is satisfied if
    \begin{align*}
        &~ \max_{0\le t\le E} \eta_t \le \min_{0\le t\le E} \sqrt{1/\big(2c_3 (t+1)(1+2(t+1))\big)}\\
       \Longleftrightarrow &~ \eta_0 \le \sqrt{\frac{1}{(4E^2+10E+6)c_3}} \\
       \Longleftrightarrow &~ \gamma \ge \beta\sqrt{(4E^2+10E+6)c_3} = \frac{2}{\Tilde{\mu}} \sqrt{(4E^2+10E+6)c_3}= \frac{2}{\Tilde{\mu}} \sqrt{(4E^2+10E+6)(12\sigma^2+18 L^2(1+\epsilon_{\max})^2)}.
    \end{align*}
\end{proof}
\subsection{Deferred Proofs of Key Lemmas}
\begin{proof}[Proof of \ref{lem:partial_unbiased}]
    Let $\{x_i\}_{i=1}^N$ denote any fixed deterministic sequence. We sample a multiset $\mathcal{S}_t$ with $|\mathcal{S}_t| = K$ by the procedure where each sampling time, we sample $x_k$ with probability $q_k$ for each time. Note that two samples are not necessarily independent. We only require each sampling distribution is identical. Let $\mathcal{S}_t = \{i_1, \dots, i_K\} \subset [N]$ (some $i_k$'s may have the same value if sampling with replacement). Then
    \begin{equation*}
        \mathbb{E}_{\mathcal{S}_t} \sum_{k \in \mathcal{S}_t} x_k 
        = \mathbb{E}_{\mathcal{S}_t}  \sum_{k=1}^K x_{i_k} = K \mathbb{E}_{\mathcal{S}_t} x_{i_1} = K \sum_{k=1}^K q_k x_k.
    \end{equation*}
    For Scheme I, $q_k = p_k$ and for Scheme II, $q_k = \frac{1}{N}$, replacing the values into the above proves the lemma.
\end{proof}

\begin{proof}[Proof of Lemma \ref{lem:descent_partial}]
    When $t+1\notin \mathcal{I}_E$, because $\overline{\boldsymbol{\theta}}^{t+1} = \overline{\boldsymbol{w}}^{t+1}$ for both schemes, by Lemma \ref{lem:descent}, we got the conclusion. When $t+1\in \mathcal{I}_E$, we have
    \begin{align*}
        \mathbb{E}\| \overline{\boldsymbol{\theta}}^{t+1} - \boldsymbol{\theta}^{PS} \|_2^2 = &~ \mathbb{E}\|\overline{\boldsymbol{\theta}}^{t+1}- \overline{\boldsymbol{w}}^{t+1}\|_2^2 + \mathbb{E}\| \overline{\boldsymbol{w}}^{t+1} - \boldsymbol{\theta}^{PS} \|_2^2 + 2\mathbb{E} \langle\overline{\boldsymbol{\theta}}^{t+1}- \overline{\boldsymbol{w}}^{t+1}, \overline{\boldsymbol{w}}^{t+1} - \boldsymbol{\theta}^{PS}\rangle .
    \end{align*}
   By Lemma \ref{lem:partial_unbiased} and the law of total expectation, we have
   \begin{align*}
       \mathbb{E} \langle\overline{\boldsymbol{\theta}}^{t+1}- \overline{\boldsymbol{w}}^{t+1}, \overline{\boldsymbol{w}}^{t+1} - \boldsymbol{\theta}^{PS}\rangle = \mathbb{E} \big[\mathbb{E}_{\mathcal{S}_{t+1}}\langle\overline{\boldsymbol{\theta}}^{t+1}- \overline{\boldsymbol{w}}^{t+1}, \overline{\boldsymbol{w}}^{t+1} - \boldsymbol{\theta}^{PS}\rangle\big] =0.
   \end{align*}
   Next we focus on upper bounding $\mathbb{E} [\| \overline{\boldsymbol{\theta}}^{t+1} - \overline{\boldsymbol{w}}^{t+1} \|^2_2]$ under two sampling schemes.

   Denote $\mathcal{S}_{t+1}=\{i_1,\ldots, i_K\}$, then for scheme I, $\overline{\boldsymbol{\theta}}^{t+1}=\frac{1}{K} \sum_{l=1}^K \boldsymbol{w}^{t+1}_{i_l}$. Thus by the law of total expectation, we have 
   \begin{align*} 
        \mathbb{E}\| \overline{\boldsymbol{\theta}}^{t+1} - \overline{\boldsymbol{w}}^{t+1} \|_2^2=&~  \mathbb{E} \big[\mathbb{E}_{\mathcal{S}_{t+1}} \| \overline{\boldsymbol{\theta}}^{t+1} - \overline{\boldsymbol{w}}^{t+1} \|_2^2\big]\\
        =&~  \mathbb{E} \big[\mathbb{E}_{\mathcal{S}_{t+1}} \| \frac{1}{K} \sum_{l=1}^K \boldsymbol{w}^{t+1}_{i_l} - \overline{\boldsymbol{w}}^{t+1} \|_2^2\big]\\
        \le &~  \mathbb{E} \big[\mathbb{E}_{\mathcal{S}_{t+1}} \frac{1}{K^2} \sum_{l=1}^K \| \boldsymbol{w}_{i_l}^{t+1} - \overline{\boldsymbol{w}}^{t+1} \|_2^2\big]\\
        =&~ \frac{1}{K} \sum_{k=1}^N p_k \mathbb{E}\| \boldsymbol{w}_{k}^{t+1} - \overline{\boldsymbol{w}}^{t+1} \|_2^2.
    \end{align*}
    Again with $\overline{\boldsymbol{\theta}}^{t+1}=\frac{1}{K} \sum_{l=1}^K \boldsymbol{w}^{t+1}_{i_l}$, for scheme II, by the law of total expectation, we have
    \begin{align*}
        \mathbb{E}\| \overline{\boldsymbol{\theta}}^{t+1} - \overline{\boldsymbol{w}}^{t+1} \|_2^2=&~  \mathbb{E} \big[\mathbb{E}_{\mathcal{S}_{t+1}} \| \overline{\boldsymbol{\theta}}^{t+1} - \overline{\boldsymbol{w}}^{t+1} \|_2^2\big]\\
        = &~ \mathbb{E} \bigg[\mathbb{E}_{\mathcal{S}_{t+1}} \bigg[ \| \frac{1}{K} \sum_{l=1}^K \boldsymbol{w}_{i_l}^{t+1} - \overline{\boldsymbol{w}}^{t+1} \|_2^2 \bigg] \bigg] \\
        = &~ \mathbb{E} \bigg[\frac{1}{K^2} \mathbb{E}_{\mathcal{S}_{t+1}} \bigg[ \| \sum_{i=1}^N \boldsymbol{1}\{ i \in \mathcal{S}_{t+1} \}  (\boldsymbol{w}_{i}^{t+1} - \overline{\boldsymbol{w}}^{t+1}) \|_2^2 \bigg]\bigg]  \\
        \le &~ \frac{1}{K^2} \mathbb{E} \bigg[ \sum_{i=1}^N \mathbb{P}(i \in \mathcal{S}_{t+1}) \|\boldsymbol{w}_{i}^{t+1} - \overline{\boldsymbol{w}}^{t+1}\|_2^2 + \sum_{j \neq i} \mathbb{P}(i,j\in \mathcal{S}_{t+1}) \left \langle \boldsymbol{w}_{i}^{t+1} - \overline{\boldsymbol{w}}^{t+1}, ~ \boldsymbol{w}_{j}^{t+1} - \overline{\boldsymbol{w}}^{t+1} \right \rangle \bigg] \nonumber \\
        = &~ \frac{1}{KN} \sum_{i=1}^N \mathbb{E}\|\boldsymbol{w}_{i}^{t+1} - \overline{\boldsymbol{w}}^{t+1}\|_2^2 + \frac{K-1}{KN(N-1)} \sum_{i \neq j} \mathbb{E}\left \langle \boldsymbol{w}_{i}^{t+1} - \overline{\boldsymbol{w}}^{t+1}, ~ \boldsymbol{w}_{j}^{t+1} - \overline{\boldsymbol{w}}^{t+1} \right \rangle \nonumber \\
        = &~ \frac{N}{K(N-1)} \bigg(1 - \frac{K}{N}\bigg) \sum_{i=1}^N p_i\mathbb{E} \|\boldsymbol{w}_{i}^{t+1} - \overline{\boldsymbol{w}}^{t+1}\|_2^2
    \end{align*}
    where we use the following equalities: (1) $\mathbb{P}\left(i \in S_{t+1}\right)=\frac{K}{N}$ and $\mathbb{P}\left(i, j \in S_{t+1}\right)=\frac{K(K-1)}{N(N-1)}$ for all $i \neq j$ and (2) $\sum_{i=1}^N \left\|\boldsymbol{w}_{i}^{t} - \overline{\boldsymbol{w}}^{t}\right\|^2+\sum_{i \neq j}\langle\boldsymbol{w}_{i}^{t+1} - \overline{\boldsymbol{w}}^{t+1}, \boldsymbol{w}_{j}^{t+1} - \overline{\boldsymbol{w}}^{t+1}\rangle=0$.

The conclusion follows from the above discussion.
\end{proof}

\begin{proof}[Proof of Lemma \ref{lem:cleandescent_partial}]
In this proof, for convenience, we will discuss with respect to $t+1$ where we assume $t+1\notin \mathcal{I}_E$ and transfer back to $t$ in the last. First by the update rule, we have when $t+1 \notin \mathcal{I}_E$
\begin{align*}
    \boldsymbol{\theta}^{t+1}_i- \overline{\boldsymbol{\theta}}^{t+1} = \boldsymbol{\theta}_i^{t} 
 - \overline{\boldsymbol{\theta}}^{t} - \eta_{t} (\nabla \ell(\boldsymbol{\theta}_i^t; Z_i^{t+1})
-{g}_t)
\end{align*}
and when $t+1 \in \mathcal{I}_E$,
\begin{align*}
    \boldsymbol{w}^{t+1}_i- \overline{\boldsymbol{w}}^{t+1} = \boldsymbol{\theta}_i^{t} 
 - \overline{\boldsymbol{\theta}}^{t} - \eta_{t} (\nabla \ell(\boldsymbol{\theta}_i^t; Z_i^{t+1})
-{g}_t).
\end{align*}
Then with the same method in Lemma \ref{lem:consensuserr}, let $t_0 := \max\{s \mid s < t+1, s\in \mathcal{I}_E\} $ and $ c_3:= 12\sigma^2+18 L^2(1+\epsilon_{\max})^2$, if $\eta_t^2 \le \frac{1}{2c_3 (t+1-t_0)\big(1+2(t+1-t_0)\big)}$, we will have: if $t+1 \notin \mathcal{I}_E$,
\begin{align*}
     &~\sum_{i=1}^N p_i\mathbb{E}\|\boldsymbol{\theta}^{t+1}_i- \overline{\boldsymbol{\theta}}^{t+1}\|^2_2 \nonumber\\
     \le & \sum_{s=0}^{t-t_0} \frac{t+2-t_0}{s+2}  \eta_t^2 (2E+1) (48\sigma^2+36\varsigma^2)\mathbb{E}\|\overline{\boldsymbol{\theta}}^s-\boldsymbol{\theta}^{PS}\|_2^2 + \sum_{s=0}^{t-t_0} \frac{t+2-t_0}{s+2}   \eta_t^2 (2E+1) (24\sigma^2+36\varsigma^2)
\end{align*}
and if $t+1 \in \mathcal{I}_E$, 
\begin{align*}
     &~\sum_{i=1}^N p_i\mathbb{E}\|\boldsymbol{w}^{t+1}_i- \overline{\boldsymbol{w}}^{t+1}\|^2_2 \nonumber\\
     \le & \sum_{s=0}^{t-t_0} \frac{t+2-t_0}{s+2}  \eta_t^2 (2E+1) (48\sigma^2+36\varsigma^2)\mathbb{E}\|\overline{\boldsymbol{\theta}}^s-\boldsymbol{\theta}^{PS}\|_2^2 + \sum_{s=0}^{t-t_0} \frac{t+2-t_0}{s+2}   \eta_t^2 (2E+1) (24\sigma^2+36\varsigma^2).
\end{align*}

With the above formula and Lemma \ref{lem:descent_partial}, we now prove that if $\eta_0$ is sufficiently small, then for any $t$, we have $\mathbb{E}\|\overline{\boldsymbol{\theta}}^t-\boldsymbol{\theta}^{PS}\|_2^2 \le \mathbb{E}\|\overline{\boldsymbol{\theta}}^0-\boldsymbol{\theta}^{PS}\|_2^2$. We first derive the following inequality, which we will use later. Note that for any $t$ where $t_0 := \max\{s \mid s < t+1, s\in \mathcal{I}_E\} $, we have
\begin{align*}
    \sum_{s=0}^{t-t_0} \frac{t+2-t_0}{s+2}  = (t+2-t_0)(\frac{1}{2}+ \ldots+\frac{1}{t-t_0+2})\le (t+2-t_0)\log (t+2-t_0)\le (E+1)\log (E+1).
\end{align*}
Then again by the same induction method in Lemma \ref{lem:consensuserr}, we have if
\begin{align*}
    \eta_t\le&~ \eta_0 \nonumber\\
    \le &~ \frac{\Tilde{\mu} \mathbb{E}\|\overline{\boldsymbol{\theta}}^0-\boldsymbol{\theta}^{PS}\|_2^2}{2\sigma^2+( c_1c_3  + c_2/6 ) (2E^2+3E+1)\log (E+1)\big( (16\sigma^2+12\varsigma^2)\mathbb{E}\|\overline{\boldsymbol{\theta}}^0-\boldsymbol{\theta}^{PS}\|_2^2+ (8\sigma^2+12\varsigma^2)\big) }= \Tilde{\eta}_0,
\end{align*}
then for any $t$, we have $\mathbb{E}\|\overline{\boldsymbol{\theta}}^t-\boldsymbol{\theta}^{PS}\|_2^2 \le \mathbb{E}\|\overline{\boldsymbol{\theta}}^0-\boldsymbol{\theta}^{PS}\|_2^2$. 

Under all these conditions, if $t\notin \mathcal{I}_E$, we have
\begin{align*}
    \sum_{i=1}^N p_i\mathbb{E}\|\boldsymbol{\theta}^{t}_i- \overline{\boldsymbol{\theta}}^{t}\|^2_2 \le&~ \eta_{t-1}^2 (2E^2+3E+1)\log (E+1) (48\sigma^2+36\varsigma^2)\mathbb{E}\|\overline{\boldsymbol{\theta}}^0-\boldsymbol{\theta}^{PS}\|_2^2 \\
    &~+ \eta_{t-1}^2(2E^2+3E+1)\log (E+1) (24\sigma^2+36\varsigma^2)\\
    \le&~ 4\eta_{t}^2 (2E^2+3E+1)\log (E+1) (48\sigma^2+36\varsigma^2)\mathbb{E}\|\overline{\boldsymbol{\theta}}^0-\boldsymbol{\theta}^{PS}\|_2^2 \\
    &~+ 4\eta_{t}^2(2E^2+3E+1)\log (E+1) (24\sigma^2+36\varsigma^2),
\end{align*}
and if $t+1\in \mathcal{I}_E$, we have
\begin{align*}
    \sum_{i=1}^N p_i\mathbb{E}\|\boldsymbol{w}^{t+1}_i- \overline{\boldsymbol{w}}^{t+1}\|^2_2 \le&~ \eta_{t-1}^2 (2E^2+3E+1)\log (E+1) (48\sigma^2+36\varsigma^2)\mathbb{E}\|\overline{\boldsymbol{\theta}}^0-\boldsymbol{\theta}^{PS}\|_2^2 \\
    &~+ \eta_{t-1}^2(2E^2+3E+1)\log (E+1) (24\sigma^2+36\varsigma^2).
\end{align*}

Note that in Lemma \ref{lem:descent_partial}, the inequality for $t\notin \mathcal{I}_E$ is looser than the inequality for $t\in \mathcal{I}_E$. Therefore, we can apply the inequality for $t\in \mathcal{I}_E$ for all $t$. Combining this inequality with the above formula gives us that:
\begin{enumerate}
   \item for scheme I,
        \begin{align*}
        &~\mathbb{E} [\| \overline{\boldsymbol{\theta}}^{t+1} - \boldsymbol{\theta}^{PS} \|^2_2] \\
        \leq &~ \frac{1}{K} \sum_{k=1}^N p_k \mathbb{E}\| \boldsymbol{w}_{k}^{t+1} - \overline{\boldsymbol{w}}^{t+1} \|_2^2 + (1 - \Tilde{\mu} \eta_t) \mathbb{E}\|\overline{\boldsymbol{\theta}}^{t} - \boldsymbol{\theta}^{PS}\|_2^2 + 2 \sigma^2 \eta_t^2  + ( c_1 \eta_t + c_2 \eta_t^2 ) \sum_{i=1}^N p_i \mathbb{E}\| \boldsymbol{\theta}_i^t - \overline{\boldsymbol{\theta}}^{t} \|_2^2\\
        \le &~ (1 - \Tilde{\mu} \eta_t) \mathbb{E}\|\overline{\boldsymbol{\theta}}^{t} - \boldsymbol{\theta}^{PS}\|_2^2 + 2 \sigma^2 \eta_t^2 \\
        &~+ (4c_1 \eta_t + 4c_2 \eta_t^2 +K^{-1}) (2E^2+3E+1)\log (E+1) \eta_t^2 \bigg(  (48\sigma^2+36\varsigma^2)\mathbb{E}\|\overline{\boldsymbol{\theta}}^0-\boldsymbol{\theta}^{PS}\|_2^2 +(24\sigma^2+36\varsigma^2)\bigg),
    \end{align*}
      \item for scheme II, 
        \begin{align*}
        &~\mathbb{E} [\| \overline{\boldsymbol{\theta}}^{t+1} - \boldsymbol{\theta}^{PS} \|^2_2] \\
        \leq &~ \frac{N-K}{KN(N-1)} \sum_{k=1}^N p_k \mathbb{E}\| \boldsymbol{w}_{k}^{t+1} - \overline{\boldsymbol{w}}^{t+1} \|_2^2 + (1 - \Tilde{\mu} \eta_t) \mathbb{E}\|\overline{\boldsymbol{\theta}}^{t} - \boldsymbol{\theta}^{PS}\|_2^2 + 2 \sigma^2 \eta_t^2  + ( c_1 \eta_t + c_2 \eta_t^2 ) \sum_{i=1}^N p_i \mathbb{E}\| \boldsymbol{\theta}_i^t - \overline{\boldsymbol{\theta}}^{t} \|_2^2\\
        \le &~ (1 - \Tilde{\mu} \eta_t) \mathbb{E}\|\overline{\boldsymbol{\theta}}^{t} - \boldsymbol{\theta}^{PS}\|_2^2 + 2 \sigma^2 \eta_t^2 \\
        &~+ \bigg(4c_1 \eta_t + 4c_2 \eta_t^2 +\frac{N-K}{K(N-1)}\bigg ) (2E^2+3E+1)\log (E+1) \eta_t^2 \bigg(  (48\sigma^2+36\varsigma^2)\mathbb{E}\|\overline{\boldsymbol{\theta}}^0-\boldsymbol{\theta}^{PS}\|_2^2 +(24\sigma^2+36\varsigma^2)\bigg).
        \end{align*}
\end{enumerate}
\end{proof}

\begin{proof}[Proof of Lemma \ref{lem:cleandescent_partial}]
    The conclusion follows directly from Lemma \ref{lem:consensuserr_partial}.
\end{proof}

\section{Proof of convergence under the alternative assumption in \eqref{asm:uniform_gradnorm_bound}} \label{appensec:alternative}

\begin{assumption} \label{asm:appen_uniform_gradnorm_bound}
    Suppose the following hold
    \begin{equation}
        \mathbb{E}_{Z_i \sim \mathcal{D}_i(\boldsymbol{\theta})} [ \| \nabla \ell(\boldsymbol{\theta} ; Z_i) \|_2^2 ] \leq G^2.
    \end{equation}
\end{assumption}

\begin{lemma} (Bound on the divergence of parameters, i.e., consensus error bound)\label{lem:appen_consensus}\\
    When $E>1$, under Assumption \ref{asm:a1_obj_str_conv}, \ref{asm:a2_obj_smooth}, \ref{asm:a3_map_sensitivity}, \ref{asm:a4_bound_var_sto_grad}, and if $\eta_t$ is non-increasing and $\eta_{t} \leq 2 \eta_{t+E}$ holds for all $t \geq 0$, we have
     \begin{equation}
        \mathbb{E} \bigg[\sum_{i=1}^N p_i \|  \boldsymbol{\theta}_i^t - \overline{\boldsymbol{\theta}}^t\|_2^2 \bigg] \leq 4 (E-1)^2 \eta_{t}^2 G^2
    \end{equation}
\end{lemma}

\begin{proof}
    FedAvg requires a communication every $E$ steps, so for any $t \geq 0$, there exists a $t_0 \leq t, t_0 \in \mathcal{I}_E$, such that $t - t_0 \leq E-1$ and $\boldsymbol{\theta}^{t_0}_i = \overline{\boldsymbol{\theta}}^{t_0}, \forall i$. Also, we use the fact that $\eta_{t_0} \leq 2 \eta_t$ for all $t-t_0 \leq E-1$, then
    \begin{align}
        \mathbb{E} \bigg[\sum_{i=1}^N p_i \| \boldsymbol{\theta}_i^t -\overline{\boldsymbol{\theta}}^t\|_2^2 \bigg] 
        = \mathbb{E} \bigg[\sum_{i=1}^N p_i \| (\boldsymbol{\theta}_i^t -\overline{\boldsymbol{\theta}}^{t_0}) - (\overline{\boldsymbol{\theta}}^t - \overline{\boldsymbol{\theta}}^{t_0}) \|_2^2 \bigg] 
        \leq \mathbb{E} \bigg[\sum_{i=1}^N p_i \| \boldsymbol{\theta}_i^t -\overline{\boldsymbol{\theta}}^{t_0}\|_2^2 \bigg],
    \end{align}
    since $\mathbb{E} \|X - \mathbb{E}X\|_2^2 \leq \mathbb{E}\|X\|_2^2$ where $X=\boldsymbol{\theta}_i^t -\overline{\boldsymbol{\theta}}^{t_0}$. Using Jensen's inequality, we further have
    \begin{align}
        \| \boldsymbol{\theta}_i^t -\overline{\boldsymbol{\theta}}^{t_0}\|_2^2 
        = \bigg\| \sum_{s=t_0}^{t-1} \eta_{s} \nabla \ell(\boldsymbol{\theta}_i^{s}; Z_i^{s+1}) \bigg\|_2^2 
        \leq (t-t_0) \sum_{s=t_0}^{t-1} \eta_{s}^2 \big\| \nabla \ell(\boldsymbol{\theta}_i^{s}; Z_i^{s+1}) \big\|_2^2,
    \end{align}

    \begin{align}
        \mathbb{E} \bigg[\sum_{i=1}^N p_i \| \boldsymbol{\theta}_i^t -\overline{\boldsymbol{\theta}}^{t_0}\|_2^2 \bigg]
        = \mathbb{E} \bigg[\bigg\| \sum_{i=1}^N p_i \sum_{s=t_0}^{t-1} \eta_{s}^2 \nabla \ell(\boldsymbol{\theta}_i^{s}; Z_i^{s+1}) \bigg\|_2^2 \bigg]
        \leq (t-t_0) \sum_{s=t_0}^{t-1} \eta_{s}^2 \sum_{i=1}^N p_i  \mathbb{E} \bigg[ \big\| \nabla \ell(\boldsymbol{\theta}_i^{s}; Z_i^{s+1}) \big\|_2^2 \bigg],
    \end{align}
    
    \begin{align}
        \mathbb{E} \bigg[\sum_{i=1}^N p_i \| \boldsymbol{\theta}_i^t -\overline{\boldsymbol{\theta}}^{t_0}\|_2 \bigg]
        = \mathbb{E} \bigg[\bigg\| \sum_{i=1}^N p_i \sum_{s=t_0}^{t-1} \eta_{s} \nabla \ell(\boldsymbol{\theta}_i^{s}; Z_i^{s+1}) \bigg\|_2 \bigg]
        \leq \sum_{s=t_0}^{t-1} \eta_{s} \sum_{i=1}^N p_i  \mathbb{E} \bigg[ \big\| \nabla \ell(\boldsymbol{\theta}_i^{s}; Z_i^{s+1}) \big\|_2 \bigg],
    \end{align}
    where we used $\eta_{s} \leq \eta_{t_0}$. Therefore, based on \textbf{A5}, we have
    \begin{align} \label{eqn:bound_para_var_with_G}
        \mathbb{E} \bigg[\sum_{i=1}^N p_i \| \boldsymbol{\theta}_i^t -\overline{\boldsymbol{\theta}}^t\|_2^2 \bigg] 
        \leq &~ \sum_{i=1}^N p_i \mathbb{E} \bigg[ \sum_{s=t_0}^{t-1} (E-1) \eta_{s}^2 \| \nabla \ell(\boldsymbol{\theta}_i^{s}; Z_i^{s+1}) \|_2^2 \bigg] \nonumber \\
        \leq &~ \sum_{i=1}^N p_i \bigg[\sum_{s=t_0}^{t-1} (E-1) \eta_{s}^2 G^2 \bigg] \nonumber \\
        \leq &~ \sum_{i=1}^N p_i (E-1)^2 \eta_{t_0}^2 G^2  \nonumber \\
        \leq &~ 4 (E-1)^2 \eta_{t}^2 G^2
    \end{align}
    since $\eta_s \leq \eta_{t_0} \leq 2 \eta_{t_0+E} \leq 2\eta_{t}$ in the last two inequalities.
\end{proof}

\begin{lemma}\cite{ArXiv_2022_Li_MPP}\label{lem:a}
Consider a sequence of non-negative, non-increasing step sizes $\left\{\eta_t\right\}_{t \geq 1}$. Let $a>0, p \in \mathbb{Z}_{+}$and $\eta_1<2 / a$. If $\eta_t^p / \eta_{t+1}^p \leq 1+(a / 2) \eta_{t+1}^p$ for any $t \geq 1$, then
\begin{equation}
    \sum_{j=1}^t \eta_j^{p+1} \prod_{\ell=j+1}^t\left(1-\eta_{\ell} a\right) \leq \frac{2}{a} \eta_t^p, \quad \forall t \geq 1
\end{equation}
\end{lemma}

\begin{lemma} \label{lem:alt_full_convergence}
    Under Assumptions \ref{asm:a1_obj_str_conv}, \ref{asm:a2_obj_smooth}, \ref{asm:a3_map_sensitivity}, \ref{asm:appen_uniform_gradnorm_bound} and the condition that $\eta_t \leq \Tilde{\mu}/c_2$, , $\eta_t \leq \eta_{t_0} \leq 2 \eta_t$ where $t_0 = \max_{s} \{s\in \mathbb{N}| Es\leq t\}$, $\eta_{t+1} < \eta_t$ for any $t\geq0$, $\eta_1 < \frac{2}{\Tilde{\mu}}$ and $\eta_t^q / \eta_{t+1}^q \leq 1+(\Tilde{\mu} / 2) \eta_{t+1}^q$ for any $t \geq 0$ and $q=1,2,3$.
    
    \begin{equation}
        \mathbb{E} [\| \overline{\boldsymbol{w}}^{t+1} - \boldsymbol{\theta}^{PS} \|^2_2]  
        \leq \prod_{i=0}^t (1 - \Tilde{\mu} \eta_i)  \big\|\overline{\boldsymbol{\theta}}^{0} - \boldsymbol{\theta}^{PS}\big\|_2^2 +  \frac{2 c_2 c_7}{\Tilde{\mu}} \eta_t^3 + \frac{2 c_1 c_7}{\Tilde{\mu}} \eta_t^2 + \frac{4 \sigma^2}{\Tilde{\mu}} \eta_t,
    \end{equation}
    where $c_7 := 4(E-1) G^2$.
\end{lemma}

\begin{proof} 
From Lemma \ref{lem:descent}, we have
\begin{equation} \label{eqn:full_conv_thm_deriv}
\begin{aligned} 
        \mathbb{E} \big[\| \overline{\boldsymbol{w}}^{t+1} - \boldsymbol{\theta}^{PS} \|^2_2\big]  
        &\leq (1 - \Tilde{\mu} \eta_t) \mathbb{E} \bigg[\big\|\overline{\boldsymbol{\theta}}^{t} - \boldsymbol{\theta}^{PS}\big\|_2^2\bigg] + ( c_1 \eta_t + c_2 \eta_t^2 ) \mathbb{E} \bigg[\sum_{i=1}^N p_i \big\| \boldsymbol{\theta}_i^t - \overline{\boldsymbol{\theta}}^{t} \big\|_2^2\bigg] + 2 \sigma^2 \eta_t^2\\
        &\leq (1 - \Tilde{\mu} \eta_t) \mathbb{E} \bigg[\big\|\overline{\boldsymbol{\theta}}^{t} - \boldsymbol{\theta}^{PS}\big\|_2^2\bigg] + c_2 c_7 \eta_t^4 + c_1 c_7 \eta_t^3 + 2 \sigma^2 \eta_t^2\\
        &= \prod_{i=0}^{t} (1 - \Tilde{\mu} \eta_i) \big\|\overline{\boldsymbol{\theta}}^{0} - \boldsymbol{\theta}^{PS}\big\|_2^2 + \sum_{s=1}^{t}\prod_{i=s+1}^{t} (1 - \Tilde{\mu} \eta_i)\Big( c_2 c_7 \eta_s^4 + c_1 c_7 \eta_s^3 + 2 \sigma^2 \eta_s^2 \Big).
\end{aligned}
\end{equation}
The second inequality holds because of Lemma \ref{lem:appen_consensus}. Using Lemma \ref{lem:a},
\begin{equation} \label{eqn:full_conv_thm_res}
\begin{aligned}
    \sum_{s=1}^{t}\prod_{i=s+1}^{t} (1 - \Tilde{\mu} \eta_i)\Big( c_2 c_7 \eta_s^4 + c_1 c_7 \eta_s^3 + 2 \sigma^2 \eta_s^2 \Big) \leq \frac{2 c_2 c_7}{\Tilde{\mu}} \eta_t^3 + \frac{2 c_1 c_7}{\Tilde{\mu}} \eta_t^2 + \frac{4 \sigma^2}{\Tilde{\mu}} \eta_t.
\end{aligned}
\end{equation}
\end{proof}

\begin{theorem} (Full participation convergence theorem, alternative assumption)\\
    Under Assumption \ref{asm:a1_obj_str_conv}, \ref{asm:a2_obj_smooth}, \ref{asm:a3_map_sensitivity}, \ref{asm:a4_bound_var_sto_grad}, the full participation scheme has convergence rate $\mathcal{O}(\frac{1}{T})$, i.e., denote $\Delta_t := \mathbb{E} [\| \overline{\boldsymbol{\theta}}^{t} - \boldsymbol{\theta}^{PS} \|^2_2]$, then for some $\gamma > 0$,
    \begin{equation}
        \Delta_t \leq \frac{\upsilon}{\gamma + t},
    \end{equation}
    where $\upsilon := \max \{ \frac{c_2 c_7 \beta^4 \gamma^{-2} + c_1 c_7 \beta^3 \gamma^{-1} + 2 \sigma^2 \beta^2}{\beta \mu - 1} , (\gamma + 1) \Delta_1 \}$.
\end{theorem}

\begin{proof}
    We will show it on the partial participation algorithm, and the proof for the full participation is similar. 
    
    For a diminishing step size $\eta_t = \frac{\beta}{t + \gamma}$ for some $\beta > \frac{1}{\Tilde{\mu}}$ and $\gamma >0$ such that $\eta_1 \leq \min\{\frac{1}{\Tilde{\mu}}, \frac{1}{4L}\} = \frac{1}{4L}$ and $\eta_t \leq 2 \eta_{t+E}$. We will prove $\Delta_t := \mathbb{E} [\| \overline{\boldsymbol{\theta}}^{t} - \boldsymbol{\theta}^{PS} \|^2_2] \leq \frac{v}{\gamma + t}$, where $v := \max \{ \frac{c_2 c_7 \beta^4 \gamma^{-2} + c_1 c_7 \beta^3 \gamma^{-1} + 2 \sigma^2 \beta^2}{\beta \mu - 1} , (\gamma + 1) \Delta_1 \}$.
    We prove this by induction. Firstly, the definition if $v$ ensures it holds for $t=1$. Assume it holds for some $t$, i.e., $\eta_t = \frac{\beta}{t + \gamma}$, then it follows from Lemma \ref{lem:alt_full_convergence} that 
    \begin{align}
        \Delta_{t+1} \leq &~ (1 - \eta_t \Tilde{\mu}) \Delta_t + c_2 c_7 \eta_t^4 + c_1 c_7 \eta_t^3 + 2 \sigma^2 \eta_t^2 \nonumber \\
        \leq &~ (1 - \frac{\beta \Tilde{\mu}}{t + \gamma})\frac{v}{t+\gamma} + \frac{c_2 c_7 \beta^4}{(t + \gamma)^4} + \frac{c_1 c_7 \beta^3}{(t + \gamma)^3} + \frac{2 \sigma^2 \beta^2}{(t + \gamma)^2} \nonumber \\
        = &~ \frac{t + \gamma -1}{(t + \gamma)^2}v + \left[ \frac{c_2 c_7 \beta^4}{(t + \gamma)^4} + \frac{c_1 c_7 \beta^3}{(t + \gamma)^3} + \frac{2 \sigma^2 \beta^2}{(t + \gamma)^2} - \frac{\beta \mu - 1}{(t + \gamma)^2}v \right] \nonumber \\
        \leq &~ \frac{t + \gamma -1}{(t + \gamma)^2}v + \left[ \frac{c_2 c_7 \beta^4}{(t + \gamma)^2 \gamma^2} + \frac{c_1 c_7 \beta^3}{(t + \gamma)^2 \gamma} + \frac{2 \sigma^2 \beta^2}{(t + \gamma)^2} - \frac{\beta \mu - 1}{(t + \gamma)^2}v \right] \nonumber \\
        \leq &~ \frac{v}{t + \gamma + 1}
    \end{align}
    where $\Tilde{\mu}$, $c_1, c_2, c_3, c_7$ are defined the same as in earlier proofs, and thus the $\mathcal{O}(1/T)$ convergence rate is shown. 
\end{proof}

\begin{lemma} \label{lem:appen_partial_diff_bound}(Bounding the difference $~\overline{\boldsymbol{w}}^{t+1} - \overline{\boldsymbol{\theta}}^{t+1}$ in partial participation)\\
    Suppose Assumption \ref{asm:a1_obj_str_conv}, \ref{asm:a2_obj_smooth}, \ref{asm:a3_map_sensitivity}, and \ref{asm:a4_bound_var_sto_grad} hold.
    For $t+1 \in \mathcal{I}_E$, assume that $\eta_t$ is non-increasing and $\eta_t \leq 2 \eta_{t+E}$ for all $t$, then we have the following results
    \begin{enumerate}
        \item For Scheme I, the expected difference $\overline{\boldsymbol{w}}^{t+1} - \overline{\boldsymbol{\theta}}^{t+1}$ is bounded by
        \begin{equation}
            \mathbb{E}_{\mathcal{S}_{t}} \| \overline{\boldsymbol{\theta}}^{t+1} - \overline{\boldsymbol{w}}^{t+1} \|_2^2 \leq \frac{4}{K} \eta_t^2 E^2 G^2.
        \end{equation}

        \item For Scheme II, assuming $p_1 = p_2 = \cdots = p_N = \frac{1}{N}$, the expected difference $\overline{\boldsymbol{w}}^{t+1} - \overline{\boldsymbol{\theta}}^{t+1}$ is bounded by
        \begin{equation}
            \mathbb{E}_{\mathcal{S}_{t}} \| \overline{\boldsymbol{\theta}}^{t+1} - \overline{\boldsymbol{w}}^{t+1} \|_2^2 \leq \frac{4(N-K)}{K(N-1)}\eta_t^2 E^2 G^2
        \end{equation}
    \end{enumerate}
\end{lemma}

\begin{proof}
    We prove the bound for Scheme I as follows. Since $\overline{\boldsymbol{\theta}}^{t+1} = \frac{1}{K} \sum_{l=1}^K \boldsymbol{w}_{i_l}^t$, taking expectation over $\mathcal{S}_{t+1}$, we have
    \begin{equation} \label{eqn:partial_lem2_scheme1}
        \mathbb{E}_{\mathcal{S}_{t}} \| \overline{\boldsymbol{\theta}}^{t+1} - \overline{\boldsymbol{w}}^{t+1} \|_2^2 = \mathbb{E}_{\mathcal{S}_{t}} \frac{1}{K^2} \sum_{l=1}^K \| \boldsymbol{w}_{i_l}^{t+1} - \overline{\boldsymbol{w}}^{t+1} \|_2^2 = \frac{1}{K} \sum_{k=1}^N p_k \| \boldsymbol{w}_{k}^{t+1} - \overline{\boldsymbol{w}}^{t+1} \|_2^2
    \end{equation}
    We note that since $t+1 \in \mathcal{I}_E$, we know that the time $t_0 = t - E + 1 \in \mathcal{I}_E$ is the communication time, which implies $\{\boldsymbol{\theta}^k_{t_0}\}$ is identical. Then
    \begin{align}
        \sum_{k=1}^N p_k \| \boldsymbol{w}_{k}^{t+1} - \overline{\boldsymbol{w}}^{t+1} \|_2^2 
        = &~ \sum_{i=1}^N p_k \| (\boldsymbol{w}_{k}^{t+1} - \overline{\boldsymbol{\theta}}^{t_0}) - (\overline{\boldsymbol{w}}^{t+1} - \overline{\boldsymbol{\theta}}^{t_0}) \|_2^2 \leq \sum_{i=1}^N p_k \| \boldsymbol{w}_{k}^{t+1} - \overline{\boldsymbol{\theta}}^{t_0} \|_2^2
    \end{align}
    Similar to Lemma \ref{lem:appen_consensus}, the last inequality is due to $\mathbb{E}\|X - \mathbb{E}X\|_2^2 \leq \mathbb{E}\|X\|_2^2$ where $X = \boldsymbol{w}_{k}^{t+1} - \overline{\boldsymbol{\theta}}^{t_0}$, and $\sum_{k=1}^N p_k (\boldsymbol{w}_{k}^{t+1} - \overline{\boldsymbol{\theta}}^{t_0}) = \overline{\boldsymbol{w}}^{t+1} - \overline{\boldsymbol{\theta}}^{t_0}$. Similarly, we have
    \begin{align}
        \mathbb{E}_{\mathcal{S}_{t}} \big[ \| \overline{\boldsymbol{\theta}}^{t+1} - \overline{\boldsymbol{w}}^{t+1} \|_2^2 \big]
        \leq &~ \frac{1}{K} \sum_{k=1}^N p_k \mathbb{E} \big[\|\boldsymbol{w}_{k}^{t+1} - \overline{\boldsymbol{\theta}}^{t_0}\|_2^2\big] \nonumber \\
        = &~ \frac{1}{K} \sum_{k=1}^N p_k \mathbb{E} \big[\|\boldsymbol{w}_{k}^{t+1} - \boldsymbol{\theta}_k^{t_0}\|_2^2 \big]\nonumber \\
        = &~ \frac{1}{K} \sum_{k=1}^N p_k \mathbb{E} \bigg[ \big\|\sum_{s=t_0}^t \eta_s \triangledown l (\boldsymbol{\theta}_k^{s}; Z_k^{s+1})\big\|_2^2 \bigg] \nonumber \\
        \leq &~ \frac{1}{K} \sum_{k=1}^N p_k E \sum_{s=t_0}^t \mathbb{E} \big[\| \eta_s \triangledown l (\boldsymbol{\theta}_k^{s}; Z_k^{s+1}) \|_2^2 \big] \nonumber \\
        \leq &~ \frac{1}{K} E^2 \eta_{t_0}^2 G^2 \leq \frac{4}{K} \eta_t^2 E^2 G^2.
    \end{align}

    Then we prove the bound for Scheme II. Since $\overline{\boldsymbol{\theta}}^{t+1} = \frac{1}{K} \sum_{l=1}^K  \boldsymbol{w}_{i_l}^{t+1}$, we have
    \begin{align}
        &~ \mathbb{E}_{\mathcal{S}_{t}} \big[ \| \overline{\boldsymbol{\theta}}^{t+1} - \overline{\boldsymbol{w}}^{t+1} \|_2^2 \big] \nonumber \\
        = &~ \mathbb{E}_{\mathcal{S}_{t}} \bigg[ \big\| \frac{1}{K} \sum_{l=1}^K \boldsymbol{w}_{i_l}^{t+1} - \overline{\boldsymbol{w}}^{t+1} \big\|_2^2 \bigg] \nonumber \\
        = &~ \frac{1}{K^2} \mathbb{E}_{\mathcal{S}_{t}} \bigg[ \big\| \sum_{i=1}^N \boldsymbol{1}\{ i \in \mathcal{S}_t \}  (\boldsymbol{w}_{i}^{t+1} - \overline{\boldsymbol{w}}^{t+1}) \big\|_2^2 \bigg] \nonumber \\
        = &~ \frac{1}{K^2} \bigg[ \sum_{i=1}^N \mathbb{P}(i \in \mathcal{S}_t) \|\boldsymbol{w}_{i}^{t+1} - \overline{\boldsymbol{w}}^{t+1}\|_2^2 + \sum_{j \neq i} \mathbb{P}(i,j\in \mathcal{S}_t) \left \langle \boldsymbol{w}_{i}^{t+1} - \overline{\boldsymbol{w}}^{t+1}, ~ \boldsymbol{w}_{j}^{t+1} - \overline{\boldsymbol{w}}^{t+1} \right \rangle \bigg] \nonumber \\
        = &~ \frac{1}{KN} \sum_{i=1}^N \|\boldsymbol{w}_{i}^{t+1} - \overline{\boldsymbol{w}}^{t+1}\|_2^2 + \frac{K-1}{KN(N-1)} \sum_{i \neq j} \left \langle \boldsymbol{w}_{i}^{t+1} - \overline{\boldsymbol{w}}^{t+1}, ~ \boldsymbol{w}_{j}^{t+1} - \overline{\boldsymbol{w}}^{t+1} \right \rangle \nonumber \\
        = &~ \frac{1}{K(N-1)} (1 - \frac{K}{N}) \sum_{i=1}^N \|\boldsymbol{w}_{i}^{t+1} - \overline{\boldsymbol{w}}^{t+1}\|_2^2.
    \end{align}
    Note that the second last equality holds because $\mathbb{P}(i \in \mathcal{S}_t) = \frac{K}{N}$ and $\mathbb{P}(i,j\in \mathcal{S}_t) = \frac{K(K-1)}{N(N-1)}$; and the last equality holds because
    \begin{align*}
        &~ \sum_{i=1}^N \|\boldsymbol{w}_{i}^{t+1} - \overline{\boldsymbol{w}}^{t+1}\|_2^2 + \sum_{i \neq j} \left \langle \boldsymbol{w}_{i}^{t+1} - \overline{\boldsymbol{w}}^{t+1}, ~ \boldsymbol{w}_{j}^{t+1} - \overline{\boldsymbol{w}}^{t+1} \right \rangle \\
        = &~ \sum_{i=1}^N \left \langle \boldsymbol{w}_{i}^{t+1} - \overline{\boldsymbol{w}}^{t+1}, ~ \bigg(\sum_{j=1}^N \boldsymbol{w}_{j}^{t+1}\bigg) - N \overline{\boldsymbol{w}}^{t+1} \right \rangle = 0.
    \end{align*}
    Recall that 
    \begin{align*}
        \sum_{k=1}^N p_k \| \boldsymbol{w}_{k}^{t+1} - \overline{\boldsymbol{w}}^{t+1} \|_2^2 
        \leq \sum_{i=1}^N p_k \| \boldsymbol{w}_{k}^{t+1} - \overline{\boldsymbol{\theta}}^{t_0} \|_2^2,
    \end{align*}
    we get
    \begin{align}
        \mathbb{E} \big[ \| \overline{\boldsymbol{\theta}}^{t+1} - \overline{\boldsymbol{w}}^{t+1} \|_2^2 \big] 
        = &~ \frac{1}{K(N-1)} (1 - \frac{K}{N}) \mathbb{E} \bigg[ \sum_{i=1}^N \|\boldsymbol{w}_{i}^{t+1} - \overline{\boldsymbol{w}}^{t+1}\|_2^2 \bigg] \nonumber \\
        \leq &~ \frac{N}{K(N-1)} (1 - \frac{K}{N}) \mathbb{E} \bigg[\sum_{i=1}^N \frac{1}{N} \|\boldsymbol{w}_{i}^{t+1} - \overline{\boldsymbol{\theta}}^{t_0}\|_2^2 \bigg] \nonumber \\
        \leq &~ \frac{N}{K(N-1)} (1 - \frac{K}{N}) 4 \eta_t^2 E^2 G^2 = \frac{4(N-K)}{K(N-1)}\eta_t^2 E^2 G^2
    \end{align}
    where the last inequality can be found in \eqref{eqn:bound_para_var_with_G} in the proof of Lemma \ref{lem:appen_consensus}.
\end{proof}

\begin{lemma} \label{lem:alt_partial_convergence}
    Under Under Assumption \ref{asm:a1_obj_str_conv}, \ref{asm:a2_obj_smooth}, \ref{asm:a3_map_sensitivity}, \ref{asm:a4_bound_var_sto_grad}, and the condition that $\eta_t \leq \Tilde{\mu}/c_2$, , $\eta_t \leq \eta_{t_0} \leq 2 \eta_t$ where $t_0 = \max_{s} \{s\in \mathbb{N}| Es\leq t\}$, $\eta_{t+1} < \eta_t$ for any $t\geq0$, $\eta_1 < \frac{2}{\Tilde{\mu}}$ and $\eta_t^q / \eta_{t+1}^q \leq 1+(\Tilde{\mu} / 2) \eta_{t+1}^q$ for any $t \geq 0$ and $q=1,2,3$, we have
    \begin{equation}
        \mathbb{E}_{\mathcal{S}_t} [|| \overline{\boldsymbol{\theta}}^{t+1} - \boldsymbol{\theta}^{PS} ||^2_2]  
        \leq \prod_{i=0}^t (1 - \Tilde{\mu} \eta_i)  \big|\big|\overline{\boldsymbol{\theta}}^{0} - \boldsymbol{\theta}^{PS}\big|\big|_2^2 +  \frac{2 c_2 c_3}{\Tilde{\mu}} \eta_t^3 + \frac{2 c_1 c_3}{\Tilde{\mu}} \eta_t^2 + \frac{2 c_8}{\Tilde{\mu}} \eta_t,
    \end{equation}
    ($c_8$ for Scheme I, replace $c_8$ with $c_9$ in Scheme II)
    where we define $c_8 := 2 \sigma^2 +\frac{4}{K} E^2 G^2$ in Scheme I, and $c_9 := 2 \sigma^2 +\frac{4(N-K)}{K(N-1)} E^2 G^2$ in Scheme II.
\end{lemma}

\begin{proof}
    Note that 
    \begin{align}
        &~ || \overline{\boldsymbol{\theta}}^{t+1} - \boldsymbol{\theta}^{PS} ||_2^2 \nonumber \\
        = &~ || \overline{\boldsymbol{\theta}}^{t+1} - \overline{\boldsymbol{w}}^{t+1} + \overline{\boldsymbol{w}}^{t+1} -  \boldsymbol{\theta}^{PS}||_2^2 \nonumber \\
        = &~ \underbrace{|| \overline{\boldsymbol{\theta}}^{t+1} - \overline{\boldsymbol{w}}^{t+1} ||_2^2}_{T_1}
        + \underbrace{|| \overline{\boldsymbol{w}}^{t+1} - \boldsymbol{\theta}^{PS} ||_2^2}_{T_2} + \underbrace{2 \langle \overline{\boldsymbol{w}}^{t+1} - \overline{\boldsymbol{\theta}}^{t+1},~ \overline{\boldsymbol{\theta}}^{t+1} - \boldsymbol{\theta}^{PS} \rangle }_{T_3}
    \end{align}
    When expectation is taken over $\mathcal{S}_{t+1}$, the last term $T_3$ vanishes due to Lemma \ref{lem:partial_unbiased}. 
    
    If $t+1 \notin \mathcal{I}_E$, $T_1$ vanishes since $\overline{\boldsymbol{\theta}}^{t+1} = \overline{\boldsymbol{w}}^{t+1}$ by definition when $t+1$ is not a communication step. For term $T_2$, it's not hard to see that we can use Lemma \ref{lem:descent} to derive one step bounds for it (and use \eqref{eqn:full_conv_thm_deriv} in Lemma \ref{lem:alt_full_convergence}), and thus we have
    \begin{align}
        \mathbb{E} \big[ || \overline{\boldsymbol{\theta}}^{t+1} - \boldsymbol{\theta}^{PS} ||^2_2 \big] 
        = &~ \mathbb{E} \big[ ||\overline{\boldsymbol{w}}^{t+1} - \boldsymbol{\theta}^{PS}||_2^2 \big] \nonumber \\
        \leq &~ (1 - \Tilde{\mu} \eta_t) \mathbb{E} \big|\big|\overline{\boldsymbol{\theta}}^{t} - \boldsymbol{\theta}^{PS}\big|\big|_2^2 + c_2 c_7 \eta_t^4 + c_1 c_7 \eta_t^3 + 2 \sigma^2 \eta_t^2,
    \end{align}
    and we recall that $c_1 := \frac{L^2(1+\epsilon_{max})^2}{2 \delta \overline{\epsilon}}, c_2 := 4 [\sigma^2 + L^2(1+\epsilon_{max})^2], c3:= 4(E-1)^2G^2, \Tilde{\mu} := \mu - (1+\delta) \overline{\epsilon} L$.

    If $t+1 \in \mathcal{I}_E$, then we have the following result from Lemma \ref{lem:appen_partial_diff_bound},
    \begin{align} \label{eqn:partial_conv_thm_res}
        &~ \mathbb{E} \big[ || \overline{\boldsymbol{\theta}}^{t+1} - \boldsymbol{\theta}^{PS} ||^2_2 \big] \nonumber \\
        = &~ \mathbb{E} \big[ || \overline{\boldsymbol{\theta}}^{t+1} -\overline{\boldsymbol{w}}^{t+1} ||_2^2 \big] + \mathbb{E} \big[ ||\overline{\boldsymbol{w}}^{t+1} - \boldsymbol{\theta}^{PS} ||_2^2 \big] \nonumber \\
        \leq &~ (1 - \Tilde{\mu} \eta_t) \mathbb{E} \big|\big|\overline{\boldsymbol{\theta}}^{t} - \boldsymbol{\theta}^{PS}\big|\big|_2^2 + c_2 c_7 \eta_t^4 + c_1 c_7 \eta_t^3 + c_8 \eta_t^2 ,
    \end{align}
    where we recall $c_8 := 2 \sigma^2 +\frac{4}{K} E^2 G^2$ in Scheme I, and $c_9 := 2 \sigma^2 +\frac{4(N-K)}{K(N-1)} E^2 G^2$ in Scheme II.

    The only difference between \eqref{eqn:full_conv_thm_res} and \eqref{eqn:partial_conv_thm_res} is in $(c_8 - 2 \sigma^2) \eta_t^2$. Therefore, we can use similar techniques to show the convergence, 
    \begin{equation*}
    \begin{aligned} 
        \mathbb{E} [|| \overline{\boldsymbol{\theta}}^{t+1} - \boldsymbol{\theta}^{PS} ||^2_2]  
        &\leq (1 - \Tilde{\mu} \eta_t) \mathbb{E} \big|\big|\overline{\boldsymbol{\theta}}^{t} - \boldsymbol{\theta}^{PS}\big|\big|_2^2 + c_2 c_7 \eta_t^4 + c_1 c_7 \eta_t^3 + c_8 \eta_t^2\\
        &= \prod_{i=0}^{t} (1 - \Tilde{\mu} \eta_i) \big|\big|\overline{\boldsymbol{\theta}}^{0} - \boldsymbol{\theta}^{PS}\big|\big|_2^2 \\
        &~~~~+ \sum_{s=1}^{t}\prod_{i=s+1}^{t} (1 - \Tilde{\mu} \eta_i)\Big( c_2 c_7 \eta_s^4 + c_1 c_7 \eta_s^3 + c_8 \eta_s^2 \Big) \\
        &\leq \prod_{i=0}^t (1 - \Tilde{\mu} \eta_i)  \big|\big|\overline{\boldsymbol{\theta}}^{0} - \boldsymbol{\theta}^{PS}\big|\big|_2^2 +  \frac{2 c_2 c_7}{\Tilde{\mu}} \eta_t^3 + \frac{2 c_1 c_7}{\Tilde{\mu}} \eta_t^2 + \frac{2 c_8}{\Tilde{\mu}} \eta_t
    \end{aligned}
    \end{equation*}
    ($c_8$ for Scheme I, replace $c_8$ with $c_9$ in Scheme II).
\end{proof}

\begin{theorem}
    (Full participation convergence theorem, alternative assumption)\\
    Under Assumption \ref{asm:a1_obj_str_conv}, \ref{asm:a2_obj_smooth}, \ref{asm:a3_map_sensitivity}, \ref{asm:a4_bound_var_sto_grad}, the full participation scheme has convergence rate $\mathcal{O}(\frac{1}{T})$, i.e., denote $\Delta_t := \mathbb{E} [\| \overline{\boldsymbol{\theta}}^{t} - \boldsymbol{\theta}^{PS} \|^2_2]$, then for some $\gamma > 0$,
    \begin{equation}
        \Delta_t \leq \frac{\upsilon}{\gamma + t},
    \end{equation}
    where $\upsilon := \max \{ \frac{c_2 c_7 \beta^4 \gamma^{-2} + c_1 c_7 \beta^3 \gamma^{-1} + c_8 \beta^2}{\beta \mu - 1} , (\gamma + 1) \Delta_1 \}$ ($c_8$ for Scheme I, replace $c_8$ with $c_9$ in Scheme II).
\end{theorem}

\begin{proof}
    We will show it on the partial participation algorithm, and the proof for the full participation is similar. 
    
    For a diminishing step size $\eta_t = \frac{\beta}{t + \gamma}$ for some $\beta > \frac{1}{\Tilde{\mu}}$ and $\gamma >0$ such that $\eta_1 \leq \min\{\frac{1}{\Tilde{\mu}}, \frac{1}{4L}\} = \frac{1}{4L}$ and $\eta_t \leq 2 \eta_{t+E}$. We will prove $\triangle_t := \mathbb{E} [|| \overline{\boldsymbol{\theta}}^{t} - \boldsymbol{\theta}^{PS} ||^2_2] \leq \frac{v}{\gamma + t}$, where $v := \max \{ \frac{c_2 c_7 \beta^4 \gamma^{-2} + c_1 c_7 \beta^3 \gamma^{-1} + c_8 \beta^2}{\beta \mu - 1} , (\gamma + 1) \triangle_1 \}$.
    We prove this by induction. Firstly, the definition if $v$ ensures it holds for $t=1$. Assume it holds for some $t$, i.e., $\eta_t = \frac{\beta}{t + \gamma}$, then it follows that 
    \begin{align}
        \triangle_{t+1} \leq &~ (1 - \eta_t \Tilde{\mu}) \triangle_t + c_2 c_7 \eta_t^4 + c_1 c_7 \eta_t^3 + c_8 \eta_t^2 \nonumber \\
        \leq &~ (1 - \frac{\beta \Tilde{\mu}}{t + \gamma})\frac{v}{t+\gamma} + \frac{c_2 c_7 \beta^4}{(t + \gamma)^4} + \frac{c_1 c_7 \beta^3}{(t + \gamma)^3} + \frac{c_8 \beta^2}{(t + \gamma)^2} \nonumber \\
        = &~ \frac{t + \gamma -1}{(t + \gamma)^2}v + \left[ \frac{c_2 c_7 \beta^4}{(t + \gamma)^4} + \frac{c_1 c_7 \beta^3}{(t + \gamma)^3} + \frac{c_8 \beta^2}{(t + \gamma)^2} - \frac{\beta \mu - 1}{(t + \gamma)^2}v \right] \nonumber \\
        \leq &~ \frac{t + \gamma -1}{(t + \gamma)^2}v + \left[ \frac{c_2 c_7 \beta^4}{(t + \gamma)^2 \gamma^2} + \frac{c_1 c_7 \beta^3}{(t + \gamma)^2 \gamma} + \frac{c_8 \beta^2}{(t + \gamma)^2} - \frac{\beta \mu - 1}{(t + \gamma)^2}v \right] \nonumber \\
        \leq &~ \frac{v}{t + \gamma + 1}
    \end{align}
    ($c_8$ for Scheme I, replace $c_8$ with $c_9$ in Scheme II), where $\Tilde{\mu}$, $c_1$ to $c_9$ are defined the same as in earlier proofs, and thus the $\mathcal{O}(1/T)$ convergence rate is shown.
\end{proof}

\newpage
\section{Experiments} \label{appensec:exp}

\subsection{Numerical simulation}
\begin{figure}[h]
    \centering
    \includegraphics[width=0.50\textwidth]{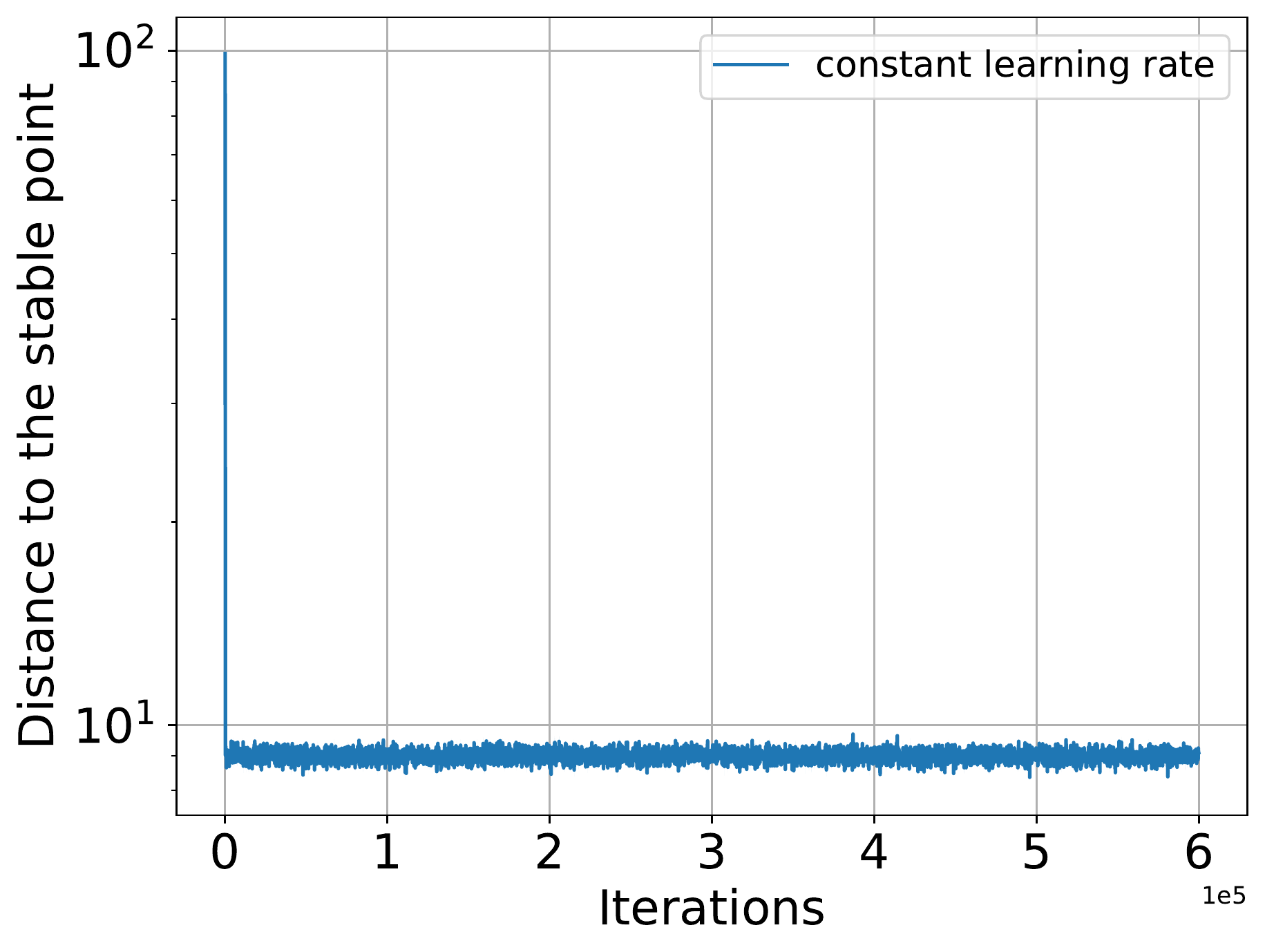}
    \caption{Constant learning rate on \expg. Full participation with $E=10$. The learning rate is set to $0.02$.}
    \label{fig:learningrate}
\end{figure}

\subsubsection{Learning rate decay (\cref{fig:learningrate})}\label{app:decay}
\begin{figure}[h]
    \centering
    \includegraphics[width=0.50\textwidth]{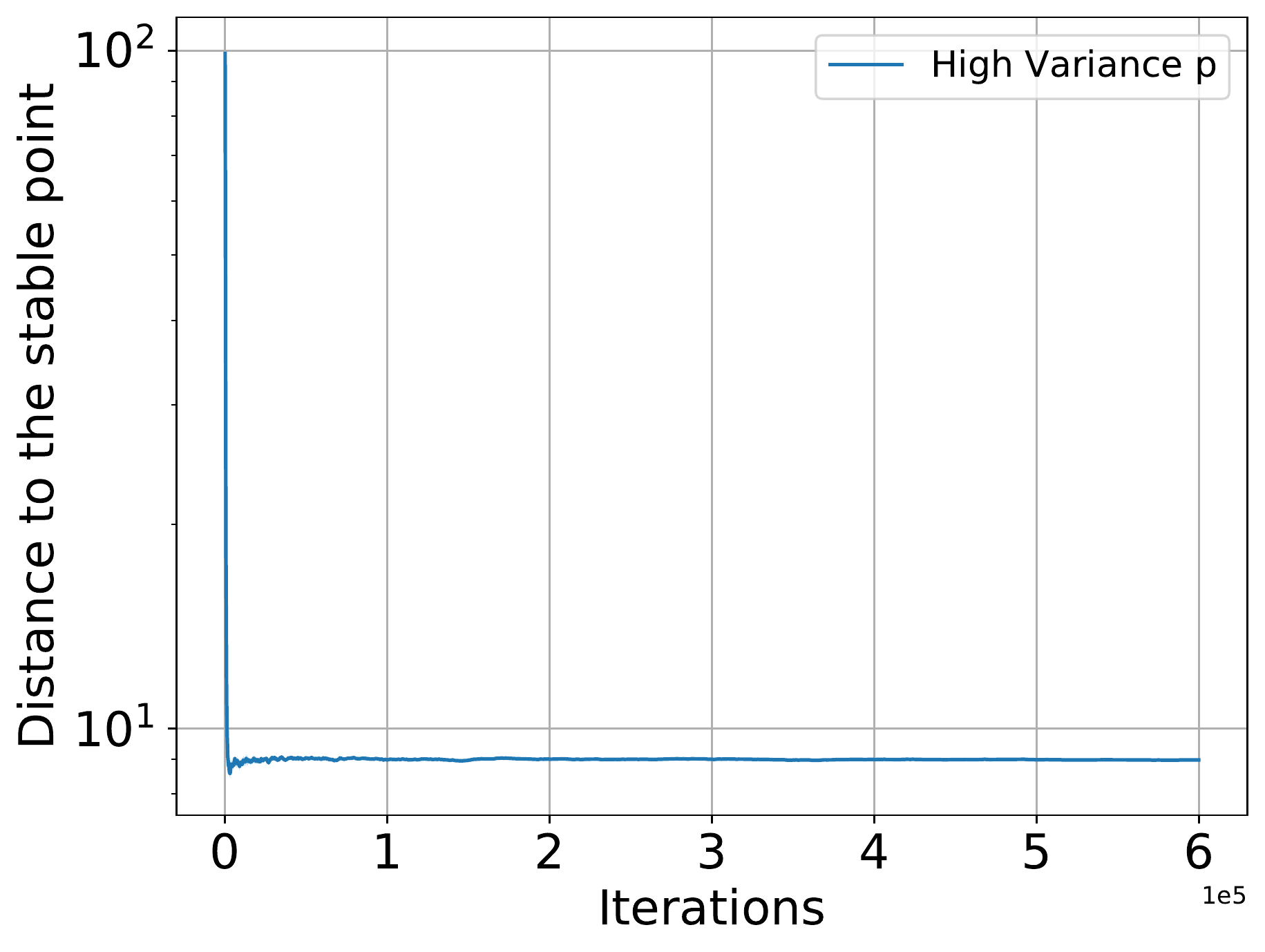}
    \caption{Sampling without replacement on \expg. The variance of $p_i$ is set to $0.01$. The learning rate is set to $\frac{5}{t + 10000}$.}
    \label{fig:highvarp}
\end{figure}

\subsubsection{Scheme II with lower learning rate (\cref{fig:highvarp})}\label{app:without}

\subsection{Credit score strategic classification}  \label{appensec:exp_credit}
\begin{figure}[htbp]
    \centering
    \subfigure[]{\includegraphics[height=0.25\textwidth]{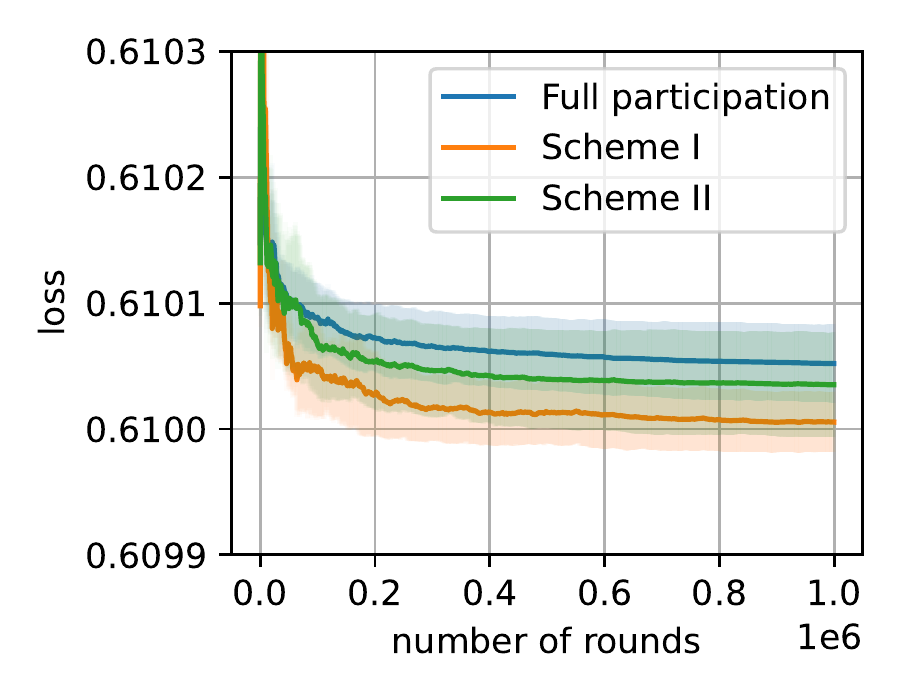}}
    \subfigure[]{\includegraphics[height=0.25\textwidth]{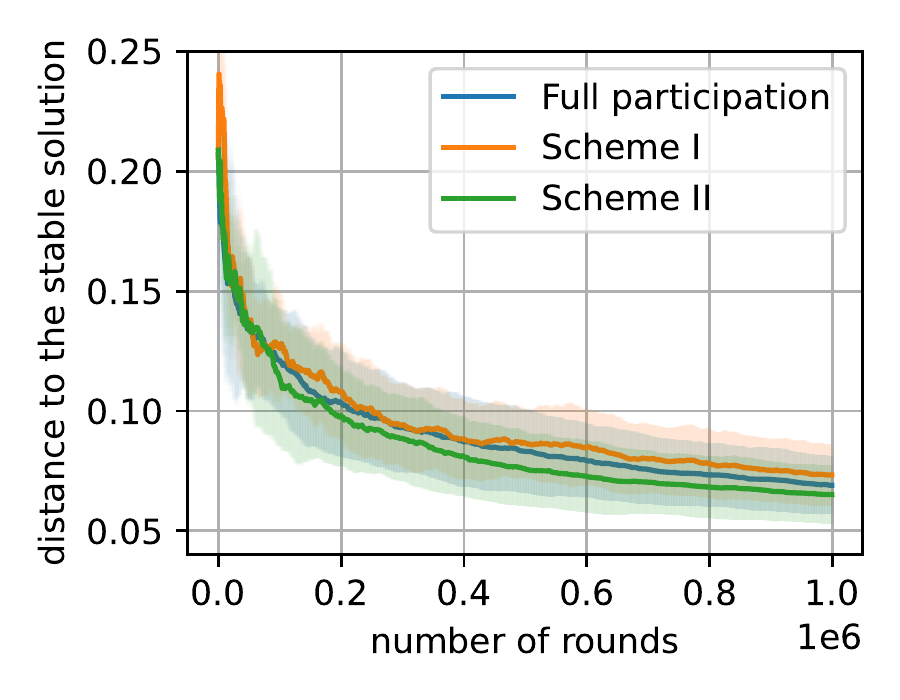}}
    \subfigure[]{\includegraphics[height=0.25\textwidth]{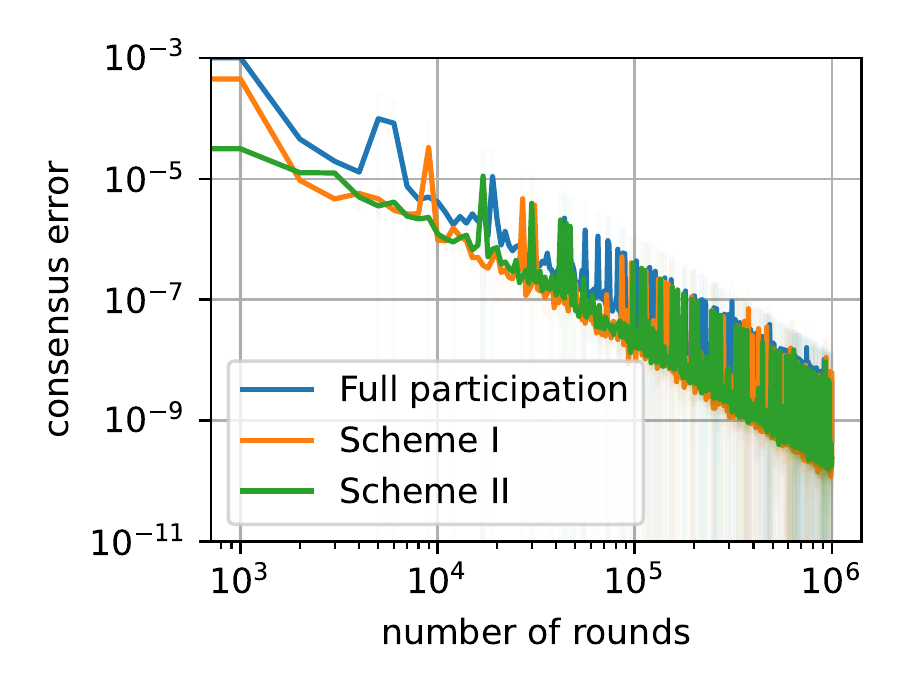}}
    \caption{The losses (a) and distances (b) to the PS solution for the full participation, Scheme I and Scheme II using batch size 1 in client gradient descent.}
    \label{fig:credit_b1}
\end{figure}

\begin{figure}[htbp]
    \centering
    \subfigure[]{\includegraphics[height=0.25\textwidth]{figures/credit/b4e5_credit_loss.pdf}}
    \subfigure[]{\includegraphics[height=0.25\textwidth]{figures/credit/b4e5_credit_dtheta.pdf}}
    \subfigure[]{\includegraphics[height=0.25\textwidth]{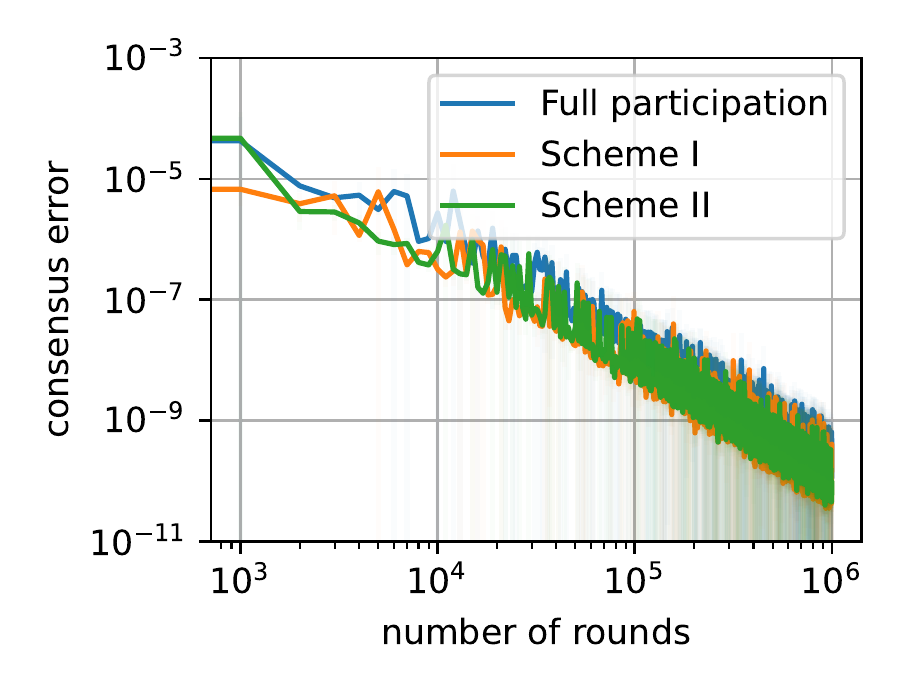}}
    \caption{Same as Figure~\ref{fig:credit_b1}, but using batch size 4.}
    \label{fig:credit_b4}
\end{figure}

\begin{figure}[htbp]
    \centering
    \subfigure[]{\includegraphics[height=0.25\textwidth]{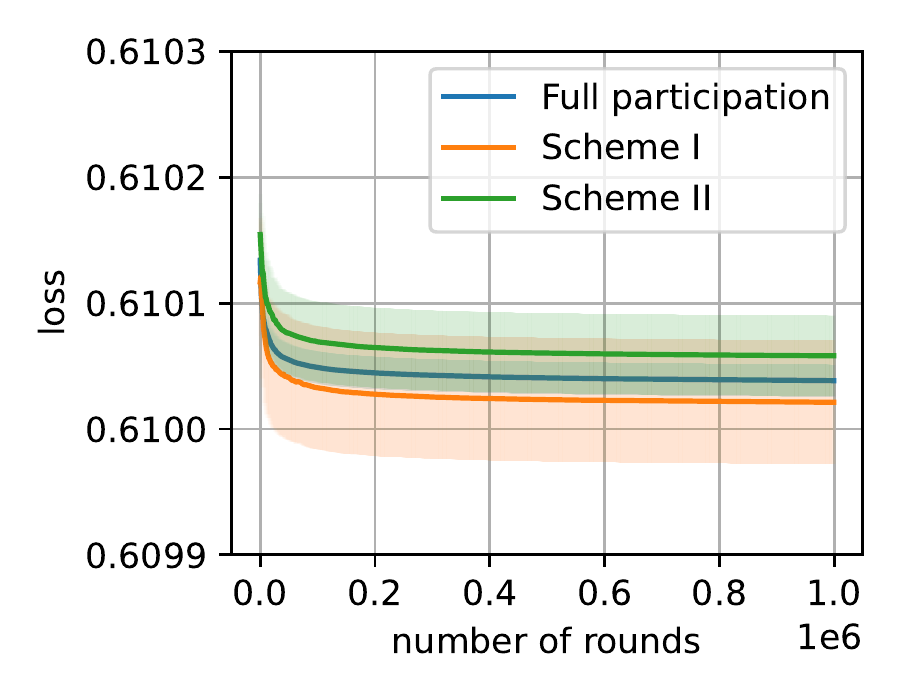}}
    \subfigure[]{\includegraphics[height=0.25\textwidth]{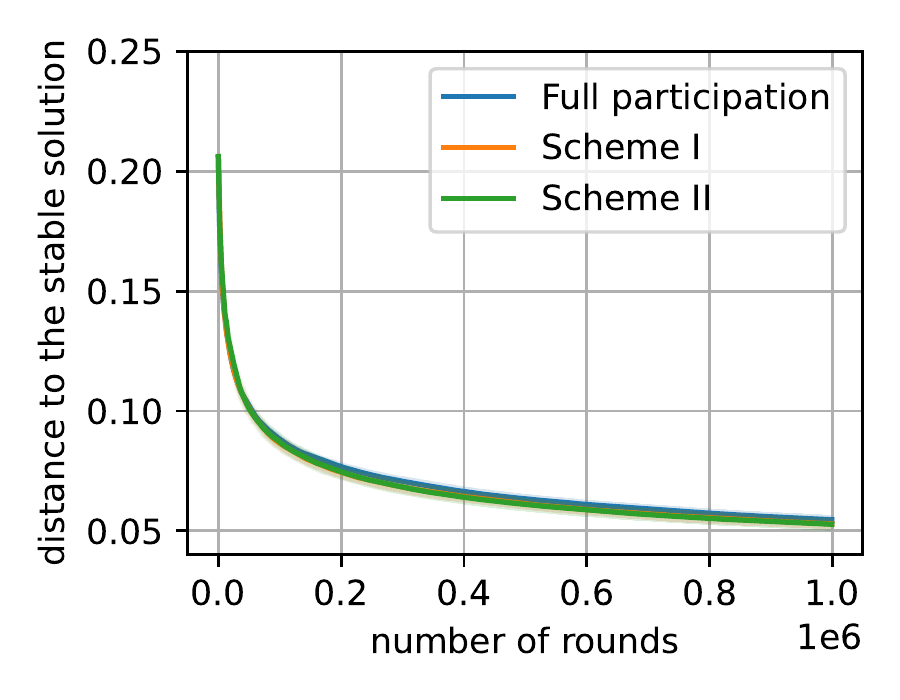}}
    \subfigure[]{\includegraphics[height=0.25\textwidth]{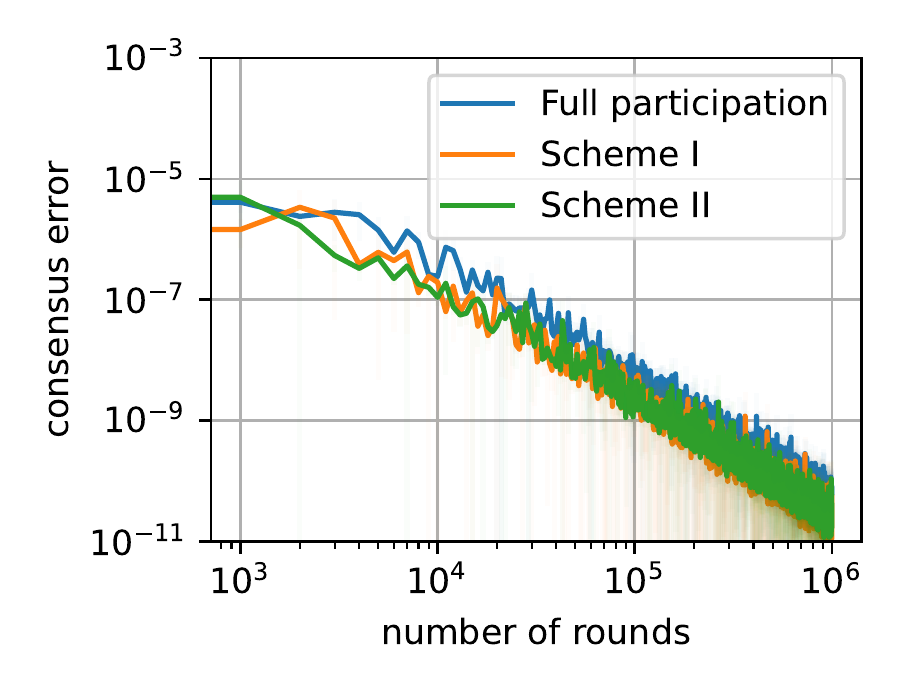}}
    \caption{Same as Figure~\ref{fig:credit_b1}, but using batch size 16.}
    \label{fig:credit_b16}
\end{figure}

Figure \ref{fig:credit_b1} shows the losses and distances to the PS solution for the full participation, Scheme I and Scheme II using batch size 1 in client gradient descent. Figure \ref{fig:credit_b4} and \ref{fig:credit_b16} show the same figure with batch size 4 and 16, respectively. The scales of y axes are set equal for convenience of comparison. Using a larger batch size improves the convergence speed for all three schemes, especially for the two schemes of partial participation, both converging as fast as the full participation with batch size 16.

\begin{figure}
    \centering
    \subfigure[Full participation, loss]{\includegraphics[height=0.25\textwidth]{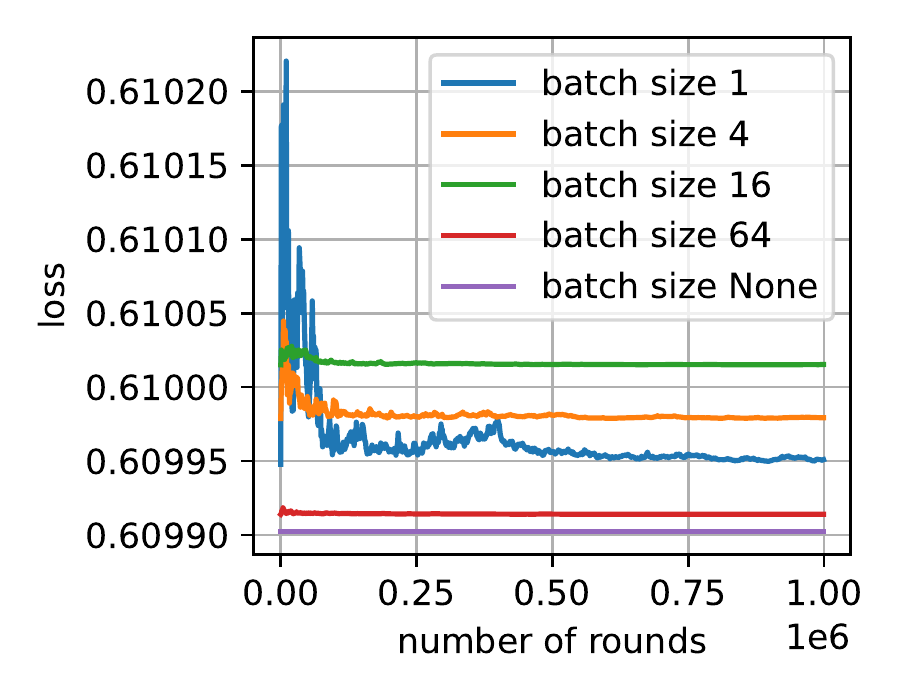}}
    \subfigure[Full participation, distance to PS solution]{\includegraphics[height=0.25\textwidth]{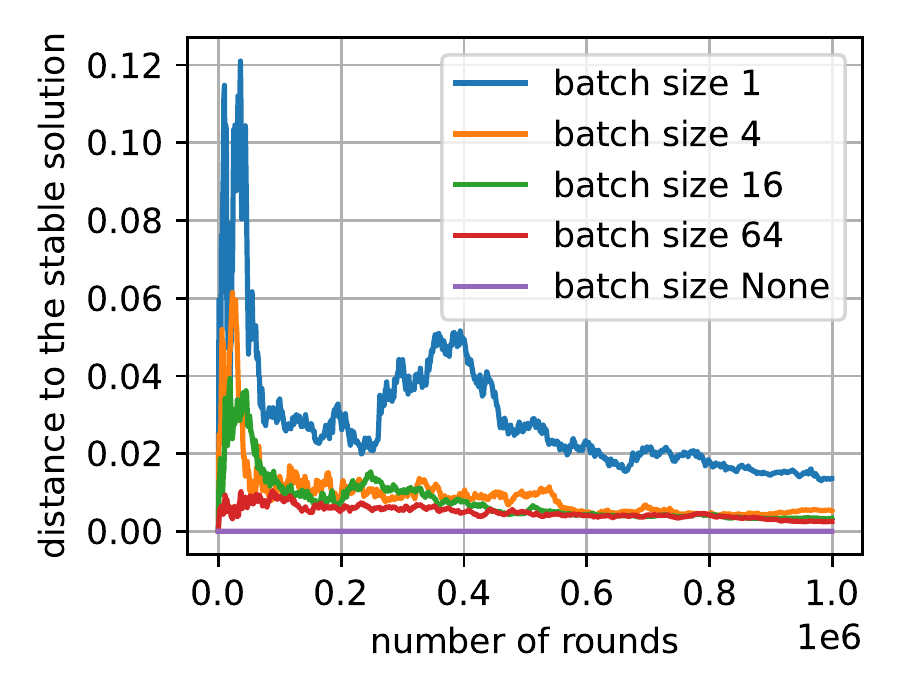}} \\
    \subfigure[Scheme I, loss]{\includegraphics[height=0.25\textwidth]{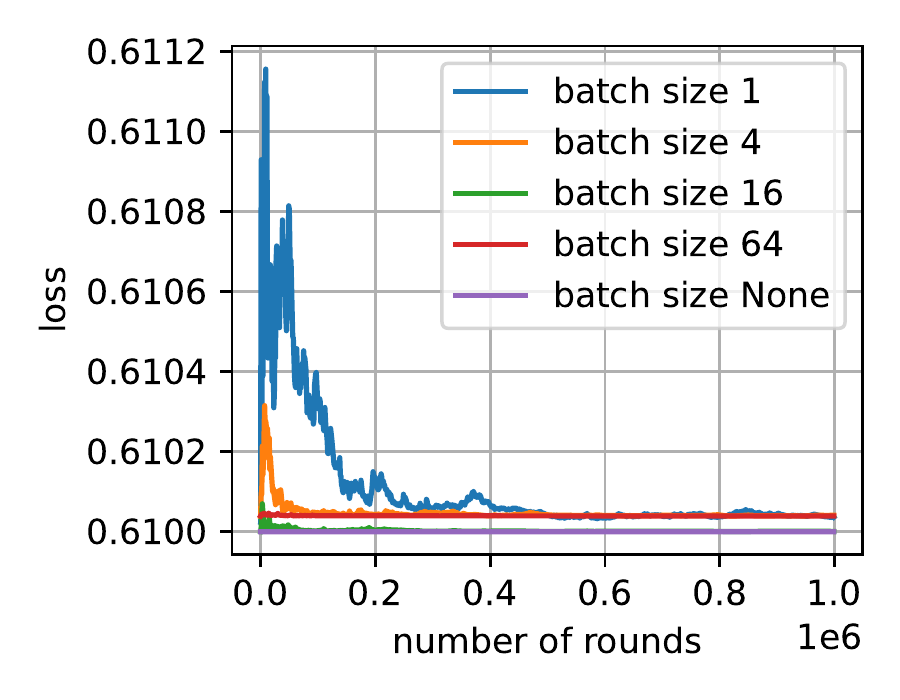}}
    \subfigure[Scheme I, distance to PS solution]{\includegraphics[height=0.25\textwidth]{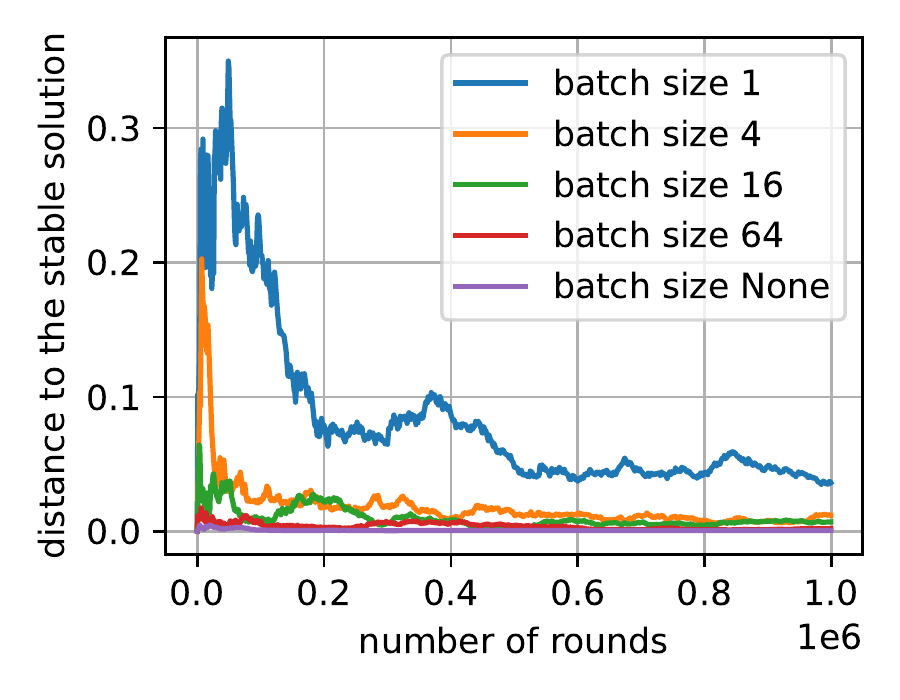}} \\
    \subfigure[Scheme II, loss]{\includegraphics[height=0.25\textwidth]{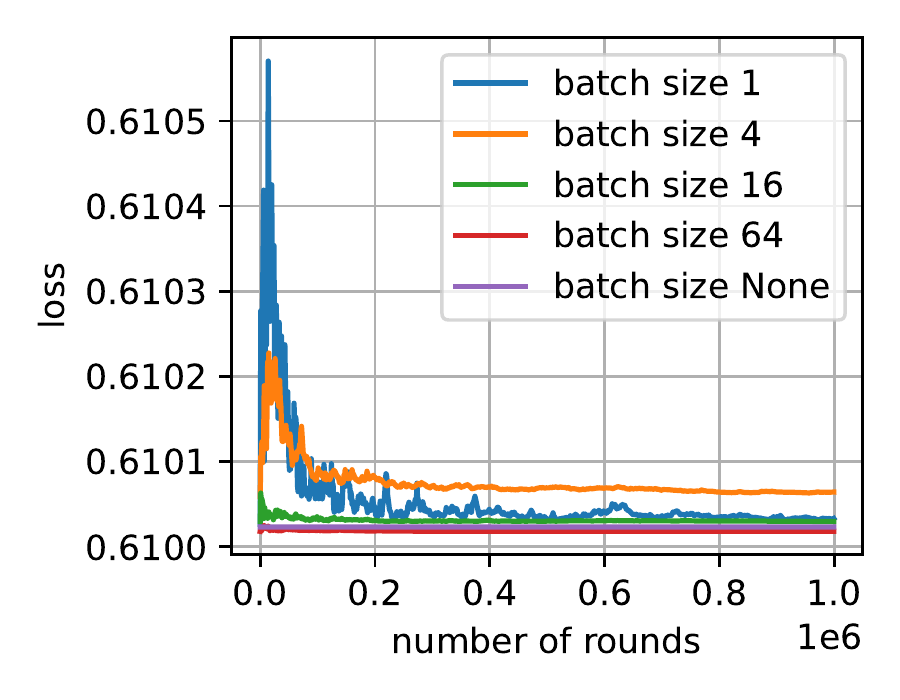}}
    \subfigure[Scheme II, distance to PS solution]{\includegraphics[height=0.25\textwidth]{figures/credit/init_PS_i_theta_dist.pdf}} \\
    \caption{The loss functions and the distance to $\boldsymbol{\theta}^{PS}$ of \texttt{P-FedAvg} initialized with $\boldsymbol{\theta}^{PS}$.}
    \label{fig:init_PS}
\end{figure}

To study how batch batch size affect the convergence, we initialize \texttt{P-FedAvg} with $\boldsymbol{\theta}^{PS}$, the solution that minimizes the performative objective function. Due to the randomness of minibatch stochastic descent, we expect the parameter to deviate from $\boldsymbol{\theta}^{PS}$ and gradually stabilize back to $\boldsymbol{\theta}^{PS}$ as the algorithm proceeds with decaying step sizes. It can be seen from Figure \ref{fig:init_PS} (b), (c) and (f) that it is indeed the case for batch sizes larger than 1. This motivates our choice of a batch size larger than 1.

\end{document}